\documentclass[final]{cvpr}

\usepackage{times}
\usepackage{epsfig}
\usepackage{graphicx}
\usepackage{amsmath}
\usepackage{amssymb}
\usepackage{paralist}
\usepackage{caption}
\usepackage{subcaption}
\usepackage{color,soul}
\usepackage{xcolor}
\usepackage{multirow}
\usepackage[titletoc,page]{appendix}
\usepackage{siunitx}
\usepackage{pifont}
\usepackage{dsfont}
\usepackage{mathtools}

\graphicspath{ {./fig/} }

\usepackage[pagebackref=true,breaklinks=true,colorlinks,bookmarks=false]{hyperref}

\def\cvprPaperID{6689} 
\def\confYear{CVPR 2021}

\renewcommand{\vec}[1]{\mathbf{#1}}
\newcommand{\set}[1]{\mathcal{#1}}

\DeclareMathOperator*{\argmax}{arg\,max}
\DeclareMathOperator*{\argmin}{arg\,min}

\renewcommand{\wrt}{w.r.t.\ }
\newcommand*{\tran}{^{\mkern-1.5mu\mathsf{T}}}

\newcommand*{\inverse}{^{\mkern-1.5mu{-1}}}
\newcommand*{\pinv}{^{\mkern-1.5mu{+}}}

\definecolor{nicegreen}{HTML}{99C000}
\definecolor{nicegreen2}{HTML}{81d41a}
\definecolor{niceyellow}{HTML}{FDCA00}
\definecolor{niceorange}{HTML}{f5a300}
\definecolor{niceblue}{HTML}{0083cc}
\definecolor{nicepurple}{HTML}{a60084}

\definecolor{c_red}{HTML}{e6194b}
\definecolor{c_blue}{HTML}{4363d8}
\definecolor{c_green}{HTML}{aaffc3}
\definecolor{c_purple}{HTML}{911eb4}
\definecolor{c_cyan}{HTML}{46f0f0}
\definecolor{c_orange}{HTML}{f58231}

\newcommand*{\affaddr}[1]{#1} 
\newcommand*{\affmark}[1][*]{\textsuperscript{#1}}

\newcommand{\cmark}{\ding{51}}%
\newcommand{\xmark}{\ding{55}}%

\begin{document}

\title{Cuboids Revisited: Learning Robust 3D Shape Fitting to Single RGB Images}

\author{%
Florian Kluger\affmark[1], Hanno Ackermann\affmark[1], Eric Brachmann\affmark[2], Michael Ying Yang\affmark[3], Bodo Rosenhahn\affmark[1]\\
\normalsize
\affaddr{\affmark[1]Leibniz University Hannover}, 
\affaddr{\affmark[2]Niantic}, 
\affaddr{\affmark[3]University of Twente}\\
\normalsize
\vspace{-1em}
}
\date{}

\maketitle

\begin{abstract}

Humans perceive and construct the surrounding world as an arrangement of simple parametric models. 
In particular, man-made environments commonly consist of volumetric primitives such as cuboids or cylinders.
Inferring these primitives is an important step to attain high-level, abstract scene descriptions. 
Previous approaches directly estimate shape parameters from a 2D or 3D input, and are only able to reproduce simple objects, yet unable to accurately parse more complex 3D scenes. 
In contrast, we propose a robust estimator for primitive fitting, which can meaningfully abstract real-world environments using cuboids. 
A RANSAC estimator guided by a neural network fits these primitives to 3D features, such as a depth map. 
We condition the network on previously detected parts of the scene, thus parsing it one-by-one. 
To obtain 3D features from a single RGB image, we additionally optimise a feature extraction CNN in an end-to-end manner.
However, naively minimising point-to-primitive distances leads to large or spurious cuboids occluding parts of the scene behind. 
We thus propose an occlusion-aware distance metric correctly handling opaque scenes.
The proposed algorithm does not require labour-intensive labels, such as cuboid annotations, for training.
Results on the challenging NYU Depth v2 dataset demonstrate that the proposed algorithm successfully abstracts cluttered real-world 3D scene layouts.
\end{abstract}

\vspace{-1em}
\section{Introduction}
\label{sec:intro}

\begin{figure}	
		\centering
		\begin{subfigure}[t]{0.99\linewidth}
			\centering
			\includegraphics[width=0.49\linewidth]{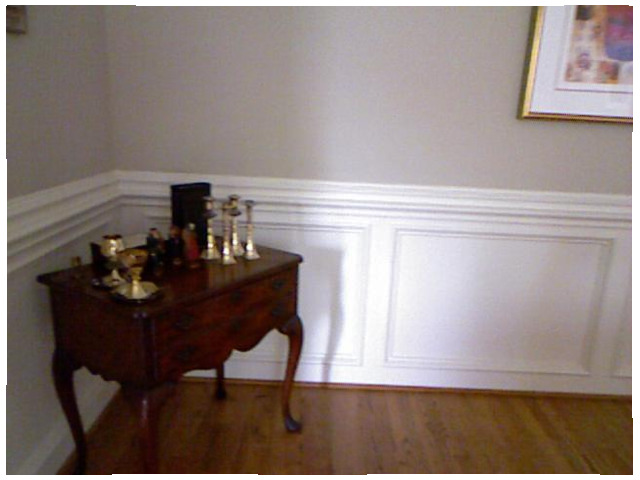}
			\includegraphics[width=0.49\linewidth]{fig/teaser/1171_original.jpg}
			\vspace{-.5em}
			\caption{Input images}
			\vspace{.35em}
		\end{subfigure}
		\begin{subfigure}[t]{0.99\linewidth}
			\centering
			\includegraphics[width=0.49\linewidth]{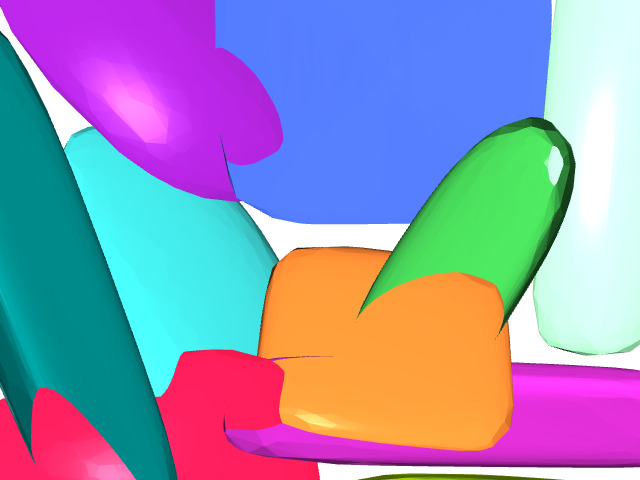}
			\includegraphics[width=0.49\linewidth]{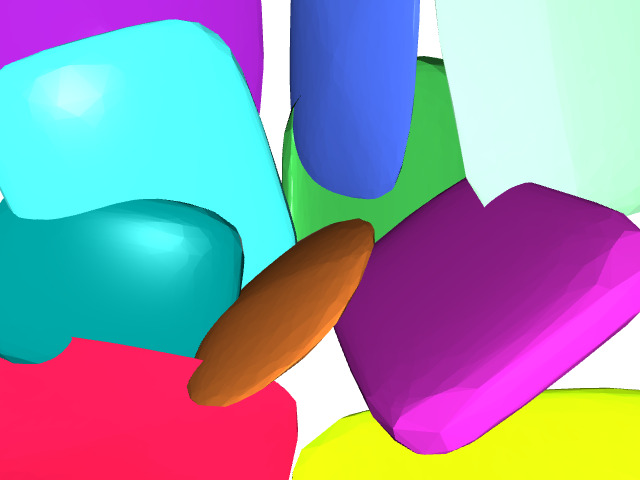}
			\vspace{-.5em}
			\caption{Recovered superquadrics~\cite{paschalidou2019superquadrics}}
			\vspace{.35em}
		\end{subfigure}
		\begin{subfigure}[t]{0.99\linewidth}
			\centering
			\includegraphics[width=0.49\linewidth]{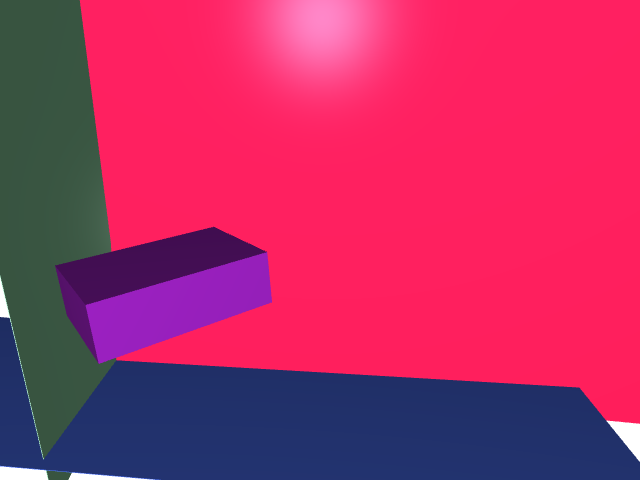}
			\includegraphics[width=0.49\linewidth]{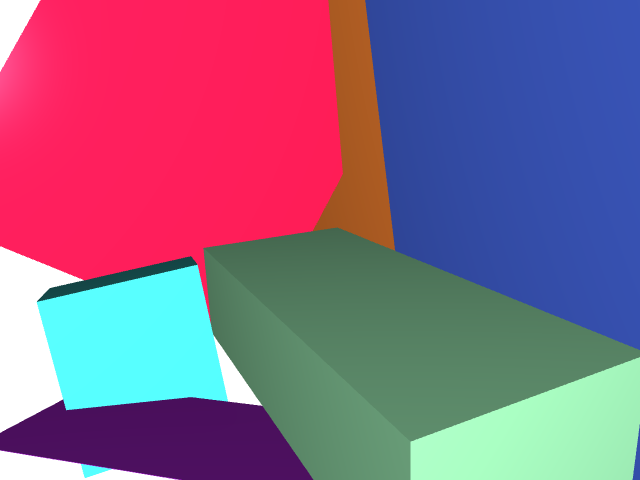}
			\vspace{-.5em}
			\caption{Recovered cuboids (ours)}
			\vspace{.35em}
		\end{subfigure}
		\vspace{-.5em}
		\caption{\textbf{Primitive-based Scene Abstractions:} We parse images of real-world scenes (a) in order to generate abstractions of their 3D structure using cuboids (c). Our method is capable of capturing scene structure more accurately than previous work~\cite{paschalidou2019superquadrics} based on superquadrics (b). }
		\vspace{-1.25em}
		\label{fig:teaser}
\end{figure}   

Humans tend to create using simple geometric forms. 
For instance, a house is made from bricks and squared timber, and a book is a cuboidal assembly of rectangles.
Consequently, it appears that humans also visually abstract environments by decomposing them into arrangements of cuboids, cylinders, ellipsoids and other simple volumetric primitives~\cite{biederman1987recognition}. 
Such an abstraction of the 3D world is also very useful for machines with visual perception. 
Scene representation based on geometric shape primitives has been an active topic since the very beginning of Computer Vision. 
In 1963, \emph{Blocks World}~\cite{roberts1963machine} from Larry Roberts was one of the earliest approaches for qualitative 3D shape recovery from 2D images using generic shape primitives.
In recent years, with rapid advances in the field of deep learning, 
high-quality 3D reconstruction from single images has become feasible. 
Most approaches recover 3D information such as depth~\cite{Eigen2014:DepthMap} and meshes~\cite{WangZLFLJ18} from RGB images.
Fewer works consider more parsimonious 3D shape descriptions such as cuboids~\cite{tulsiani2017learning} or superquadrics~\cite{paschalidou2019superquadrics, Paschalidou2020:Decomposition}.
These 3D shape parsers work well for isolated objects, but do not generalise to complex real-world scenes (cf.~Fig.~\ref{fig:teaser}). 

Robust model fitting algorithms such as RANSAC~\cite{Fischler1981:RANSAC} and its many  derivatives~\cite{barath2018graph, Chin2014:AccHypGen, PurChi2014a} have been used to fit low-dimensional parametric models, such as plane homographies, fundamental matrices or geometric primitives, to real-world noisy data.
Trainable variants of RANSAC~\cite{Brachmann2017:DSAC, Brachmann2019:NGRANSAC, KluBra2020} use a neural network to predict sampling weights from data. 
They require fewer samples and are more accurate. 

Leveraging advances in the fields of 
single image 3D reconstruction and robust multi-model fitting, 
we present a novel approach for robustly parsing real-world scenes using 3D shape primitives, such as cuboids.
A trainable RANSAC estimator fits these primitives to 3D features, such as a depth map. 
We build upon the estimator proposed in \cite{KluBra2020}, and extend it by predicting multiple sets of RANSAC sampling weights concurrently.
This enables our method to distinguish between different structures in a scene more easily.
We obtain 3D features from a single RGB image using a CNN, and show how to optimise this CNN in an end-to-end manner.
Our training objective is based on geometrical consistency with readily available 3D sensory data.

During primitive fitting, a naive maximisation of inlier counts considering point-to-primitive distances causes the algorithm to detect few but excessively large models. 
We argue that this is due to parts of a primitive surface correctly representing some parts of a scene, while other parts of the same primitive wrongly occlude other parts of the scene.
Thus, points should not be assigned to primitive surfaces which cannot be seen by the camera due to occlusion or self-occlusion. 
We therefore propose an occlusion-aware distance and a corresponding occlusion-aware inlier count.  

As no closed-form solution exists to calculate cuboid parameters, we infer them by numerical optimisation.
However, backpropagation through this optimisation is numerically unstable and computationally costly.
We therefore analytically derive the gradient of primitive parameters \wrt the features used to compute them.
Our  gradient computation allows for end-to-end training without backpropagation through the minimal solver itself.
We demonstrate the efficacy of our method on the challenging real-world NYU Depth v2 dataset~\cite{silberman2012indoor}. 

\vspace{.1em}
In summary, our \textbf{contributions}\footnote{Source code is available at: \url{https://github.com/fkluger/cuboids_revisited}} are as follows:
\begin{compactitem}
\item A 3D scene parser which can process more complex real-word scenes than previous works on 3D scene abstraction.
\item An occlusion-aware distance metric for opaque scenes.
\item Analytical derivation of the gradient of cuboids \wrt input features, in order to circumvent infeasible backpropagation through our minimal solver, thus enabling end-to-end training. 
\item Our method does not require labour-intensive labels, such as cuboid or object annotations, and can be trained on readily available sensory data instead.
\end{compactitem}
 
\begin{figure*}	
\centering
\includegraphics[width=0.95\linewidth]{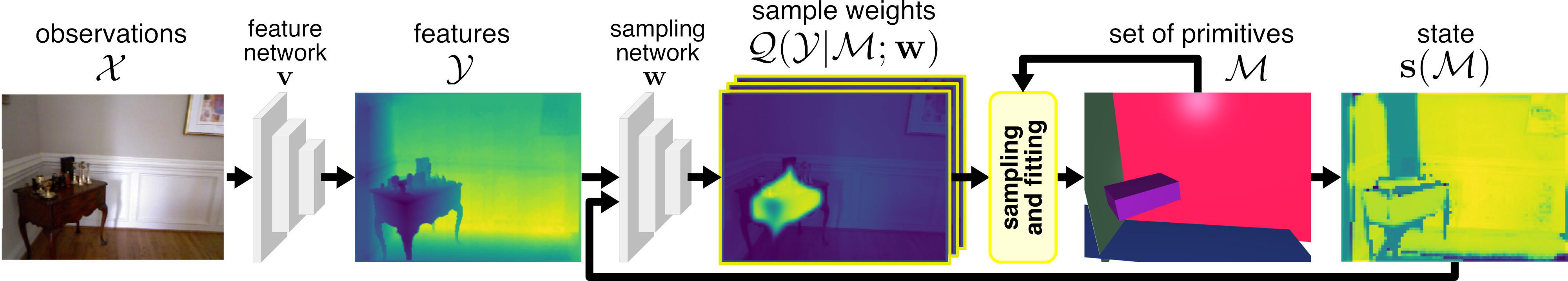}
\vspace{-.5em}
\caption{ \textbf{Overview:} 
Given observations $\set{X}$ (RGB image), we predict 3D features $\set{Y}$ (depth map) using a neural network with parameters $\vec{v}$.
Conditioned on a state $\vec{s}$, a second neural network with parameters $\vec{w}$ predicts sampling weights $p(\vec{y}|\vec{s}; \vec{w}) \in \set{Q}$ for each feature $\vec{y} \in \set{Y}$.
Using these weights, a RANSAC-based estimator samples minimal sets of features, and generates primitive (cuboid) hypotheses $\set{H}$.
It selects the best hypothesis $\vec{\hat{h}} \in \set{H}$ and appends it to the set of previously recovered primitives $\set{M}$.
We update the state $\vec{s}$ based on $\set{M}$ and repeat the process in order to recover all primitives step-by-step.
}
\vspace{-1.2em}
\label{fig:system}
\end{figure*} 

\section{Related Work}
\label{sec:related}

\vspace{-2mm}
\paragraph{Monocular Depth Estimation.}
Monocular depth estimation has become feasible with advances in deep learning and progressed steadily over recent years.
Instead of fully supervised learning~\cite{Eigen2014:DepthMap, Liu2015:Depth, lee2019big}, later works approach the problem via semi-supervised learning~\cite{Kuznietsov2017:Semi}, unsupervised learning~\cite{Garg0216:UnsupervisedDepth, Mahjourian2018:Unsupervised, Godard2017:DepthEstimation, Yin2018:GeoNet} or self-supervised learning~\cite{Godard2019:SelfSupervised, Ma2019:SelfSupervised}. 
These methods predict dense 3D information instead of the more parsimonious primitive based descriptions we are interested in.
However, we show how to leverage monocular depth estimation to this end.

\vspace{-5mm}
\paragraph{Single Image 3D Reconstruction.}
The authors of~\cite{Liu2018:PlaneNet, liu2019planercnn} propose methods for piece-wise planar reconstruction from single images.
Older works~\cite{criminisi2000single} rely on geometric image features, such as vanishing points~\cite{kluger2017deep} or horizon lines~\cite{kluger2020temporally}.
Like depth estimation, however, these methods can only describe the visible surfaces of a scene and do not capture volumetric characteristics.
Although some methods~\cite{Pan2019:Mesh, groueix2018papier, WangZLFLJ18, mescheder2019occupancy} perform 3D mesh reconstruction, they demonstrably only work on images of isolated single objects.
The authors of~\cite{Nie2020:Total3dUnderstanding, Zhang2017:DeepContext, gkioxari2019mesh} present mesh reconstructions for multiple objects in real-world scenarios.
However, the latter methods require ground-truth object meshes and object class labels for training, which are both costly to obtain.
Our approach, on the other hand, requires neither.

\vspace{-5mm}
\paragraph{3D Bounding Box Regression.}
For predefined object classes, 3D bounding box regression is a well investigated topic in 3D object detection for RGB images~\cite{Ren2016:3dObjectDetection, Dwibedi2016:Beyond2dBBs,Kehl2018:3dDetection,Mousavian2017:3dBB},
RGB-D images~\cite{Qi2018:FrustumPNets}, and 3D point clouds~\cite{Shi2019:PRCNN,Thomas2019ICCV}. 
These works not only detect bounding boxes but also classify objects. 
During training, target 3D boxes as well as category labels are needed. 
In contrast, our proposed method does not require such annotations.

\vspace{-5mm}
\paragraph{Robust Multi-Model Fitting.}
Robust fitting of parametric models, for instance using RANSAC~\cite{Fischler1981:RANSAC}, is a key problem in Computer Vision. 
Several approaches~\cite{barath2018multi,barath2019progressive,pham2014interacting, amayo2018geometric} fit \emph{multiple instances} of a model simultaneously by optimising an energy-based functional.
Moreover, the algorithms in \cite{barath2018multi, barath2019progressive, magri2019fitting} approach the \emph{multi-class} problem, \ie when models of multiple types may fit the data. 
Sampling efficiency is improved in~\cite{Brachmann2017:DSAC, Brachmann2019:NGRANSAC, KluBra2020, PurChi2014a}. 
In this work, we extend~\cite{KluBra2020} for robustly parsing real-world scenes using 3D shape primitives.
Unlike~\cite{KluBra2020}, we not just learn the parameters of the sampling weight from data, but train it conjointly with a depth estimation network in an end-to-end manner. 

\vspace{-5mm}
\paragraph{3D Primitive Fitting.}
Although an old topic in Computer Vision (cf. Blocks World~\cite{roberts1963machine}), 3D shape recovery from 2D images using volumetric primitives is still considered non-trivial~\cite{GuptaEfrosHebert_ECCV10}. 
The methods of~\cite{Xiao2012:Localizing} and~\cite{Jiang2013:Linear,Lin2013:Holistic} localise cuboids in images and RGB-D data, respectively.
While~\cite{Jiang2013:Linear,Lin2013:Holistic} use the challenging NYU Depth v2 dataset~\cite{silberman2012indoor}, all three works require ground truth cuboids.

The works most related to ours are~\cite{tulsiani2017learning,paschalidou2019superquadrics,Paschalidou2020:Decomposition}.
In~\cite{tulsiani2017learning}, the authors propose a method for 3D shape abstraction using cuboids. 
It is based on a neural network, which directly regresses cuboid parameters from either an RGB image or a 3D input, and is trained in an unsupervised fashion.
Similarly, the authors of~\cite{paschalidou2019superquadrics} decompose objects into sets of superquadric surfaces.
They extend this to hierarchical sets of superquadrics in~\cite{Paschalidou2020:Decomposition}.
However, these approaches are only evaluated on relatively simple shapes, such as those in ShapeNet~\cite{Chang2015:ShapeNet}. 
While they yield good results for such objects, they are unable to predict reasonable primitive decompositions for complex real-world scenes  (cf.~Sec.~\ref{sec:exp}).

\vspace{-.5em}
\section{Method}
\vspace{-.5em}
\label{sec:method}
We formulate the problem of abstracting a target scene $\set{Z}$ by fitting a set of shape primitives $\set{M}$ to features $\set{Y}$, which may either be provided directly, or extracted from observations $\set{X}$.
Here, $\set{Z}$ represents the ground truth depth or its corresponding point cloud, and $\set{X}$ is an RGB image of the scene aligned with $\set{Z}$.
Features $\set{Y}$ are either equal to $\set{Z}$, or estimated via a function $\set{Y}=f_\vec{v}(\set{X})$ if $\set{Z}$ is unknown.
We implement function $f_\vec{v}$ as a neural network with parameters $\vec{v}$. 
Each primitive $\vec{h} \in \set{M}$ is a cuboid with variable size and pose.
See Fig.~\ref{fig:system} for an example of $\set{X}$, $\set{Y}$ and $\set{M}$.

For primitive fitting, we build upon the robust multi-model estimator of Kluger et al.~\cite{KluBra2020}. 
This estimator predicts sampling weights $\vec{p} = f_{\vec{w}}(\set{Y}, \vec{s})$ from observations $\set{Y}$ and a state $\vec{s}$ via a neural network with parameters $\vec{w}$, which are learnt from data.
It samples minimal sets of features from $\set{Y}$ according to $\vec{p}$, and fits primitive hypotheses $\set{H}$ via a minimal solver $f_h$.
From these hypotheses, it selects the best primitive $\vec{\hat{h}} \in \set{H}$ according to an inlier criterion, and adds it to the current set of primitives $\set{M}$. 
Based on $\set{M}$, it then updates the state $\vec{s}$ and predicts new sampling weights $\vec{p}$ in order to sample and select the next primitive.
This process, as visualised in Fig.~\ref{fig:system}, is repeated until all primitives have been found one by one.
Unlike~\cite{KluBra2020}, we learn the parameters of the sampling weight predictor $f_{\vec{w}}$ jointly with the feature extractor $f_{\vec{v}}$ in an end-to-end manner. 
We show how to achieve this despite backpropagation through the minimal solver $f_h$ being numerically unstable and computationally costly.
In addition, we generalise $f_{\vec{w}}$ so that it predicts multiple sets of sampling weights $\set{Q} = \{ \vec{p}_1, \dots, \vec{p}_Q \} $ concurrently, which enables it to distinguish between different primitive instances more effectively.

The scenes we are dealing with in this work have been captured with an RGB-D camera.
We therefore do not have full 3D shapes available as ground-truth for the scenes.
Instead, we only have 2.5D information, \ie 3D information for visible parts of the scene, and no information about the occluded parts.
If not taken into account, this fact can lead to spurious, oversized or ill-fitting primitives being selected, as visualised in Fig.~\ref{fig:occlusion}.
For this reason we present occlusion-aware distance and inlier metrics.

\vspace{-.25em}
\subsection{Feature Extraction}
\vspace{-.25em}
In order to fit 3D shapes such as cuboids to an RGB image $\set{X}$, we have to extract 3D features $\set{Y}$ from $\set{X}$.
We employ a depth estimator $f_{\vec{v}}$ which gives us the desired features $\set{Y} = f_{\vec{v}}(\set{X})$ in form of a pixel-wise depth map.
The depth estimator is realised as a convolutional neural network with parameters $\vec{v}$.
We then convert $\set{Y}$ into a point cloud via backprojection using known camera intrinsics $\vec{K}$.

\vspace{-.25em}
\subsection{Cuboid Parametrisation}
\vspace{-.25em}
A generic cuboid is described by its {shape} $(a_x, a_y, a_z)$ and {pose} $(\vec{R}, \vec{t})$. 
The shape corresponds to its width, height and length in a cuboid-centric coordinate system, while its pose translates the latter into a world coordinate system.
We represent the rotation $\vec{R}$ in angle-axis notation $\vec{r} = \theta \vec{u}$.
Each cuboid thus has nine degrees of freedom, and we require minimal sets of $C=9$ points to estimate them.

\vspace{-.75em}
\subsubsection{Point-to-Cuboid Distance }
\vspace{-.25em}
\label{subsubsec:point_to_cuboid_distance}
When computing the distance between point $\vec{y} = (x, y, z)\tran$ and cuboid $\vec{h} = (a_x, a_y, a_z, \vec{R}, \vec{t})$, we first translate $\vec{y}$ into the cuboid-centric coordinate frame:
$  \vec{\hat{y}} = \vec{R}(\vec{y} - \vec{t}) \, .$
We then compute its squared distance to the cuboid surface:
\vspace{-2mm}
\begin{align}
&d( \vec{h}, \vec{y})^2 =  
 \max(\min(a_x-|\hat{x}|, a_y-|\hat{y}|, a_z-|\hat{z}|), 0)^2 + \nonumber  \\
&\!\max(|\hat{x}|-a_x, 0)^2 \! + \! \max(|\hat{y}|-a_y, 0)^2 \! + \! 
 \max(|\hat{z}|-a_z, 0)^2 \nonumber  .
\end{align}
Similarly, we can compute the distances to any of the six individual sides of the cuboid.
Defining, for example, the plane orthogonal to the $x$-axis in its positive direction as the first side, we define the distance of a point $\vec{y}$ to it as:
\vspace{-2mm}
\begin{align}
    d^{\text{P}}_{1}(\vec{h}, \vec{y})^2 = & (\hat{x}-a_x)^2 + \\ & \max(|\hat{y}|-a_y, 0)^2 +  \max(|\hat{z}|-a_z, 0)^2 \, . \nonumber
\end{align}
Distances $d^{\text{P}}_2, \dots, d^{\text{P}}_6$ to the other sides of the cuboid are calculated accordingly.

\vspace{-3mm}
\subsubsection{Occlusion Handling}
\vspace{-1mm}
\label{subsubsec:oa_point_cuboid_distance}

\begin{figure}[http]
\centering
\includegraphics[width=0.5\linewidth]{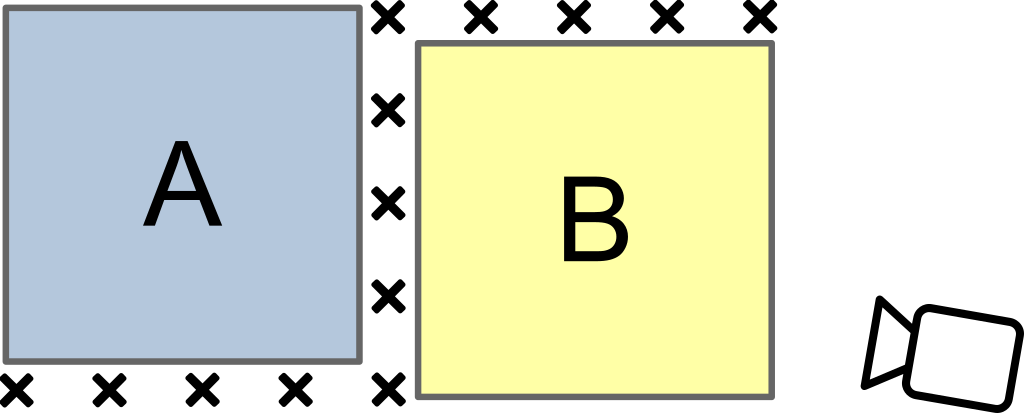}
\caption{ \textbf{Occlusion:} Given are a point cloud (\ding{53}), two cuboids (A and B) and a camera observing the scene. Cuboid A is a better fit since it does not occlude any points.}
\vspace{-10pt}
\label{fig:occlusion}
\end{figure}

When dealing with 2.5D data which only represents visible parts of the scene, simply using minimal point-to-cuboid distances as in Sec.~\ref{subsubsec:point_to_cuboid_distance} is not adequate. 
Fig.~\ref{fig:occlusion} gives an intuitive example:
The mean distance of all points to their closest surface of either cuboid A or B is the same.
However, the {visible} surfaces of cuboid B occlude all points while not representing any structure present in the scene.
Cuboid A, on the other hand, does not occlude any points, and its visible surfaces fit very well to an existing structure.
Although other parts of its surface do not represent any structure either, they are self-occluded and thus of no interest.
So, cuboid A is a much better fit than cuboid B.

\vspace{-1.25em}
\paragraph{Occlusion Detection.}
In order to detect whether a cuboid $\vec{{h}}$ occludes a point $\vec{y}$ from the perspective of a camera with centre $\vec{c}$, we must first translate $\vec{c}$ into the cuboid-centric coordinate frame.
We parametrise the line of sight for a point $\vec{y}$ and determine its intersections with each of the six cuboid planes. Via this procedure we define the following indicator function:
\vspace{-2mm}
\begin{equation}
    \chi_{\text{o}}( \vec{h}, \vec{y}, i) =
\begin{cases}
1 ~&\text{ if }~ i\text{-th plane of } \vec{h}  \text{ occludes }  \vec{y}, \\
0 ~&\text{ else }, \, \text{with } i \in \{1, \dots, 6\} \, .
\end{cases}
\end{equation}

\vspace{-1.25em}
\paragraph{Occlusion-Aware Point-to-Cuboid Distance.}
In order to correctly evaluate scenarios with occlusions, we propose an \emph{occlusion-aware} point-to-cuboid distance:
Given a point $\vec{y}$ and set of cuboids $\set{M} = \{\vec{h}_1, \dots, \vec{h}_{|\set{M}|}\}$, we compute its distance to the most distant occluding surface.
In other words, we need to know the minimal distance $\vec{y}$ would have to travel in order to become visible:
\vspace{-2mm}
\begin{equation}
d_{\text{o}}(\set{M}, \vec{y}) = 
\max_{\vec{h} \in \set{M}, \, i \in \{1, \dots, 6\}} 
\left( \,
\chi_{\text{o}}(\vec{h}, \vec{y}, i) \cdot d^{\text{P}}_{i}(\vec{h}, \vec{y}) 
\, \right) \, . \nonumber
\label{eq:occlusion.metric}
\end{equation}
If $\vec{y}$ is not occluded at all, this distance naturally becomes zero.
We hence define the occlusion-aware distance of a point to a set of cuboids as:
\vspace{-2mm}
\begin{equation}
    d_{\text{oa}}(\set{M}, \vec{y}) = 
    \max 
    \left( 
    \min_{\vec{h} \in \set{M}} d(\vec{h}, \vec{y}), \; d_{\text{o}}(\set{M}, \vec{y}) 
    \right) \, .
\end{equation}

\vspace{-.8em}
\subsection{Robust Fitting}
\label{subsec:robustfitting}
\begin{figure}	
\centering
\includegraphics[width=0.9\linewidth]{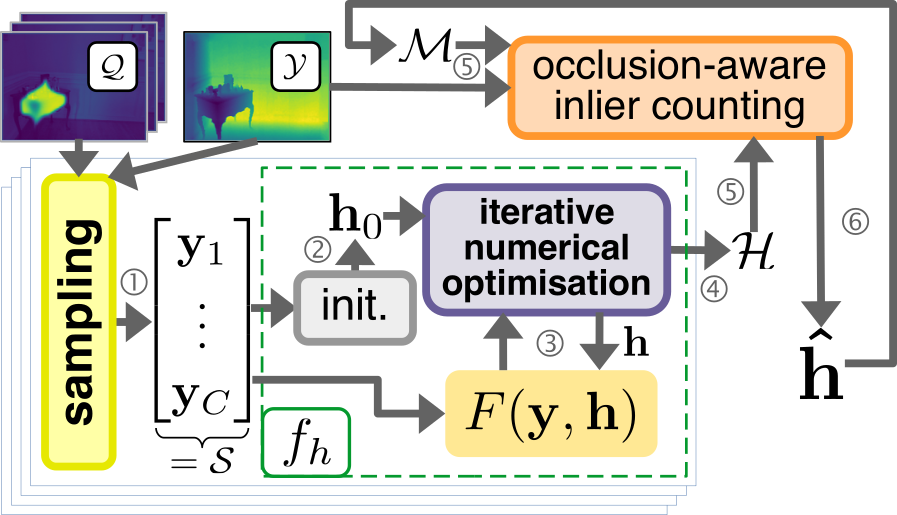}
\caption{ 
\textbf{Sampling and Fitting:}
(1) We sample a minimal set of features $\set{S} \subset \set{Y}$ using sampling weights $\set{Q}$ (Sec.~\ref{par:sampling}).
(2) Within minimal solver $f_h$ (Sec.~\ref{subsubsec:fitting}), we initialise cuboid parameters $\vec{h}_0$.
(3) We optimise these parameters iteratively using Eq.~\ref{eq:def.F}, resulting in a hypothesis $\vec{h}$. 
(4) We compute multiple cuboid hypotheses concurrently, resulting in a set of hypotheses $\set{H}$.
(5) Using occlusion-aware inlier counting (Sec.~\ref{subsec:occlusion_inlier}), we select the best hypothesis $\vec{\hat{h}}$ and (6) add it to the set of recovered cuboids $\set{M}$.
}
\vspace{-10pt}
\label{fig:fitting}
\end{figure}

We seek to robustly fit a set of cuboids $\set{M} = \{\vec{h}_1, \dots, \vec{h}_{|\set{M}|}\}$ to the possibly noisy 3D features $\vec{y} \in \set{Y}$.
To this end, we build upon the robust multi-model fitting approach of Kluger et al.~\cite{KluBra2020}:
\begin{compactenum}
\item We predict sets of sampling weights $\vec{p}$ from data $\set{Y}$ using a neural network $f_{\vec{w}}$.
\item Using these sampling weights, a RANSAC-based~\cite{Fischler1981:RANSAC} estimator generates a cuboid instance $\vec{h}$, which we append to $\set{M}$.
\item Conditioned on $\set{M}$, we update the sampling weights $\vec{p}$ via $f_{\vec{w}}$ and generate the next cuboid instance.
\end{compactenum}
We repeat these steps multiple times, until all cuboids have been recovered one-by-one. 
Fig.~\ref{fig:system} gives an overview of the algorithm, while Fig.~\ref{fig:fitting} depicts the sampling and fitting stage in more detail.

\vspace{-3mm}
\subsubsection{Sampling}
\vspace{-1mm}
\label{par:sampling}
\begin{figure}	
\centering
\includegraphics[width=0.99\linewidth]{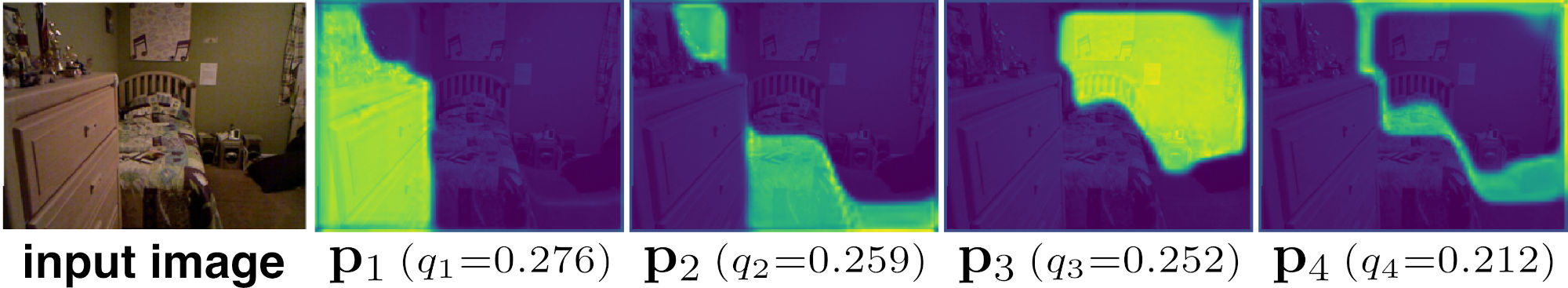}
\vspace{-.3em}
\caption{ \textbf{Sampling Weights: } 
For each image, we predict multiple sets of sampling weights $\set{Q} = \{\vec{p}_1, \dots, \vec{p}_Q\}$ and corresponding selection probabilities $\vec{q} = [q_1, \dots, q_Q]$. In this example, the first three sampling weight sets roughly cover distinct parts of the scene.
The fourth set $\vec{p}_4$ does not, but also has the lowest selection probability. }
\vspace{-1em}
\label{fig:probability_maps}
\end{figure} 
In~\cite{KluBra2020}, one set of sampling weights $\vec{p}(\set{Y} \, | \, \set{M})$ is predicted at each step.
Optimally, these weights should highlight a single coherent structure in $\set{Y}$ and suppress the rest, in order to maximise the likelihood of sampling an all-inlier set of features.
However, we often have multiple important structures present in a scene.
Which structure to emphasise and sample first may therefore be ambiguous, and the approach proposed in~\cite{KluBra2020} struggles with primitive fitting for that reason. 
In order to deal with this, we allow the network to 
instead predict several sets of sampling weights $\set{Q} = \{\vec{p}_1, \dots, \vec{p}_Q\}$ and corresponding selection probabilities $\vec{q}(\set{Y} \, | \, \set{M}) \in \mathbb{R}^Q$, as shown in Fig.~\ref{fig:probability_maps}.
This results in a two-step sampling procedure:
First, we randomly select one of the sampling weight sets $\vec{p} \in \set{Q}$ according to $\vec{q}$.
Then we sample a minimal set of features according to the selected sampling weights $\vec{p}$ in order to generate a cuboid hypothesis.
This allows the neural network $(\set{Q}, \vec{q}) = f_{\vec{w}}(\set{Y} | \set{M})$ to highlight multiple structures at once, without them interfering with each other.
Ideally, $\vec{q}$ contains non-zero values for all sampling weight sets highlighting valid structures, and zero values otherwise.
At worst, it would degenerate to $\vec{q} = \vec{e}_i$, with $i \in \{1, \dots, Q\}$, effectively setting $Q=1$.
We avoid this via regularisation during training (cf. Sec.~\ref{sec:training}).

\vspace{-2mm}
\subsubsection{Fitting}
\vspace{-1.5mm}
\label{subsubsec:fitting}

Given a minimal set of sampled features $\set{S} = \{\vec{y}_1, \dots, \vec{y}_C\} \subset \set{Y}$, we want to find parameters of a cuboid $\vec{h} = f_h(\set{S})$ such that it fits these features optimally.
We define the minimal solver $f_h$ via this objective function:
\vspace{-2mm}
\begin{equation}
    F(\vec{y}, \vec{h}) = d(\vec{h}, \vec{y})^2 \cdot (a_x + a_y + a_z) \, .
    \label{eq:def.F}
\vspace{-2mm}
\end{equation}
As the size of the cuboid may be ambiguous when dealing with 2.5D data, we add the regularisation term $(a_x + a_y + a_z)$ in order to favour smaller cuboids.
Unfortunately, no closed form solution for $\vec{h} \in \argmin_{\vec{h}} \| F(\set{S}, \vec{h}) \|_1$ exists. 
Starting from an initial solution\footnote{Please refer to the supplementary (Sec.~A.5) for details.} $\vec{h}_0$, we therefore approximate it via an iterative numerical optimisation method such as \mbox{L-BFGS}~\cite{liu89} or Adam~\cite{KingmaB14}. 

\vspace{-2.5mm}
\subsubsection{Occlusion-Aware Inlier Counting}
\vspace{-1.5mm}
\label{subsec:occlusion_inlier}
In order to select the cuboid hypothesis $\vec{h} \in \set{H}$  which fits best to features $\set{Y}$, given a set of existing cuboids $\set{M}$, we need to define an inlier function $f_{\text{I}}(\vec{y}, \vec{h})$.
We could naively take $f_{\text{I}}(\vec{y}, \vec{h}) \in [0, 1]$, with $f_{\text{I}}(\vec{y}, \vec{h}) = 1$ if feature $\vec{y}$ is well represented by cuboid $\vec{h}$, and $f_{\text{I}}(\vec{y}, \vec{h}) = 0$ otherwise.
However, as described in Sec.~\ref{subsubsec:oa_point_cuboid_distance}, we want to avoid cuboids which create occlusions.
Hence, we define an occlusion-aware inlier function $f_{\text{OAI}}(\vec{y}, \set{M}) \in [-1, 1]$ with the additional property $f_{\text{OAI}}(\vec{y}, \set{M}) = -1$ if $\vec{y}$ is occluded by cuboids in $\set{M}$, but \emph{only} if it is occluded by cuboid sides to which it is not also an inlier. 
We define:
\vspace{-2mm}
  $$  f_{\text{IO}}(\vec{y}, \vec{h}, i) = f_{\text{I}}(\vec{y}, \vec{h}, i) - \chi_{\text{o}}(\vec{y}, \vec{h}, i) \cdot (1 - f_{\text{I}}(\vec{y}, \vec{h}, i)) \, , $$
to determine whether $\vec{y}$ is an inlier to the $i$-th side of cuboid $\vec{h}$ ($f_{\text{IO}}>0$), occluded by it ($f_{\text{IO}}<0$), or a regular outlier ($f_{\text{IO}}=0$). If $\vec{y}$ is occluded by any cuboid side in $\set{M}$, it must be marked as occluded. Otherwise it should be marked as an inlier to its closest cuboid, or as an outlier:
\begin{equation}
     f_{\text{OAI}}(\vec{y}, \set{M}) = \\
\begin{dcases}
\min_{\vec{h}, i} f_{\text{IO}}(\vec{y}, \vec{h}, i)                                          
    ~&\text{if }~ (\cdot) < 0  \, , \\ 
\max_{\vec{h}, i} f_{\text{IO}}(\vec{y}, \vec{h}, i)
    ~&\text{else} \, ,
\end{dcases}
\end{equation}
with $\vec{h} \in \set{M}$ and $i \in \{1, \dots, 6\}$, \ie minimising or maximising over all sides of all cuboids.
This implies that features which are occluded indeed reduce the inlier count $I_{\text{c}}$, which we use to determine which cuboid hypothesis $\vec{h}$ shall be added to our current set of cuboids $\set{M}$:
\vspace{-.5mm}
\begin{equation}
    I_{\text{c}}(\set{Y}, \set{M} \cup \{\vec{h}\}) = \sum_{\vec{y} \in \set{Y}} f_{\text{OAI}}(\vec{y}, \set{M}\cup \{\vec{h}\}) \, .
\end{equation}

\vspace{-3mm}
\subsection{Training}
\label{sec:training}
As in~\cite{KluBra2020}, we want to optimise parameters $\vec{w}$ of the sampling weight estimation network in order to increase the likelihood of sampling all-inlier minimal sets of features.
In order to achieve this, we minimise the expectation of a task loss $\ell(\vec{h}, \set{M})$ which measures how well the resulting cuboid $\vec{h}$ and previously determined cuboids $\set{M}$ fit to a scene:
\vspace{-2mm}
\begin{equation}
    \mathcal{L}(\vec{w}) = \mathbb{E}_{\set{H} \sim p(\set{H} | \set{M}; \vec{w})} \left[ \ell (\vec{\hat{h}}, {\set{{M}}}) \right] \, ,
    \label{eq:exp_task_loss_w}
\end{equation}
with $\vec{\hat{h}} \in \set{H}$ being the cuboid hypothesis selected according to the inlier criterion:
\vspace{-1mm}
\begin{equation}
    \vec{\hat{h}} \in \argmax_{\vec{h} \in \set{H}} I_{\text{c}}(\set{Y}, \set{M} \cup \{\vec{h}\}) \, .
\end{equation}
However, as described by Brachmann et al.~\cite{Brachmann2017:DSAC, Brachmann2019:NGRANSAC}, this discrete hypothesis selection prohibits also learning parameters $\vec{v}$ of the feature extraction network.
We must therefore turn hypothesis selection into a probabilistic action:
\vspace{-2mm}
\begin{equation}
    \vec{\hat{h}} \sim p(\vec{\hat{h}} | \set{H}, \set{M}) = \frac{\exp I_{\text{c}}(\set{Y}, \set{M} \cup \{\vec{\hat{h}}\}) } { \sum_{\vec{{h}} \in \set{H} } \exp I_{\text{c}}(\set{Y}, \set{M} \cup \{\vec{{h}}\}) } \, .
\end{equation}
This allows us to compute the expected loss over cuboid hypotheses $\set{H}$, which is differentiable:
\vspace{-1mm}
\begin{equation}
    \mathcal{L}(\vec{v}) = \mathbb{E}_{\vec{h} \sim p(\vec{h} | \set{H}, \set{M}) } \left[ \ell (\vec{h}, \set{M}) \right] \, .
    \label{eq:exp_task_loss_v}
\end{equation}
Combining Eq.~\ref{eq:exp_task_loss_w} and Eq.~\ref{eq:exp_task_loss_v}, we can train both networks together end-to-end:
\vspace{-1mm}
\begin{equation}
    \mathcal{L}(\vec{v}, \vec{w}) = \mathbb{E}_{\set{H} \sim p(\set{H} | \set{M}; \vec{w})} \mathbb{E}_{\vec{h} \sim p(\vec{h} | \set{H}, \set{M}) } \left[ \ell (\vec{h}, \set{M}) \right] \, .
    \label{eq:exp_task_loss_vw}
\end{equation}
Computing the exact expectation in Eq.~\ref{eq:exp_task_loss_w} is intractable, so we approximate its gradient  by drawing $K$ samples of $\set{H}$: %
\vspace{-2mm}
\begin{align}
    \frac{\partial\mathcal{L}(\vec{v},\vec{w})}{\partial (\vec{v},\vec{w})}   \approx 
     \frac{1}{K} \sum_{k=1}^K \left[ 
    \mathbb{E}_{\vec{h}} \left[ \ell \right] \frac{\partial \log p(\set{H}_k; \vec{w})}{\partial (\vec{v},\vec{w})}  +
    \frac{\partial \mathbb{E}_{\vec{h}}[\ell]]}{\partial (\vec{v},\vec{w})} 
    \right] \, . 
    \label{eq:exp_task_loss_grad_approx}
\end{align}

\vspace{-7mm}
\paragraph{Differentiable Solver.}
The above requires that the gradient for a particular 3D feature can be computed \wrt a resulting cuboid. 
However, cuboid parameters are computed via iterative numerical optimisation (cf. Sec.~\ref{subsubsec:fitting}). 
Tracking the operations through this step results in inaccurate gradients and prohibitively high computational costs. 
In the following, we describe a solution which does not necessitate tracking the computations.
Instead, it relies on the implicit function theorem to directly compute the desired gradients. 

Given a set of features $\set{S} = \{\vec{y}_1, \dots, \vec{y}_C\}$ and 
an optimally fit cuboid $\vec{h}$, we seek to compute the partial derivatives ${\partial\vec{h}}/{\partial\set{S}}$.
From Eq.~\ref{eq:def.F}, we have $F(\set{S}, \vec{h}) = \vec{0}$. Via the implicit function theorem, we can therefore obtain: 
\vspace{-2mm}
\begin{equation}
   \frac{\partial\vec{h}}{\partial\set{S}} = - {\left(\frac{\partial F}{\partial\vec{h}}(\set{S}, \vec{h})\right)}\inverse \cdot \frac{\partial F}{\partial\set{S}}(\set{S}, \vec{h}) \, .
 \end{equation}
In practice, the partial derivatives by the size parameters $\partial F / \partial a_{\{x,y,z\}}$ are mostly zero or close to zero.
Inversion of the Jacobian $\frac{\partial F}{\partial\vec{h}}(\set{S}, \vec{h})$ is not possible or numerically unstable in this case. 
We thus approximate it by masking out the possibly zero derivatives and using the pseudo-inverse:
\begin{equation}
    \frac{\partial\vec{h}}{\partial\set{S}} \approx - \left( \frac{\partial F}{\partial(\vec{R}, \vec{t})}(\set{S}, \vec{h})\right)\pinv \cdot \frac{\partial F}{\partial\set{S}}(\set{S}, \vec{h}) \, .
\end{equation}
This allows us to perform end-to-end backpropagation even without a differentiable minimal solver $\vec{h} = f_h(\set{S})$.

\vspace{-2mm}
\paragraph{Task Loss.}
During training, we aim to maximise the expected inlier counts of the sampled cuboids, \ie we minimise the following loss:
\vspace{-2mm}
\begin{equation}
    \ell(\vec{{h}}, {\set{{M}}}) = -I_{\text{c}}(\set{Y}, \set{M} \cup \{\vec{h}\})
\end{equation}
In order enable gradient computation according to Eq.~\ref{eq:exp_task_loss_grad_approx}, $\ell$ must also be differentiable.
Instead of a hard threshold, we thus use a soft inlier measure derived from~\cite{KluBra2020}:
\vspace{-2mm}
\begin{equation}
    f_{\text{I}}(\vec{y}, \vec{h}, i) = 1 - \sigma(\beta (\frac{1}{\tau}  d_i^{\text{P}}(\vec{y}, \vec{h})^2 - 1)) \, ,
\end{equation}
with softness parameter $\beta$, inlier threshold $\tau$, and $\sigma(\cdot)$ being the sigmoid function. 
This loss is based on geometrical consistency only, and does not need any additional labels.

\vspace{-2mm}
\paragraph{Regularisation.}
In order to prevent mode collapse of the sampling weights $\set{Q}$ (cf. Sec.~\ref{par:sampling}), we apply a regularisation term during training.
We minimise the correlation coefficients between individual sets of sampling weights $\vec{p} \in \set{Q}$:
\vspace{-2mm}
\begin{equation}
    \ell_{\text{corr}}(\set{Q}) = \sum_{\substack{\vec{p}_i, \vec{p}_j \in \set{Q},\\  i \neq j}} \frac{\operatorname{cov}(\vec{p}_i, \vec{p}_j)}{\sigma_{\vec{p}_i}, \sigma_{\vec{p}_j}} \, ,
\end{equation}
with covariances $\operatorname{cov}(\cdot,\cdot)$ and standard deviations $\sigma$. 
For the same reason, we also maximise the entropy of selection probabilities $\vec{q}$, \ie $\ell_{\text{entropy}} = -\operatorname{H}(\vec{q})$.

\section{Experiments}
\label{sec:exp}
\paragraph{Dataset.}
We provide qualitative and quantitative results on the NYU Depth v2 dataset~\cite{silberman2012indoor}. 
It contains 1449 images of indoor scenes with corresponding ground truth depth recorded with a Kinect camera.
We use the same dataset split as~\cite{lee2019big}: 654 images for testing and 795 images for training, of which we reserved 195 for validation.

\vspace{.5mm}
\noindent\textbf{Implementation Details. }
For experiments with RGB image input, we use the BTS~\cite{lee2019big} depth estimator pre-trained on NYU as our feature extraction network. 
BTS achieves state-of-the-art results on NYU with a publicly available implementation\footnote{https://github.com/cogaplex-bts/bts}, and is thus a sensible choice.
We implemented the sample weight estimator as a variant of the fully convolutional neural network used in~\cite{Brachmann2019:NGRANSAC}. 
We pretrain the sample weight estimation network with ground truth depth as input for $20$ epochs. 
This is also the network we use for experiments with depth input.
We then continue training both networks end-to-end with RGB input for another $25$ epochs.
For cuboid fitting (cf. Sec.~\ref{subsubsec:fitting}), we implement the minimal solver by applying the Adam~\cite{KingmaB14} optimiser to perform gradient descent \wrt cuboid parameters for 50 iterations.
We provide additional implementation details and a listing of hyperparameters in the supplementary. 

\begin{table*}
\setlength\tabcolsep{.37em}
	\begin{center}
	\begin{tabular}{|l|cc|cc|cc|cc|cc|cc|}
	\cline{2-13}
	\multicolumn{1}{c|}{} & \multicolumn{10}{c|}{occlusion-aware $L_2$-distance}                 & \multicolumn{2}{c|}{$L_2$} \\
	\multicolumn{1}{c|}{} & \multicolumn{2}{c|}{AUC$_{@\SI{50}{\centi\meter}}$} & \multicolumn{2}{c|}{AUC$_{@\SI{20}{\centi\meter}}$} & \multicolumn{2}{c|}{AUC$_{@\SI{10}{\centi\meter}}$} & \multicolumn{2}{c|}{AUC$_{@\SI{5}{\centi\meter}}$}    & \multicolumn{2}{c|}{mean (cm)}      & \multicolumn{2}{c|}{mean (cm)} \\
	\hline
	\multicolumn{13}{|c|}{\small{RGB input}} \\
	\textbf{Ours}                                                           & $\mathbf{57.0}\%$ & $\scriptstyle \pm 0.20$  
	                                                                        & $\mathbf{33.1}\%$ & $\scriptstyle \pm 0.17$  
	                                                                        & $\mathbf{18.9}\%$ & $\scriptstyle \pm 0.09$  
	                                                                        & $\mathbf{10.0}\%$ & $\scriptstyle \pm 0.06$  
	                                                                        & $\mathbf{34.5}$ &   $\scriptstyle \pm 0.24$  
	                                                                        & $\mathbf{30.1}$ &   $\scriptstyle \pm 0.34$ \\
	SQ-Parsing~\cite{paschalidou2019superquadrics} + BTS~\cite{lee2019big}  & $30.1\%$ &  & $11.6\%$ &  & $4.3\%$ &  & $1.1\%$ &  & $65.9$ &  & $33.9$ &   \\
	SQ-Parsing (RGB)~\cite{paschalidou2019superquadrics}                    & $19.4\%$ &  & $6.3\%$ &  & $1.9\%$ &  & $0.4\%$ &  & $87.4$ &  & $42.4$ &   \\
	\hline
	\multicolumn{13}{|c|}{\small{depth input}} \\
	\textbf{Ours}                                                           & $\mathbf{77.2}\%$ & $\scriptstyle \pm 0.08$  
	                                                                        & $\mathbf{62.7}\%$ & $\scriptstyle \pm 0.07$  
	                                                                        & $\mathbf{49.1}\%$ & $\scriptstyle \pm 0.12$  
	                                                                        & $\mathbf{34.3}\%$ & $\scriptstyle \pm 0.14$  
	                                                                        & $\mathbf{20.8}$ &   $\scriptstyle \pm 0.51$  
	                                                                        & $\mathbf{17.9}$ &   $\scriptstyle \pm 0.50$ \\
	Sequential RANSAC~\cite{vincent2001seqransac}                           & ${59.9}\%$ & $\scriptstyle \pm 0.17$  
	                                                                        & ${42.8}\%$ & $\scriptstyle \pm 0.09$  
	                                                                        & ${30.1}\%$ & $\scriptstyle \pm 0.84$  
	                                                                        & ${18.8}\%$ & $\scriptstyle \pm 0.10$  
	                                                                        & ${37.9}$ &   $\scriptstyle \pm 0.74$  
	                                                                        & $32.7$ &   $\scriptstyle \pm 0.72$ \\
	SQ-Parsing~\cite{paschalidou2019superquadrics}                          & $50.3\%$ &  & $22.4\%$ &  & $8.8\%$ &  & $2.5\%$ &  & $34.7$ &  & ${20.9}$ &   \\
	\hline
	\end{tabular}
    \end{center}
 \vspace{-4mm}
	\caption{\textbf{Quantitative Results on NYU:} We evaluate on NYU Depth v2~\cite{silberman2012indoor}, for both RGB and depth inputs. We compare our method against variants of SQ-Parsing~\cite{paschalidou2019superquadrics} and Sequential RANSAC~\cite{vincent2001seqransac}. We present AUC values (higher is better) for various upper bounds of the occlusion-aware (OA) $L_2$ distance (cf. Sec.~\ref{subsubsec:oa_point_cuboid_distance}). We also report mean OA-$L_2$ and regular $L_2$ distances (lower is better). See Sec.~\ref{sec:exp} for details.}
	\label{tab:main_results}
\vspace{-1.1em}
\end{table*}

\vspace{.5mm}
\noindent\textbf{Baselines. }
While multiple works in the field of primitive based 3D shape parsing have been published in recent years, not all of them can be used for comparison. 
We cannot use methods which require ground truth shape annotations~\cite{niu2018im2struct, zou20173d} or watertight meshes~\cite{genova2019learning, deng2020cvxnet} for training, as the data we are concerned with does not provide these.
Unfortunately, we also cannot compare against~\cite{Paschalidou2020:Decomposition}, as they also evaluated on different data and have not yet provided their source code.
For the cuboid based approach of~\cite{tulsiani2017learning}, source code is available\footnote{https://github.com/shubhtuls/volumetricPrimitives}. %
However, following the instructions they provided, we were unable to obtain sensible results on NYU.
The neural network of~\cite{tulsiani2017learning} appears to degenerate when trained on NYU, predicting similar unreasonable cuboid configuration for all scenes (cf.~qualitative results in the supplementary material). 
We therefore did not include it in our evaluation.
On the other hand, we were able to train the superquadric-based approach of~\cite{paschalidou2019superquadrics} (SQ-Parsing) on NYU using their provided source code\footnote{https://github.com/paschalidoud/superquadric\_parsing}.
Since SQ-Parsing is designed for meshes as input, we preprocess the ground truth depth point clouds of NYU by applying Poisson surface reconstruction~\cite{kazhdan2006poisson}.
We also compare against a variant of~\cite{paschalidou2019superquadrics} which directly operates on RGB images (SQ-Parsing RGB).
We further compare our method against a variant of Sequential RANSAC~\cite{vincent2001seqransac} from depth input.
Lastly, we evaluate the results of SQ-Parsing when applied to the prediction of the same depth estimation network (BTS~\cite{lee2019big}) which we employ for our method. 
We used the same parameters as~\cite{paschalidou2019superquadrics} for these experiments.

\vspace{-1.25em}
\paragraph{Metrics.}
Previous works~\cite{paschalidou2019superquadrics, Paschalidou2020:Decomposition} evaluated their results using Chamfer distance and volumetric IoU.
These works, however, deal with full 3D shapes of watertight objects, while we fit 3D primitives to scenes where only a 2.5D ground truth is available.
As we explain in Sec.~\ref{subsubsec:oa_point_cuboid_distance}, we therefore evaluate using the occlusion-aware distance metric, OA-$L_2$ for short, instead.
We calculate the occlusion-aware distances of all ground-truth points to the recovered primitives per scene.
Using these distances, we then compute the relative area under the recall curve (AUC, in percent) for multiple upper bounds: ${\SI{50}{\centi\meter}}$, ${\SI{20}{\centi\meter}}$, ${\SI{10}{\centi\meter}}$, ${\SI{5}{\centi\meter}}$. 
The AUC values are less influenced by outliers than the mean distance, and gauge how many points are covered by the primitives within the upper bound.
In addition, we also report mean OA-$L_2$ as well as the mean of the regular $L_2$ distance.
As our method is random sampling based, we report the mean and variance over five runs for all metrics.

\begin{figure*}	
		\begin{subfigure}[t]{0.99\linewidth}
			\includegraphics[width=0.155\linewidth]{fig/examples/007_0362_0836_original.jpg}
			\includegraphics[width=0.155\linewidth]{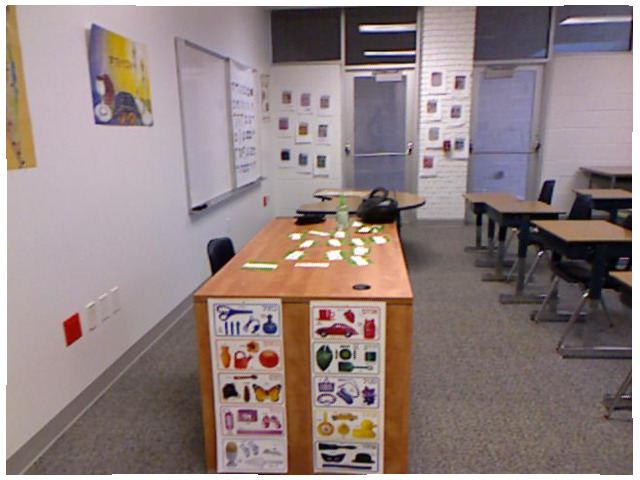}
			\includegraphics[width=0.155\linewidth]{fig/examples/012_0236_0570_original.jpg}
			\includegraphics[width=0.155\linewidth]{fig/examples/006_0603_1347_original.jpg}
			\includegraphics[width=0.155\linewidth]{fig/examples/001_0336_0778_original.jpg}
			\includegraphics[width=0.155\linewidth]{fig/examples/001_0160_0396_original.jpg}
			\vspace{-.6em}
			\caption{Input images}
			\vspace{.1em}
		\end{subfigure}
		\begin{subfigure}[t]{0.99\linewidth}
			\includegraphics[width=0.155\linewidth]{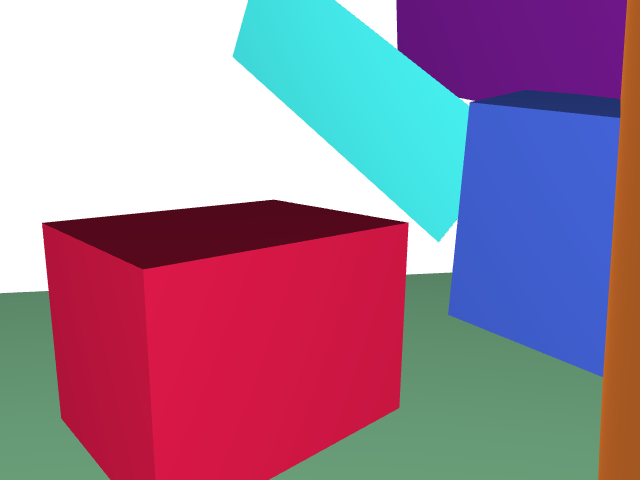}
			\includegraphics[width=0.155\linewidth]{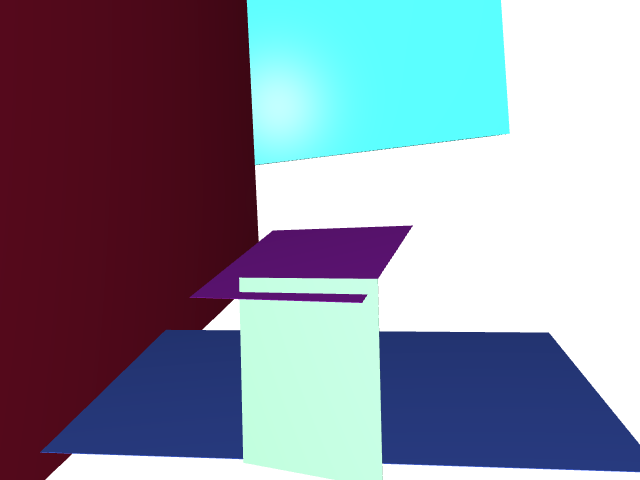}
			\includegraphics[width=0.155\linewidth]{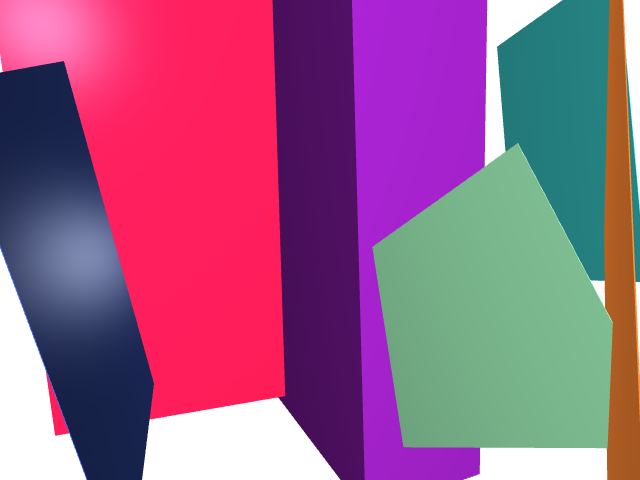}
			\includegraphics[width=0.155\linewidth]{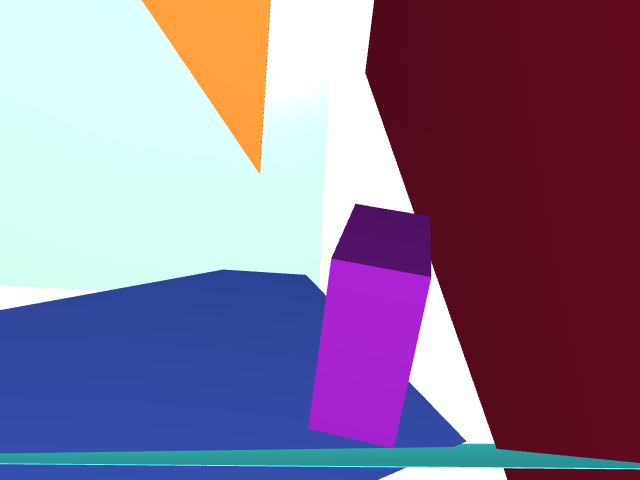}
			\includegraphics[width=0.155\linewidth]{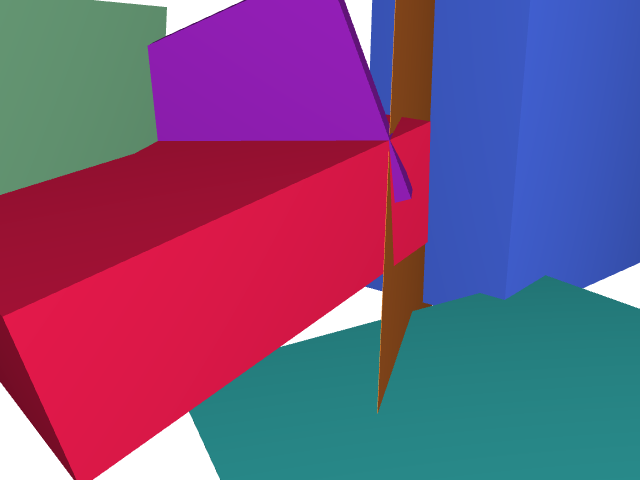}
			\includegraphics[width=0.155\linewidth]{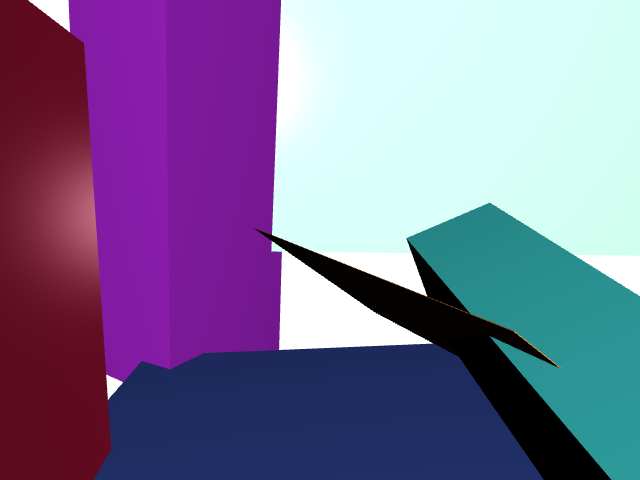}
			\vspace{-.6em}
			\caption{Recovered cuboids, original view}
			\vspace{.1em}
		\end{subfigure}
		\begin{subfigure}[t]{0.99\linewidth}
			\includegraphics[width=0.155\linewidth]{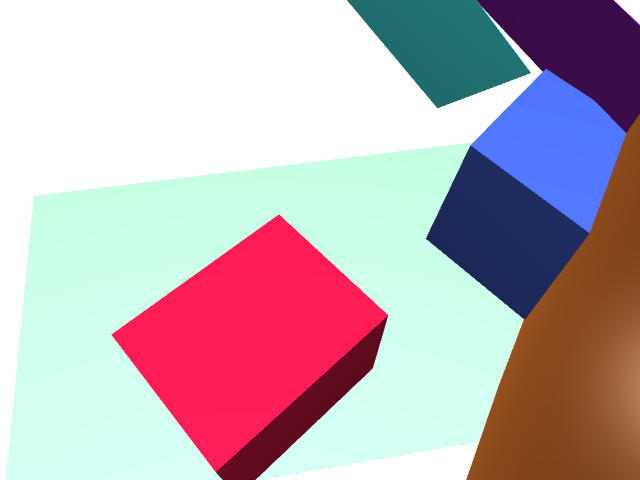}
			\includegraphics[width=0.155\linewidth]{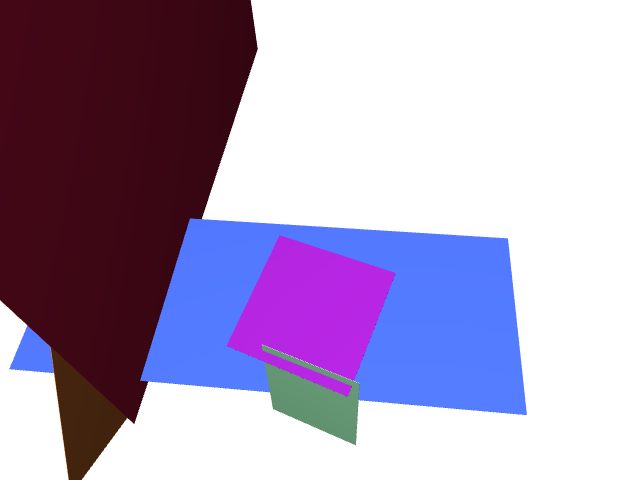}
			\includegraphics[width=0.155\linewidth]{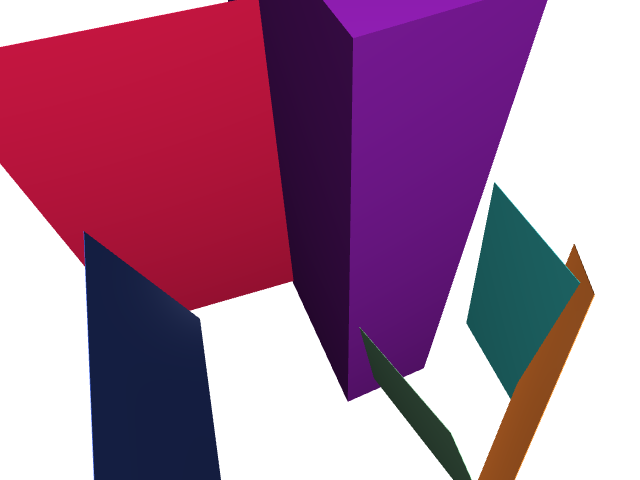}
			\includegraphics[width=0.155\linewidth]{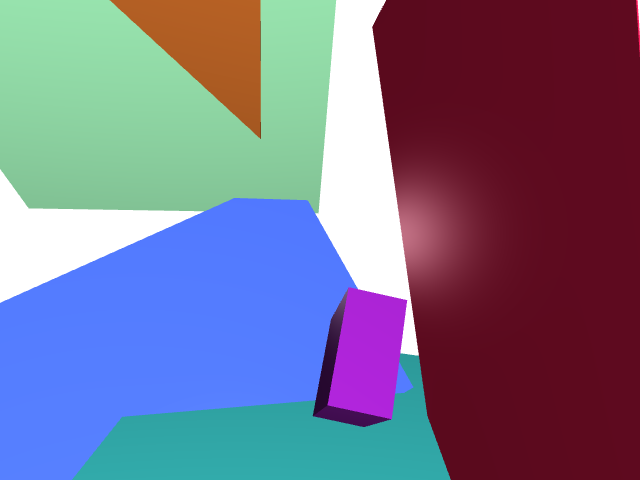}
			\includegraphics[width=0.155\linewidth]{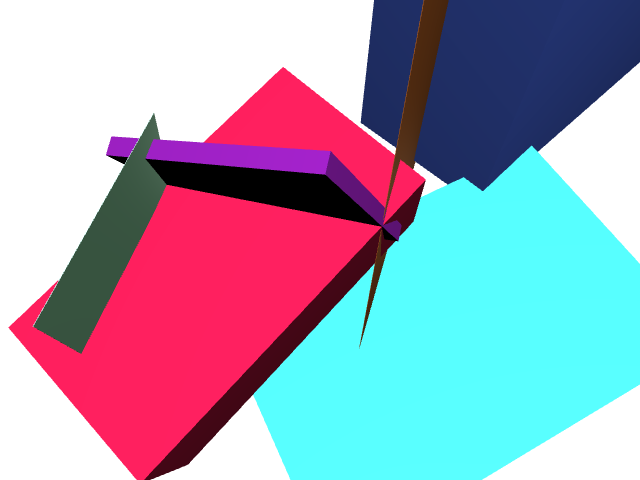}
			\includegraphics[width=0.155\linewidth]{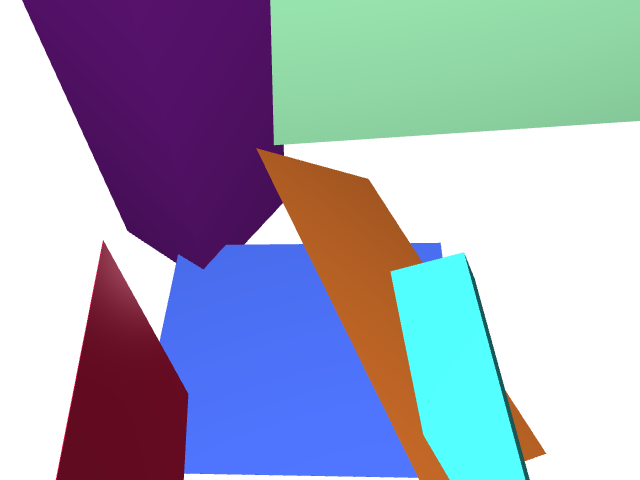}
			\vspace{-.6em}
			\caption{Recovered cuboids, top view}
			\vspace{.1em}
		\end{subfigure}
		\begin{subfigure}[t]{0.99\linewidth}
			\includegraphics[width=0.155\linewidth]{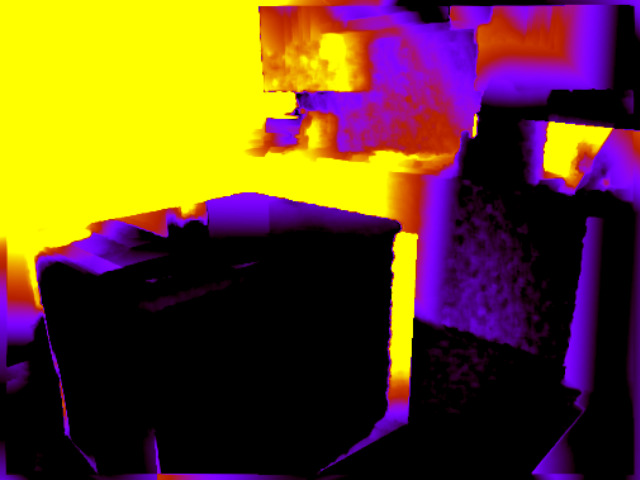}
			\includegraphics[width=0.155\linewidth]{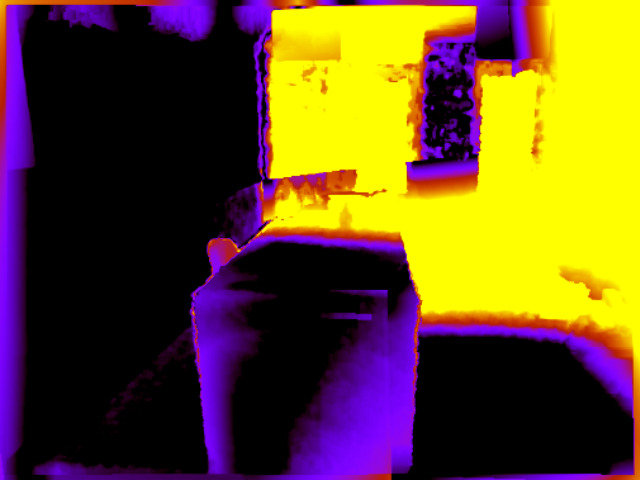}
			\includegraphics[width=0.155\linewidth]{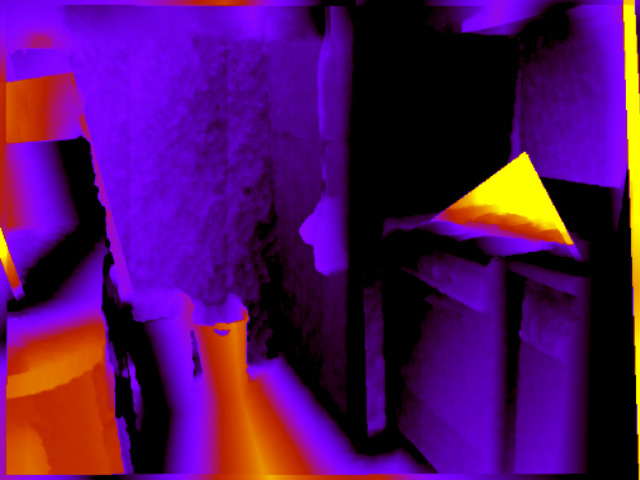}
			\includegraphics[width=0.155\linewidth]{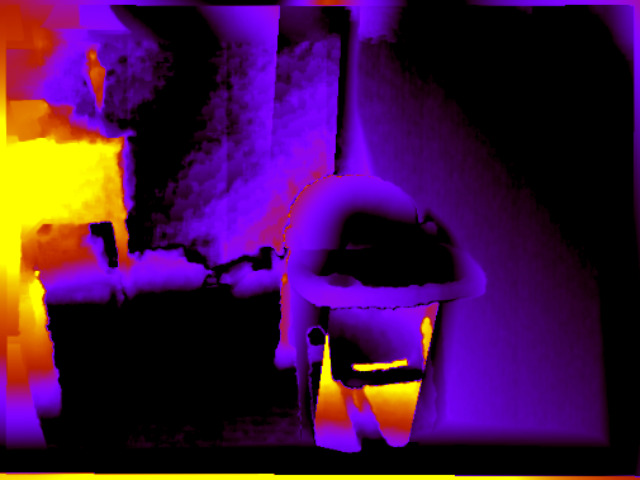}
			\includegraphics[width=0.155\linewidth]{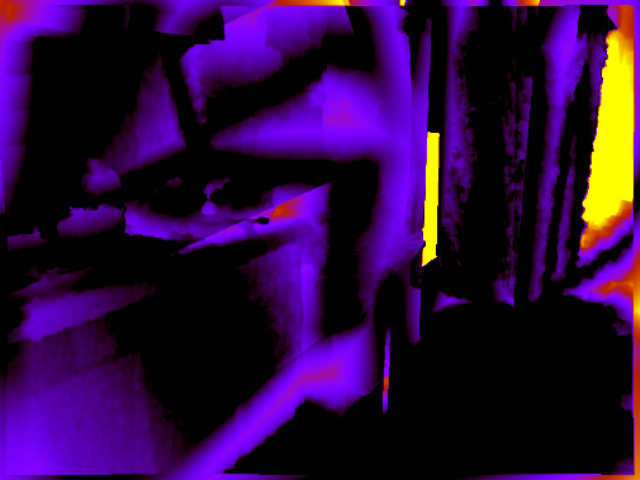}
			\includegraphics[width=0.155\linewidth]{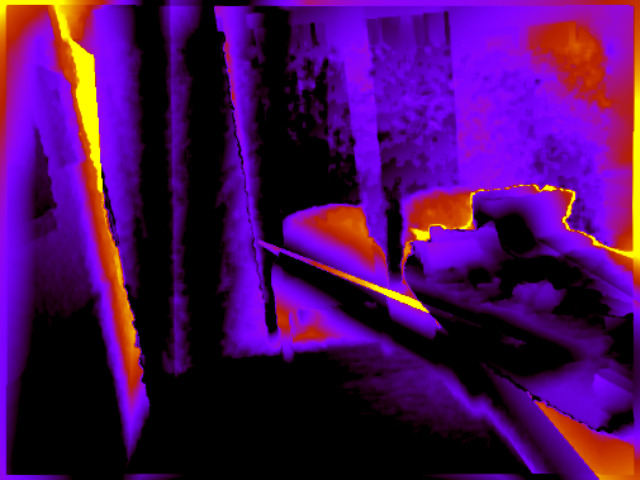}
			\includegraphics[width=0.04\linewidth]{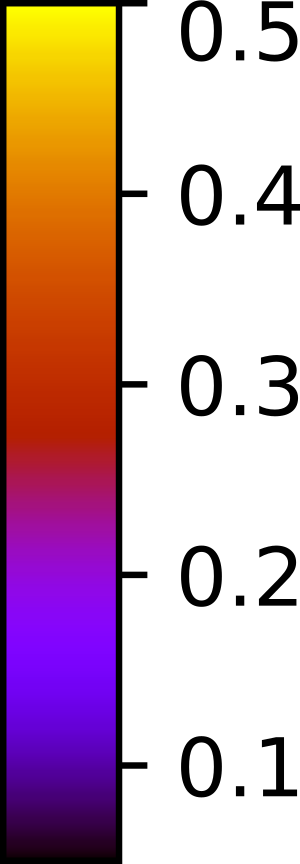}
			\vspace{-1.7em}
			\caption{Visualisation of the occlusion-aware distance (legend on the right, in metres) between the cuboids and the ground-truth depth.}
			\vspace{.1em}
		\end{subfigure}
		\begin{subfigure}[t]{0.99\linewidth}
			\includegraphics[width=0.155\linewidth]{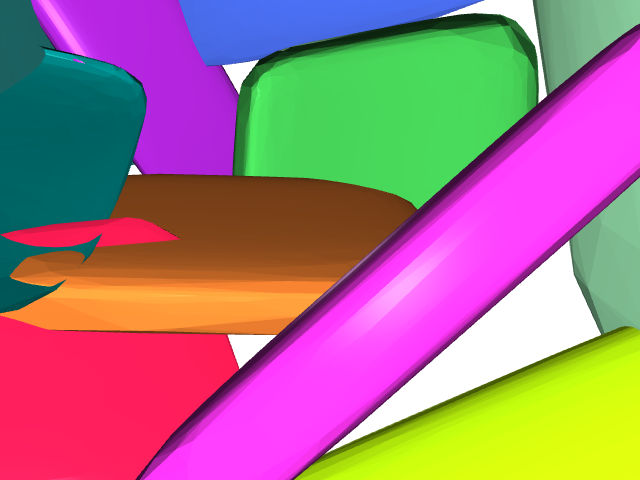}
			\includegraphics[width=0.155\linewidth]{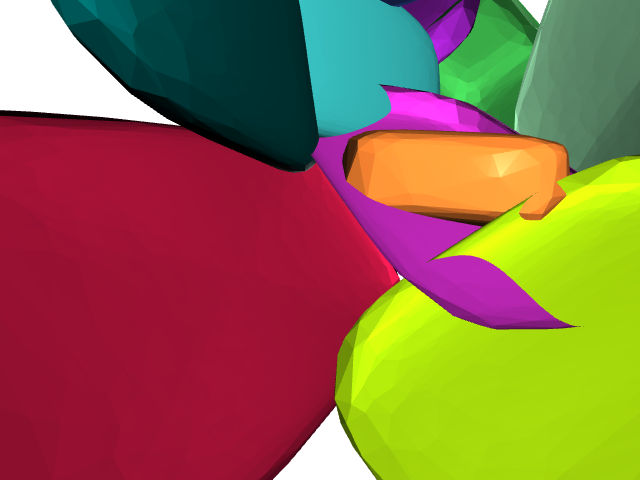}
			\includegraphics[width=0.155\linewidth]{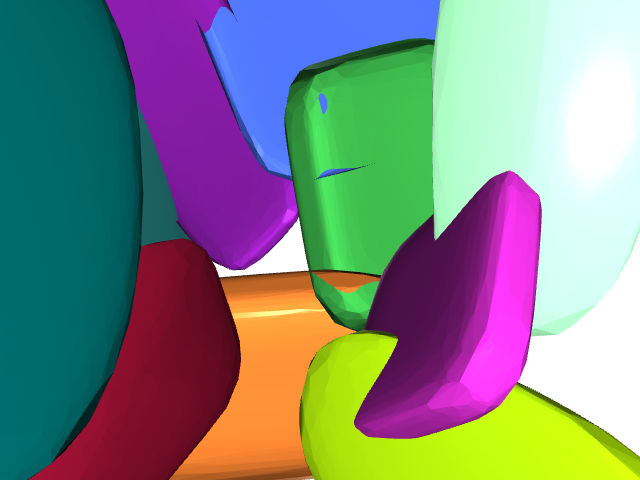}
			\includegraphics[width=0.155\linewidth]{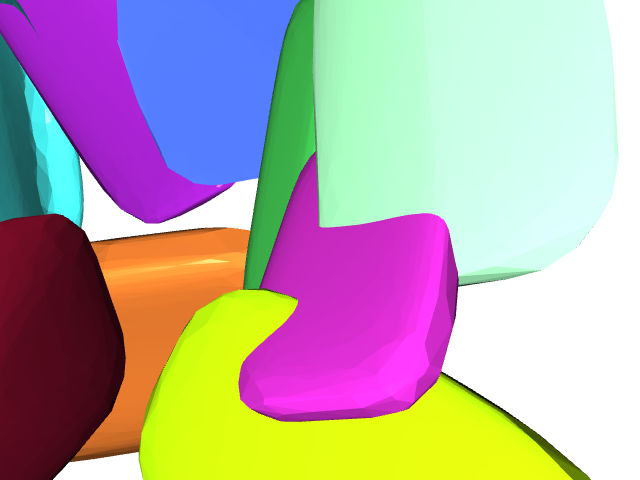}
			\includegraphics[width=0.155\linewidth]{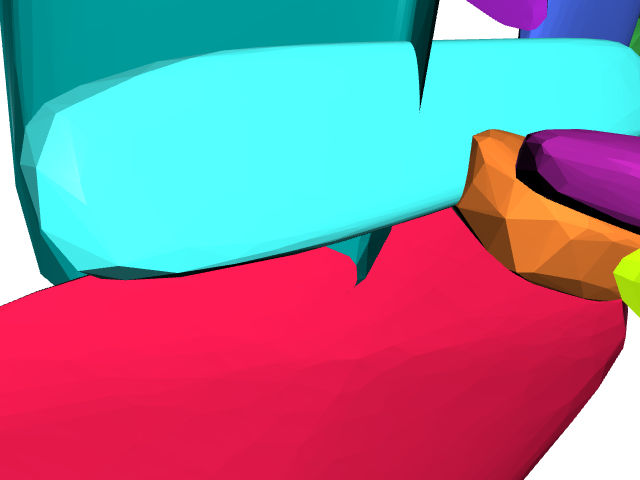}
			\includegraphics[width=0.155\linewidth]{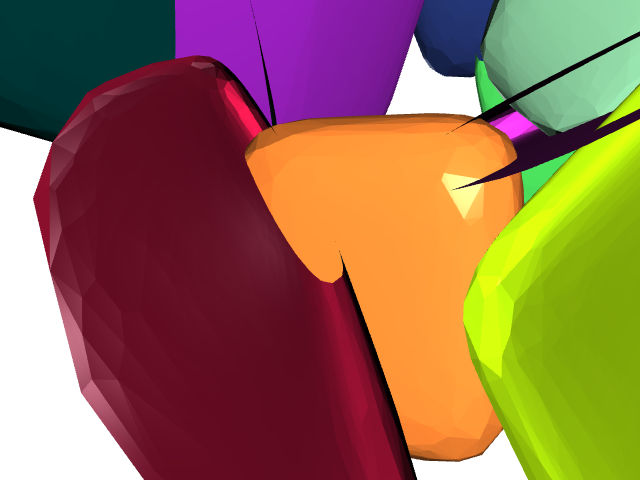}
			\vspace{-.5em}
			\caption{Recovered superquadrics~\cite{paschalidou2019superquadrics}, original view}
			\vspace{.2em}
		\end{subfigure}
\vspace{-10pt}
		\caption{\textbf{Qualitative Results.} First row: Input RGB images. 
		Second and third rows: Cuboids obtained with our proposed method 
		(same views as the original images and top views, respectively). 
		Colours convey the order in which the cuboids have been selected: \textcolor{c_red}{red}, \textcolor{c_blue}{blue}, \textcolor{c_green}{green}, \textcolor{c_purple}{purple}, \textcolor{c_cyan}{cyan}, \textcolor{c_orange}{orange}.
		Cuboids covering large parts of the image are often selected early.
		Fourth row: OA-$L_2$ distance (Sec.~\ref{subsubsec:oa_point_cuboid_distance}) between the cuboids and the ground-truth depth. Very large distances are due to missing or occluding cuboids.
		Last row: Superquadrics obtained using~\cite{paschalidou2019superquadrics}. These abstractions are very cluttered and hardly represent the original scenes. 
		Our algorithm infers abstractions that represent the original scenes more closely.}
		\label{fig:examples}
\vspace{-4mm}
\end{figure*}

\vspace{-1.25em}
\paragraph{Qualitative Results.}
We present qualitative examples from the NYU dataset comparing the results of~\cite{paschalidou2019superquadrics} against our method.
In Fig.~\ref{fig:examples}, rows 3-5, we show renderings of the cuboids predicted by by our method (RGB input in row 1) for scenes from the NYU test set.
Row three is rendered from the same perspective as the original image, while row four and five show top and side views, respectively.
As these examples show, our method is able to recover cuboids for key elements of the scene, such as walls, floors, cupboards, counters or tables.
It often recovers cuboids covering larger parts of the image first.
It does not always succeed in capturing volumetric properties of all objects, such as the fridge in column five, which is represented by two almost planar cuboids instead.
In addition, spurious, usually very thin cuboids appear occasionally, such as the cuboid intersecting the table in the last column.
For comparison, we show renderings of superquadrics obtained with SQ-Parsing + BTS -- which is our best performing competitor for RGB input -- in row 2.
Unlike our cuboid representations, these superquadrics bear little resemblance to the original scene.
Our method is thus able to abstract the depicted indoor scenes in a much more sensible way than~\cite{paschalidou2019superquadrics}.

\begin{table}
\vspace{-1.25em}
	\begin{center}
	\begin{tabular}{|l|c|c|c|c|}
	\cline{2-5}
	\multicolumn{1}{c|}{} & \multicolumn{4}{c|}{OA-$L_2$} \\
	\cline{2-5}
	\multicolumn{1}{c|}{} & \small{AUC$_{@\SI{20}{}}$} & \small{AUC$_{@\SI{10}{}}$} & \small{AUC$_{@\SI{5}{}}$} &  mean (cm)\\
	\hline
	$Q=1$ & $40.2\%$ & $29.7\%$ & $19.8\%$ & $45.3$ \\
	$Q=2$ & $57.5\%$ & $43.6\%$ & $29.7\%$ & $22.3$ \\
	$Q=4$ & $\mathbf{62.7}\%$ & $\mathbf{49.1}\%$ & $\mathbf{34.3}\%$ & $\mathbf{20.8}$ \\
	    \hline

	\end{tabular}
    \end{center}
\vspace{-4mm}
\caption{\textbf{Ablation Study:} We evaluate our approach with varying numbers $Q$ of sampling weight sets with depth input.
We present AUC values for various upper bounds of the OA-$L_2$ distance and the mean OA-$L_2$ distance.
	}
	\label{tab:ablation2_results}
\vspace{-1.2em}
\end{table}

\vspace{-1em}
\paragraph{Quantitative Evaluation.}
Table~\ref{tab:main_results} gives quantitative results on the NYU Depth v2~\cite{silberman2012indoor} test set for both RGB and depth inputs.
For depth input, our method achieves significantly higher AUC values than
Sequential RANSAC~\cite{vincent2001seqransac}, with margins between $15.5$ and $19.9$ percentage points. 
Sequential RANSAC outperforms SQ-Parsing on AUC, yet is inferior w.r.t.~the means.
Our method also achieves significantly higher AUC values than SQ-Parsing~\cite{paschalidou2019superquadrics} across the whole range, with margins between $26.9$ and $40.3$ percentage points.
We present lower mean $L_2$ and occlusion-aware $L_2$ distances as well, with improvements of $\SI{3.0}{\centi\meter}$ ($14.3\%$) and $\SI{13.9}{\centi\meter}$ ($40.0\%$) respectively. 
For RGB input, Tab.~\ref{tab:main_results} shows that SQ-Parsing + BTS, \ie the combination of~\cite{paschalidou2019superquadrics} for depth input with the monocular depth estimator from~\cite{lee2019big}, performs better than 
the RGB variant of~\cite{paschalidou2019superquadrics}.
This implies that the image based encoder network of~\cite{paschalidou2019superquadrics} is not able to extract 3D information from images as well as~\cite{lee2019big} can.
Our approach, however, also outperforms SQ-Parsing + BTS by large margins, with AUC improvements of $8.9$ to $26.9$ percentage points.
We improve mean $L_2$ and OA-$L_2$ distances by $\SI{4.0}{\centi\meter}$ ($11.8\%$) and $\SI{31.5}{\centi\meter}$ ($47.8\%$) respectively. 
In summary, our method outperforms~\cite{paschalidou2019superquadrics} on NYU in all settings on all metrics by significant margins, presenting a new state-of-the-art for this challenging task.

\vspace{-3.5mm}
\paragraph{Ablation Study.}
\label{par:ablation}
To demonstrate the efficacy of predicting multiple sets of sampling weights at once (cf. Sec.~\ref{par:sampling}), we trained our approach with varying numbers $Q$ of sampling weight sets.
We evaluated these variants with depth input and present the results in Tab.~\ref{tab:ablation2_results}.
Predicting multiple sampling weight sets ($Q=4$) performs significantly better than predicting just one ($Q=1$) or two ($Q=2$) sets.
Please refer to the supplementary for an ablation study \wrt~our proposed occlusion-aware inlier counting.

\vspace{-.9em}
\section{Conclusion}
\vspace{-0.6em}
We present a 3D scene parser which abstracts complex real-world scenes into ensembles of simpler volumetric primitives.
It builds upon a learning-based robust estimator, which we extend in order to recover cuboids from RGB images.
To this end, we propose an occlusion-aware distance metric which enables us to correctly handle opaque scenes.
We facilitate end-to-end training by circumventing backpropagation through our minimal solver and deriving the gradient of primitive parameters \wrt the input features analytically.
Our algorithm neither requires known ground truth primitive parameters nor any other costly annotations. 
It can thus be straight-forwardly applied to other datasets which lack this information.
Results on the challenging real-world NYU Depth v2 dataset demonstrate that the proposed method successfully parses and abstracts complex and cluttered 3D scenes.
In future work, we plan to address common failure cases, \ie planar and spurious cuboids, by replacing the depth estimation with more expressive 3D features and using matching data for training.

\vspace{-.9em}
\paragraph{Acknowledgements. }
This work was supported by the BMBF grant \emph{LeibnizAILab} (01DD20003), by the DFG grant \emph{COVMAP} (RO 2497/12-2), by the DFG Cluster of Excellence \emph{PhoenixD} (EXC 2122), and by the Center for Digital Innovations (ZDIN).

\renewcommand*\appendixpagename{\Large Appendix}

\begin{appendices}

We provide additional implementation details for our method, including a listing of hyperparameters (Tab.~\ref{tab:parameters}), in Sec.~\ref{sec:implementation} of this appendix. 
In Sec.~\ref{sec:oasq}, we establish the occlusion-aware distance metric for superquadrics, which we need for our quantitative evaluation. 
Sec.~\ref{sec:ablation} provides an additional ablation study analysing the effectiveness of our proposed occlusion-aware inlier counting (cf. Sec. 3.2.2 in the main paper). 
In Sec.~\ref{sec:qualitative}, we show additional qualitative results, including failure cases of our method.

\section{Implementation Details}
\label{sec:implementation}

\subsection{Feature Extraction Network}
\vspace{-.3em}
We employ the \emph{Big-to-Small} (BTS)~\cite{lee2019big} depth estimation CNN as our feature extraction network.
We use a variant of their approach using a DenseNet-161~\cite{huang2017densely} as the base network.
It is pre-trained on NYU Depth v2~\cite{silberman2012indoor} and achieves state-of-the-art results for monocular depth estimation.
Please refer to~\cite{lee2019big} for details.

\subsection{Sampling Weight Network}
\begin{figure*}	
\centering
\includegraphics[width=0.99\linewidth]{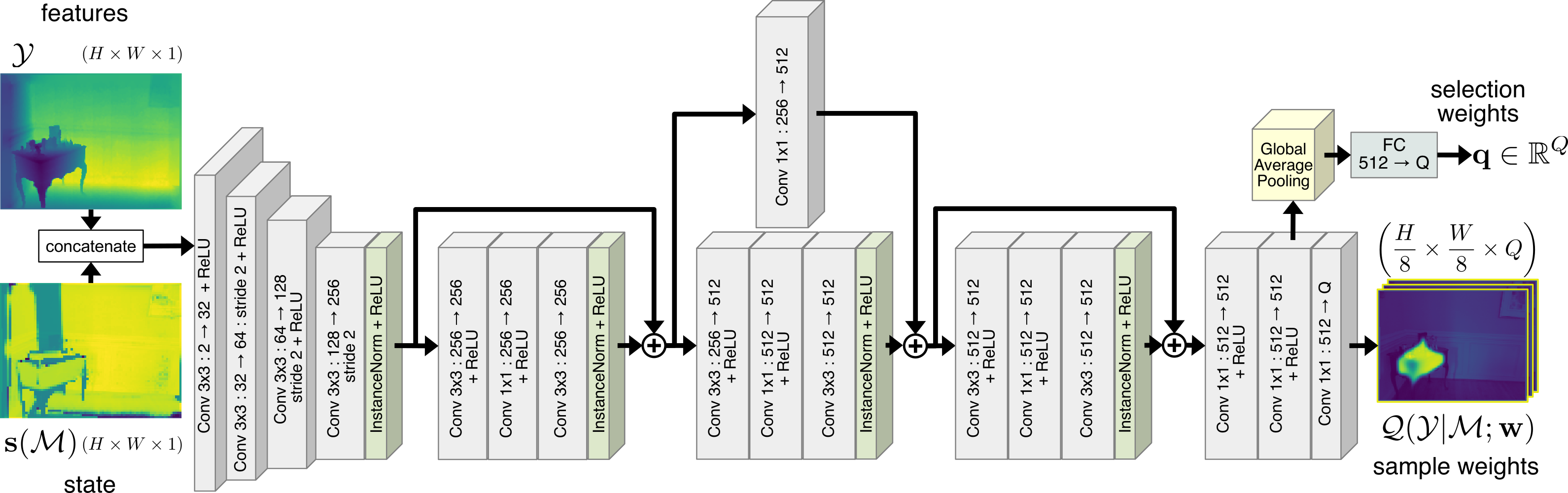}
\vspace{-5pt}
\caption{ \textbf{Sampling Weight Network Architecture:}
We stack features $\set{Y}$ and state $\vec{s}$ into a tensor of size $H \times W \times 2$, with image width $W$ and height $H$. We feed this tensor into a neural network consisting of $3\times3$ and $1\times1$ convolutional layers, instance normalisation layers~\cite{ulyanov2016instance}, and ReLU activations~\cite{he2015delving} arranged as residual blocks~\cite{he2016deep}. The last convolutional layer predicts the sampling weight sets $\set{Q}(\set{Y}|\set{M}; \vec{w})$. We apply global average pooling and a fully-connected layer to the output of the penultimate convolutional layer in order to predict the selection weights $\vec{q}$. This architecture is based on~\cite{KluBra2020, Brachmann2019:NGRANSAC}.
}
\vspace{-5pt}
\label{fig:sampling_net}
\end{figure*}

\vspace{-.3em}
For prediction of sampling weights, we use a neural network based on the architecture for scene coordinate regression used in~\cite{Brachmann2019:NGRANSAC}. 
Refer to Fig.~\ref{fig:sampling_net} for an overview.
The input of the network is a concatenation of features $\set{Y}$ and state $\vec{s}$ of size $H \times W \times 2$, with image width $W$ and height $H$.
When using ground truth depth as input, we normalise $\set{Y}$ by the mean and variance of the training set.
When using the features predicted by the feature extraction network, we add a batch normalisation~\cite{ioffe2015batch} layer between the two networks in order to take care of input normalisation.
We augmented the network of~\cite{Brachmann2019:NGRANSAC} with instance normalisation~\cite{ulyanov2016instance} layers, which proved crucial for the ability of the network to segment distinct structures in the scene.
The network thus consists of $3\times3$ and $1\times1$ convolutional layers, instance normalisation layers~\cite{ulyanov2016instance}, and ReLU activations~\cite{he2015delving} arranged as residual blocks~\cite{he2016deep}.
While most convolutions are applied with stride one, layers two to four use a stride of two, resulting in a final output spatially subsampled by a factor of eight \wrt the input.
We apply sigmoid activation to the last convolutional layer to predict the sampling weight sets $\set{Q}(\set{Y}|\set{M}; \vec{w})$ with size $\frac{H}{8} \times \frac{W}{8} \times Q$, \ie $Q$ sets of sampling weights for each input.
Additionally, we apply global average pooling and a fully-connected layer with sigmoid activation to the output of the penultimate convolutional layer in order to predict the selection weights $\vec{q}$.

\begin{table}[t]
\setlength\tabcolsep{.30em}
    \centering
    \begin{tabular}{c|lc|c|c|}
        \multicolumn{3}{c|}{}                           & depth & RGB \\
        \multicolumn{3}{c|}{}                           & input & input \\
        \hline
        \parbox[t]{2.5mm}{\multirow{9}{*}{\rotatebox[origin=c]{90}{training}}}
        &  learning rate            &               & $10^{-5}$ & $10^{-9}$ \\
        &  epochs                   &               & $20$ & $25$ \\
        &  number of instances      & $|\set{M}|$           & $6$ & $4$  \\
        &  batch size               & $B$           & \multicolumn{2}{c|}{$2$} \\
        &  IMR weight               & $\kappa_{\text{im}}$      & \multicolumn{2}{c|}{$10^{-2}$} \\
        &  correlation weight       & $\kappa_{\text{corr}}$      & \multicolumn{2}{c|}{$1.0$} \\
        &  entropy weight           & $\kappa_{\text{entropy}}$      & \multicolumn{2}{c|}{$1.0$} \\
        &  single-instance samples  & $|\set{H}|$   & \multicolumn{2}{c|}{$32$} \\
        &  sample count             & $K$           & \multicolumn{2}{c|}{$2$}  \\
        \hline
        \parbox[t]{2.5mm}{\multirow{4}{*}{\rotatebox[origin=c]{90}{both}}}
         &  inlier threshold         & $\tau$        & \multicolumn{2}{c|}{$0.004$}  \\
         &  sampling weight sets     & $Q$           & \multicolumn{2}{c|}{$4$}  \\
         &  $f_h$ iterations         &               & \multicolumn{2}{c|}{$50$}  \\
         &  $f_h$ learning rate      &               & \multicolumn{2}{c|}{$0.2$}  \\
        \hline
        \parbox[t]{2.5mm}{\multirow{3}{*}{\rotatebox[origin=c]{90}{test}}}
        &  number of instances      & $|\set{M}|$           & \multicolumn{2}{c|}{$6$}  \\
        &  single-instance samples  &$|\set{H}|$    & \multicolumn{2}{c|}{$4096$}  \\
        &  inlier cutoff (selection)& $\Theta$      & \multicolumn{2}{c|}{$10$}  \\
        
    \end{tabular}
    \caption{\textbf{User definable parameters} of our approach and the values we chose for our experiments using either ground truth depth or RGB images as input. We distinguish between values used during training and at test time. Mathematical symbols refer to the notation used either in the main paper or in this supplementary document.  }
    \label{tab:parameters}
    \vspace{-.75em}
\end{table}

\subsection{Neural Network Training}
\vspace{-.3em}
We implement our method using PyTorch~\cite{paszke2017automatic} version 1.7.0. 
We use the Adam~\cite{KingmaB14} optimiser to train the neural networks.
In order to avoid divergence induced by bad hypothesis samples frequently occurring at the beginning of training, we clamp losses to an absolute maximum value of $0.3$.
First, we train the sampling weight network by itself using ground truth depth for $20$ epochs with a learning rate of $10^{-5}$.
Then we fine-tune the sampling weight network and the feature extraction network with RGB input together for $25$ epochs with a learning rate of $10^{-6}$.
We use a batch size $B=2$ for all experiments.
Training was performed using two RTX 2080 Ti GPUs.

\paragraph{Regularisation.}
In addition to the main task loss $\ell(\vec{h}, \set{M})$, we apply the regularisation losses $\ell_{\text{corr}}$ and $\ell_{\text{entropy}}$ in order to prevent mode collapse of the sampling weight sets $\set{Q}$ (cf. Sec. 3.5 in the main paper). 
We furthermore inherit the inlier masking regularisation (IMR) loss $\ell_{\text{im}}$ from~\cite{KluBra2020}.
The final loss $\ell_{\text{final}}$ is thus a weighted sum of all these losses:
\begin{equation}
    \ell_{\text{final}} = \ell + \kappa_{\text{corr}} \cdot \ell_{\text{corr}} + \kappa_{\text{entropy}} \cdot \ell_{\text{entropy}} + \kappa_{\text{im}} \cdot \ell_{\text{im}} \, .
\end{equation}

\subsection{Occlusion Detection}
\vspace{-.3em}
In order to detect whether a cuboid $\vec{{h}}$ occludes a point $\vec{y}$ from the perspective of a camera with centre $\vec{c}$, we must first translate $\vec{c}$ into the cuboid-centric coordinate frame.
Without loss of generality, we assume $\vec{c} = (0,0,0)\tran$ and thus:
\begin{equation}
    \vec{\hat{c}} = \vec{R}(\vec{c} - \vec{t}) = -\vec{Rt} \, .
\end{equation}
We parametrise the line of sight for a point $\vec{y}$ as:
\begin{equation}
    \vec{x}(\lambda) = \vec{\hat{y}} + \lambda \vec{v} \, , \, \mathrm{with} \, \vec{v} = \vec{\hat{c}}-\vec{\hat{y}} \, .
\end{equation}
We determine its intersections with each of the six cuboid planes.
For the plane orthogonal to the $x$-axis in its positive direction, this implies that
\begin{equation}
    \lambda = \frac{a_x - \vec{\hat{y}}\tran\vec{e}_x}{\vec{v}\tran\vec{e}_x},
\end{equation}
where $\vec{e}_x$ denotes the $x$-axis unit vector. 
If $\lambda < 0$, $\vec{{y}}$ lies in front of the plane and is thus not occluded. 
Otherwise, we must check whether the intersection actually lies on the cuboid, \ie $d(\vec{h}, \vec{x}(\lambda)) = 0$.
If that is the case, then $\vec{y}$ is occluded by this part of the cuboid. 
We repeat this check accordingly for the other five cuboid planes.
Via this procedure we define an indicator function:
\begin{equation}
    \chi_{\text{o}}(\vec{y}, \vec{h}, i) =
\begin{cases}
1 ~&\text{ if }~ i\text{-th plane of } \vec{h}  \text{ occludes }  \vec{y}, \\
0 ~&\text{ else }, \, \text{with } i \in \{1, \dots, 6\} \, .
\end{cases}
\end{equation}

\subsection{Cuboid Fitting}
\vspace{-.3em}
We implement the minimal solver $\vec{h} = f_{h}(\set{S})$, which estimates cuboid parameters $\vec{h}$ from a minimal set of features $\set{S} = \{\vec{y}_1, \dots, \vec{y}_C\}$, via iterative numerical optimisation using gradient descent.
To this end, we apply the Adam~\cite{KingmaB14} optimiser to minimise the objective $\frac{1}{|\set{S}|} \|F(\set{S}, \vec{h})\|_1$ (cf. Sec. 3.3 in the main paper).
We perform 50 steps of gradient descent with a learning rate of $0.2$, starting from an initial estimate $\vec{h}_0$.

\noindent
\textbf{Initialisation of the Minimal Solver.}
As good initialisation is crucial for fast convergence, we estimate the initial position of the cuboid via the mean of the features, \ie $\vec{t}_0 = \frac{1}{|\set{S}|} \sum_{\vec{y} \in \set{S}} \vec{y}$. Secondly we estimate the rotation $\vec{R}$ via singular value decomposition:
\begin{equation}
\vec{R}_0 = \vec{V}\tran, \, \text{ with } \vec{USV}\tran = \left[\vec{y}_1 \dots \vec{y}_{|\set{S}|}  \right]\tran \, .
\end{equation}

Lastly, we initialise the cuboid size with the element-wise maximum of the absolute coordinates of the centred and rotated features, with $ i \in \{1, \dots, |\set{S}|\}$:
\begin{equation}
    \begin{pmatrix}
    a_{x0} \\ a_{y0} \\ a_{z0} 
    \end{pmatrix}
    =
    \begin{pmatrix}
    \max_{i} |\hat{x}_i| \\ \max_{i} |\hat{y}_i| \\ \max_{i} |\hat{z}_i| 
    \end{pmatrix}
    \, , \text{ with }
    \begin{pmatrix}
    \hat{x}_i \\ \hat{y}_i \\ \hat{z}_i
    \end{pmatrix}
    =
    \vec{R}_0 (\vec{y}_i - \vec{t}_0) \, .
    \nonumber
\end{equation}
This gives us an initial estimate $\vec{h}_0 = (a_{x0}, a_{y0}, a_{z0}, \vec{R}_0, \vec{t}_0)$.

\subsection{Stopping Criterion}
\vspace{-.3em}
\label{subsec:cutoff}
During evaluation, we always predict up to $|\set{M}|$ cuboids. 
However, similar to~\cite{KluBra2020}, we select primitive instances sequentially.
The set $\set{M}$ contains the already selected primitives, and $\vec{h}$ is the next selected primitive.
We only add $\vec{h}$ to $\set{M}$ is it increases the joint inlier count by at least $\Theta$, \ie:
\begin{equation}
    I_{\text{c}}(\set{Y}, \set{M} \cup \{\vec{h}\}) - I_{\text{c}}(\set{Y}, \set{M}) > \Theta \, .
\end{equation}
Otherwise the method terminates and returns $\set{M}$ as the recovered primitive configuration.

\section{OA Distance for Superquadrics}
\vspace{-.3em}
\label{sec:oasq}
The surface of a superellipsoid~\cite{barr1981superquadrics}, which is the type of superquadric used in~\cite{paschalidou2019superquadrics,Paschalidou2020:Decomposition}, can be described by its inside-outside function:
\begin{equation}
    f_{\text{sq}}(x,y,z) = \left( \left( \frac{x}{a_x} \right)^{ \frac{2}{\epsilon_2}} \!\! + \left( \frac{y}{a_y} \right)^{\frac{2}{\epsilon_2}} \right)^{\frac{\epsilon_1}{\epsilon_2}} \!\! + \left( \frac{z}{a_y} \right)^{\frac{2}{\epsilon_1}} \!\! -1 \, ,
\end{equation}
with $\epsilon_1, \epsilon_2$ describing the shape of the superquadric, and $a_x, a_y, a_z$ describing its extent along the canonical axes. 
If $f_{\text{sq}} = 0$, the point $(x,y,z)$ resides on the superquadric surface. 
For $f_{\text{sq}} > 0$ and $f_{\text{sq}} < 0$, it is outside or inside the superquadric, respectively. 
Alternatively, a point on the superquadric surface can be described by:
\begin{equation}
    \vec{p}(\eta, \omega) = 
    \begin{bmatrix}
    a_x \cos^{\epsilon_1}(\eta) \cos^{\epsilon_2}(\omega) \\
    a_y \cos^{\epsilon_1}(\eta) \sin^{\epsilon_2}(\omega) \\
    a_z \sin^{\epsilon_1}(\eta) 
    \end{bmatrix} \, ,
    \label{eq:sq_io}
\end{equation}
parametrised by two angles $\eta,\omega$. The surface normal at such a point is defined as:
\begin{equation}
    \vec{n}(\eta, \omega) = 
    \begin{bmatrix}
    a_x^{-1} \cos^{2-\epsilon_1}(\eta) \cos^{2-\epsilon_2}(\omega) \\
    a_y^{-1} \cos^{2-\epsilon_1}(\eta) \sin^{2-\epsilon_2}(\omega) \\
    a_z^{-1} \sin^{2-\epsilon_1}(\eta) 
    \end{bmatrix} \, .
    \label{eq:sq_normal}
\end{equation}
Unfortunately, no closed form solution exists for calculating a point-to-superquadric distance, which we would need in order to compute the occlusion-aware distance metric as described in Sec. 3.2. 
We therefore approximate it by sampling points and determining occlusion and self-occlusion using Eq.~\ref{eq:sq_io} and Eq.~\ref{eq:sq_normal}.

\subsection{Occlusion}
\vspace{-.3em}
Given a 3D point $\vec{y} = \left[x,\,y,\,z\right]\tran$ and a camera centre $\vec{c} = \left[0,\,0,\,0\right]\tran$, we sample $L$ points uniformly on the line of sight:
\begin{equation}
    \set{Y}_{\text{los}} = \{\frac{1}{L} \vec{y}, \, \frac{2}{L} \vec{y}, \, \dots, \vec{y} \} = 
    \{\vec{y}_1, \, \vec{y}_2, \, \dots, \vec{y}_L \} \, .
\end{equation}
For a superquadric $\vec{h} = (\epsilon_1, \epsilon_2, a_x, a_y, a_z, \vec{R}, \vec{t})$, we transform all points in $\set{Y}_{\text{los}}$ into the superquadric-centric coordinate system:
\begin{equation}
    \vec{\hat{y}} = \vec{R}(\vec{y}-\vec{t}) \, ,
\end{equation}
and determine whether they are inside or outside of the superquadric:
\begin{equation}
    \set{F}_{\text{los}} = \{ f_{\text{sq}}(\vec{\hat{y}}_1), \dots, f_{\text{sq}}(\vec{\hat{y}}_L) \} \, .
\end{equation}
We then count the sign changes in $\set{F}_{\text{los}}$: if there are none, $\vec{y}$ is not occluded by the superquadric; otherwise, it is:
\begin{equation}
    \chi_{\text{sq}}(\vec{y}, \vec{h}) =
\begin{cases}
1 ~&\text{ if }~ \vec{h}  \text{ occludes }  \vec{y}, \\
0 ~&\text{ else } \, .
\end{cases}
\end{equation}

\subsection{Self-Occlusion}
\vspace{-.3em}
Using the source code provided by the authors of~\cite{paschalidou2019superquadrics}, we sample $N$ points $\vec{p}_i$ uniformly on the surface of superquadric $\vec{h}$ and determine their corresponding surface normals $\vec{n}_i$. 
For each point, we compute the vector $\vec{v}_i = \vec{p}_i - \hat{\vec{c}}$, with $\vec{\hat{c}} = \vec{R}(\vec{c}-\vec{t})$ being the camera centre in the superquadric-centric coordinate system. 
If $\vec{v}_i \tran \vec{n}_i = 0$, \ie the two vectors are orthogonal, $\vec{p}_i$ lies on the rim of the superquadric, which partitions it into a visible and an invisible part~\cite{jaklic2000segmentation}. 
Assuming $\vec{n}_i$ points outward, it follows that $\vec{p}_i$ is invisible if $\vec{v}_i \tran \vec{n}_i > 0$ and $f_{\text{sq}}(\hat{\vec{c}}) > 0$, in which case we discard it. 
We denote the set of visible points of $\vec{h}$ as $\set{P}(\vec{h})$.

\subsection{Occlusion-Aware Distance}
\vspace{-.3em}
We define the distance of a point $\vec{y}$ to one superquadric $\vec{h}$ as the minimum distance to any of its visible points $\set{P}$:
\begin{equation}
    d_{\text{sq}}(\vec{h}, \vec{y}) = \min_{\vec{p} \in \set{P}(\vec{h})} \| \vec{p}-\vec{y} \|_2 \, .
\end{equation}
Similarly to cuboids, we compute the distance of $\vec{y}$ to the most distant occluding superquadric, given a set of superquadrics $\set{M}$:
\begin{equation}
    d_{\text{o,sq}}(\set{M}, \vec{y}) = 
\max_{\vec{h} \in \set{M}} 
\left( \,
\chi_{\text{sq}}(\vec{y}, \vec{h}) \cdot d_{\text{sq}}(\vec{h}, \vec{y}) 
\, \right) \, ,
\end{equation}
and corresponding occlusion-aware distance:
\begin{equation}
    d_{\text{oa,sq}}(\set{M}, \vec{y}) = 
    \max 
    \left( 
    \min_{\vec{h} \in \set{M}} d_{\text{sq}}(\vec{h}, \vec{y}), \; d_{\text{o,sq}}(\set{M}, \vec{y}) 
    \right) \, . \nonumber
\end{equation}

\section{Ablation Study}
\vspace{-.3em}
\label{sec:ablation}

\begin{table*}
\setlength\tabcolsep{.37em}
	\begin{center}
	\begin{tabular}{|c|c|c|cc|cc|cc|cc|cc|cc|}
	\cline{1-15}
	\multicolumn{3}{|c|}{occlusion-aware inlier} & \multicolumn{10}{c|}{occlusion-aware $L_2$-distance}                 & \multicolumn{2}{c|}{$L_2$} \\
	\multicolumn{2}{|c|}{training} & \multicolumn{1}{c|}{inference} & \multicolumn{2}{c|}{\multirow{2}{*}{AUC$_{@\SI{50}{\centi\meter}}$}} & \multicolumn{2}{c|}{\multirow{2}{*}{AUC$_{@\SI{20}{\centi\meter}}$}} & \multicolumn{2}{c|}{\multirow{2}{*}{AUC$_{@\SI{10}{\centi\meter}}$}} & \multicolumn{2}{c|}{\multirow{2}{*}{AUC$_{@\SI{5}{\centi\meter}}$}}    & \multicolumn{2}{c|}{\multirow{2}{*}{mean (cm)}}      & \multicolumn{2}{c|}{\multirow{2}{*}{mean (cm)}} \\
	sampling & loss & sampling & & & & & & & & & & & & \\
	\hline
	\cmark & \cmark & \cmark    & $\mathbf{77.2}\%$ & $\scriptstyle \pm 0.08$  
                                & $\mathbf{62.7}\%$ & $\scriptstyle \pm 0.07$  
                                & $\mathbf{49.1}\%$ & $\scriptstyle \pm 0.12$  
                                & $\mathbf{34.3}\%$ & $\scriptstyle \pm 0.14$  
                                & ${20.8}$ &   $\scriptstyle \pm 0.51$  
                                & ${17.9}$ &   $\scriptstyle \pm 0.50$ \\
	\xmark & \cmark & \cmark    & $75.4\%$ & $\scriptstyle \pm 0.13$ 
	                            & $58.8\%$ & $\scriptstyle \pm 0.19$ 
	                            & $44.3\%$ & $\scriptstyle \pm 0.22$ 
	                            & $29.8\%$ & $\scriptstyle \pm 0.23$ 
	                            & $20.1$ & $\scriptstyle \pm 0.43$ 
	                            & $16.6$ & $\scriptstyle \pm 0.47$  \\
	\cmark & \xmark & \cmark    & $77.0\%$ & $\scriptstyle \pm 0.11$ 
	                            & $61.8\%$ & $\scriptstyle \pm 0.08$ 
	                            & $47.9\%$ & $\scriptstyle \pm 0.08$ 
	                            & $33.1\%$ & $\scriptstyle \pm 0.12$ 
	                            & $\mathbf{18.9}$ & $\scriptstyle \pm 0.19$ 
	                            & $15.5$ & $\scriptstyle \pm 0.17$  \\
	\xmark & \xmark & \cmark    & $76.6\%$ & $\scriptstyle \pm 0.09$ 
	                            & $61.2\%$ & $\scriptstyle \pm 0.10$ 
	                            & $47.1\%$ & $\scriptstyle \pm 0.11$ 
	                            & $32.4\%$ & $\scriptstyle \pm 0.13$ 
	                            & $19.0$ & $\scriptstyle \pm 0.37$ 
	                            & $15.5$ & $\scriptstyle \pm 0.42$  \\
	\cmark & \cmark & \xmark    & $11.0\%$ & $\scriptstyle \pm0.29 $ 
	                            & $5.3\%$ & $\scriptstyle \pm0.20 $ 
	                            & $3.3\%$ & $\scriptstyle \pm0.11 $ 
	                            & $2.0\%$ & $\scriptstyle \pm0.06 $ 
	                            & $151.7$ & $\scriptstyle \pm1.31 $ 
	                            & ${4.4}$ & $\scriptstyle \pm0.08 $  \\
	\xmark & \cmark & \xmark    & $11.1\%$ & $\scriptstyle \pm 0.14$ 
	                            & $5.4\%$ & $\scriptstyle \pm 0.08$ 
	                            & $3.3\%$ & $\scriptstyle \pm 0.06$ 
	                            & $1.9\%$ & $\scriptstyle \pm 0.04$ 
	                            & $152.7$ & $\scriptstyle \pm 0.49$ 
	                            & $4.5$ & $\scriptstyle \pm 0.07$  \\
	\cmark & \xmark & \xmark    & $10.3\%$ & $\scriptstyle \pm 0.31$ 
	                            & $5.0\%$ & $\scriptstyle \pm 0.16$ 
	                            & $3.1\%$ & $\scriptstyle \pm 0.12$ 
	                            & $1.8\%$ & $\scriptstyle \pm 0.08$ 
	                            & $156.4$ & $\scriptstyle \pm 1.43$ 
	                            & $4.2$ & $\scriptstyle \pm 0.06$  \\
	\xmark & \xmark & \xmark    & $10.0\%$ & $\scriptstyle \pm 0.10$ 
	                            & $4.7\%$ & $\scriptstyle \pm 0.11$ 
	                            & $2.9\%$ & $\scriptstyle \pm 0.08$ 
	                            & $1.7\%$ & $\scriptstyle \pm 0.06$ 
	                            & $157.4$ & $\scriptstyle \pm 1.25$ 
	                            & $\mathbf{3.9}$ & $\scriptstyle \pm 0.08$  \\

	\hline
	\end{tabular}
    \end{center}
 \vspace{-4mm}
	\caption{\textbf{Ablation Study:} We analyse the influence of occlusion-aware inlier counting. We enable or disable it for hypothesis sampling and selection during training and inference, and for loss computation during training. We evaluate on NYU Depth v2~\cite{silberman2012indoor} for depth input. We present AUC values (higher is better) for various upper bounds of the occlusion-aware (OA) $L_2$ distance. We also report mean OA-$L_2$ and regular $L_2$ distances (lower is better). See Sec.~\ref{sec:ablation} for details.}
	\label{tab:ablation_results}
\vspace{-4mm}
\end{table*}

In order to demonstrate the impact of our proposed occlusion-aware (OA) inlier counting (cf. Sec. 3.4 in the main paper), we selectively enabled or disabled the following parts of our method:
\begin{compactitem}
    \item During training: occlusion-aware inlier counting for sampling and hypothesis selection.
    \item During training: occlusion-aware inlier counting for loss computation (cf. Sec. 3.5 main paper).
    \item During inference: occlusion-aware inlier counting for sampling and hypothesis selection.
\end{compactitem}
When disabling the occlusion-aware inlier counting for one component, we used regular inlier counting based on the minimal $L_2$ distance instead.
We then evaluated these variants using ground truth depth input on the NYU~\cite{silberman2012indoor} test set, and show the results in Tab.~\ref{tab:ablation_results}.

As these results show, performance \wrt the occlusion-aware $L_2$ distance degrades severely when disabling the occlusion-aware inlier counting during inference, regardless of whether it was used during training (cf. rows 5-8 in Tab.~\ref{tab:ablation_results}).
AUC percentages drop to single digits for smaller upper bounds ($\SI{20}{\centi\meter}$ and below), and the mean increases from around $\SI{20}{\centi\meter}$ to more than $\SI{1.5}{\meter}$.
Conversely, the mean of the regular $L_2$ distance decreases by roughly a factor of four.
This indicates that the recovered cuboids cover more points, but create significantly more occlusions doing so, which results in less reasonable scene abstractions.
However, when we look at the impact of occlusion-aware inlier counting during training (cf. rows 1-4 in Tab.~\ref{tab:ablation_results}), the differences are not as clear-cut.
Using occlusion-aware inlier counting for both sampling and loss during training still yields the highest AUC values, albeit with slimmer margins, between $0.2$ and $4.8$ percentage points.
Disabling occlusion-aware inlier counting for the loss only (row 3) yields the second best AUC values, with the smallest margin at AUC$_{@\SI{50}{\centi\meter}}$ and the largest margin at AUC$_{@\SI{10}{\centi\meter}}$.
Meanwhile, this configuration yields the lowest mean occlusion-aware distance -- $\SI{18.9}{\centi\meter}$ vs. $\SI{20.8}{\centi\meter}$, \ie $9.1\%$ less.
This indicates fewer outlier points with errors $>\SI{50}{\centi\meter}$, which strongly influence the mean error, but lower representation accuracy for inlier points with errors $<\SI{50}{\centi\meter}$.

\section{Qualitative Results}
\vspace{-.3em}
\label{sec:qualitative}

\subsection{Occlusion-Aware Inlier Counting}
\begin{figure*}	
		\centering
		\begin{subfigure}[t]{0.99\linewidth}
			\centering
			\includegraphics[width=0.16\linewidth]{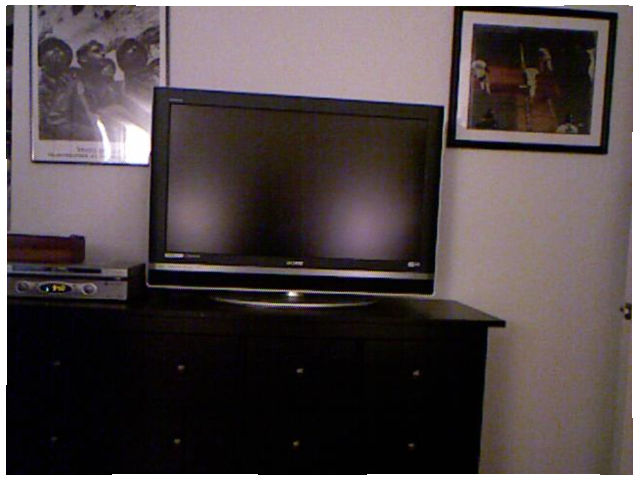}
			\includegraphics[width=0.16\linewidth]{supp/oai/008_0322_0764_original.jpg}
			\includegraphics[width=0.16\linewidth]{supp/oai/015_0459_1095_original.jpg}
			\includegraphics[width=0.16\linewidth]{supp/oai/017_0112_0282_original.jpg}
			\includegraphics[width=0.16\linewidth]{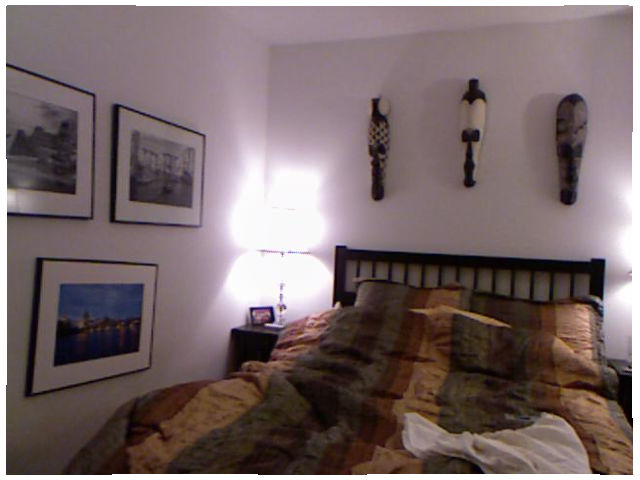}
			\includegraphics[width=0.16\linewidth]{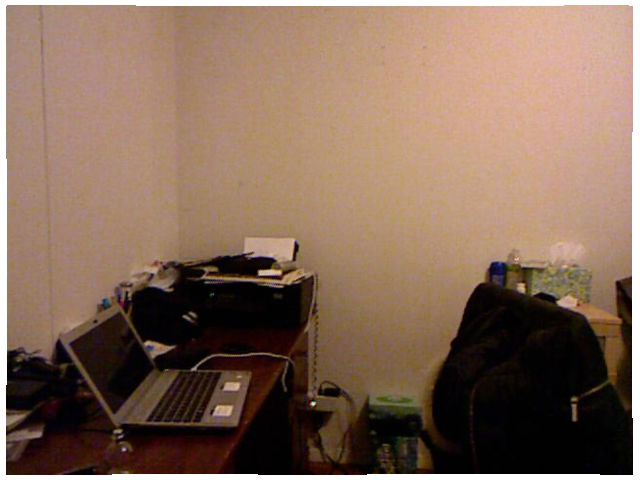}
			\vspace{-.5em}
			\caption{Original images}
			\vspace{.75em}
		\end{subfigure}
		\begin{subfigure}[t]{0.99\linewidth}
			\centering
			\makebox[0.16\linewidth][l]{\small AUC$_{@\SI{5}{\centi\meter}}$: $70.2\%$}
			\makebox[0.16\linewidth][c]{\small $57.7\%$}
			\makebox[0.16\linewidth][c]{\small $57.3\%$}
			\makebox[0.16\linewidth][c]{\small $53.7\%$}
			\makebox[0.16\linewidth][c]{\small $50.8\%$}
			\makebox[0.16\linewidth][c]{\small $56.3\%$}
			\includegraphics[width=0.16\linewidth]{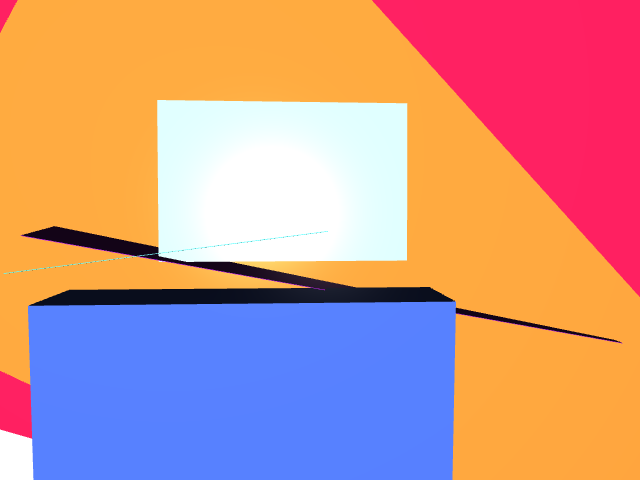}
			\includegraphics[width=0.16\linewidth]{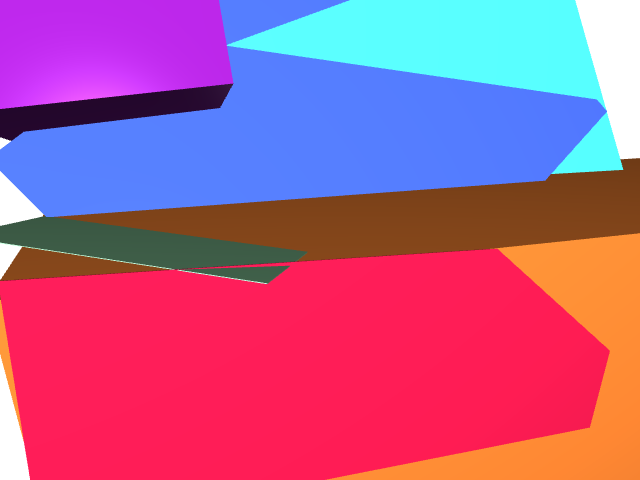}
			\includegraphics[width=0.16\linewidth]{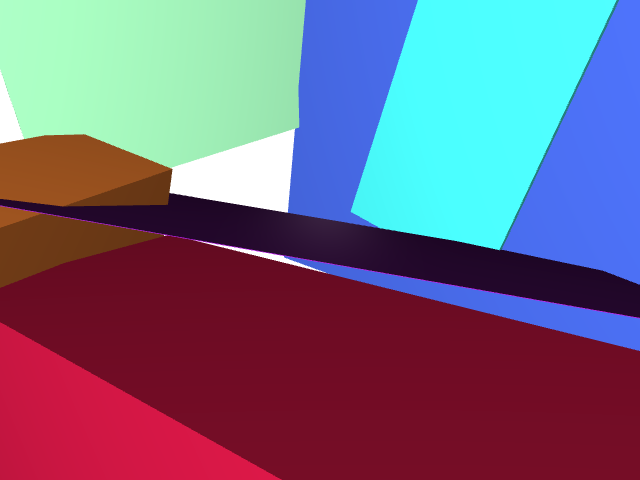}
			\includegraphics[width=0.16\linewidth]{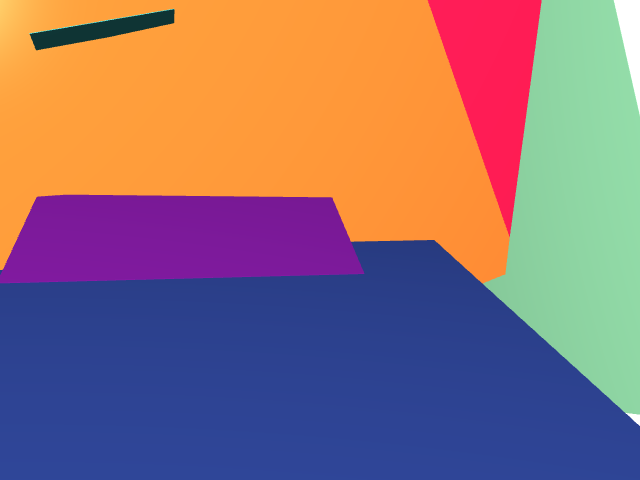}
			\includegraphics[width=0.16\linewidth]{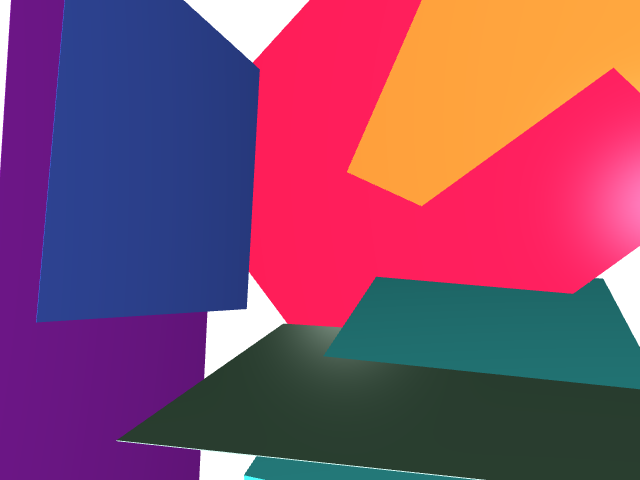}
			\includegraphics[width=0.16\linewidth]{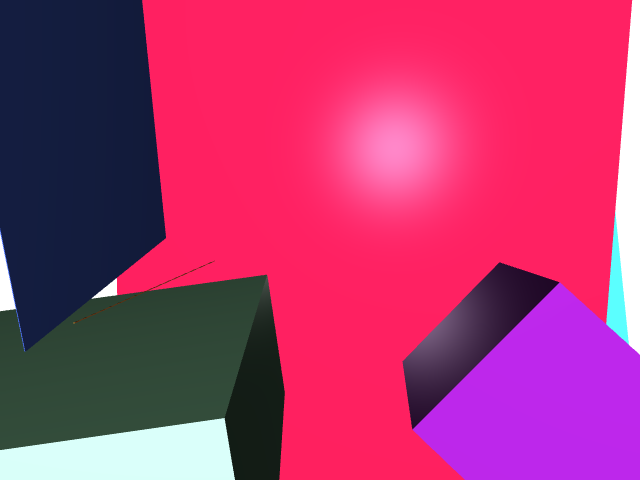}
			\vspace{-.5em}
			\caption{Recovered cuboids using occlusion-aware inlier counting, original view}
			\vspace{.35em}
		\end{subfigure}
		\begin{subfigure}[t]{0.99\linewidth}
			\centering
			\includegraphics[width=0.16\linewidth]{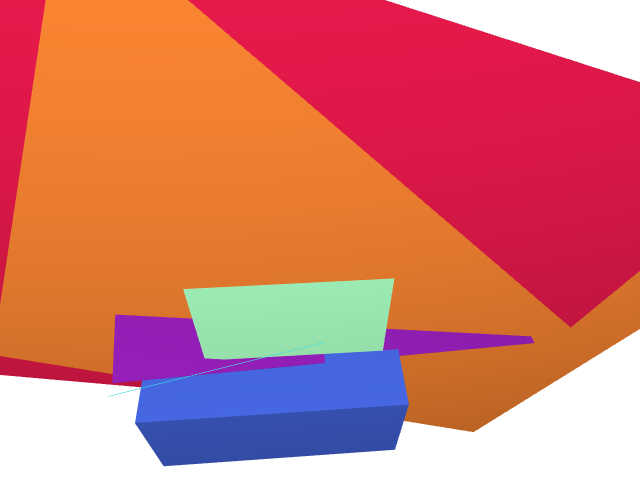}
			\includegraphics[width=0.16\linewidth]{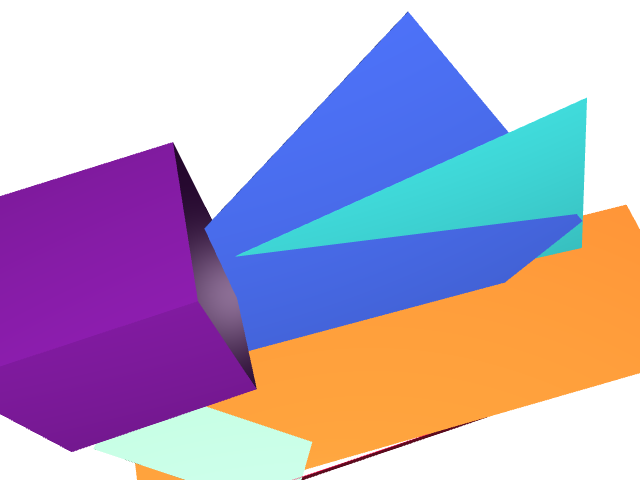}
			\includegraphics[width=0.16\linewidth]{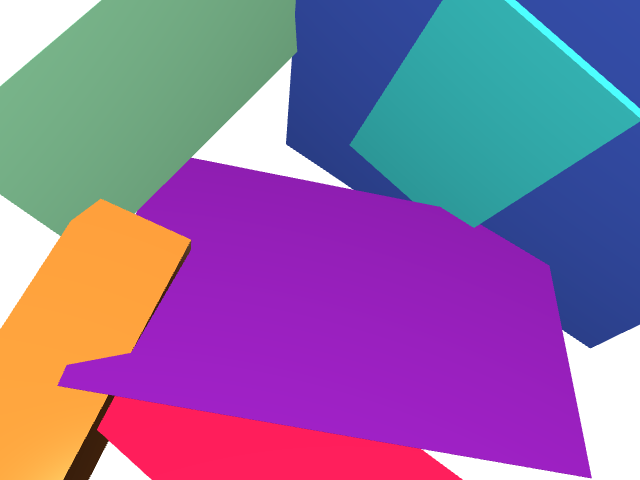}
			\includegraphics[width=0.16\linewidth]{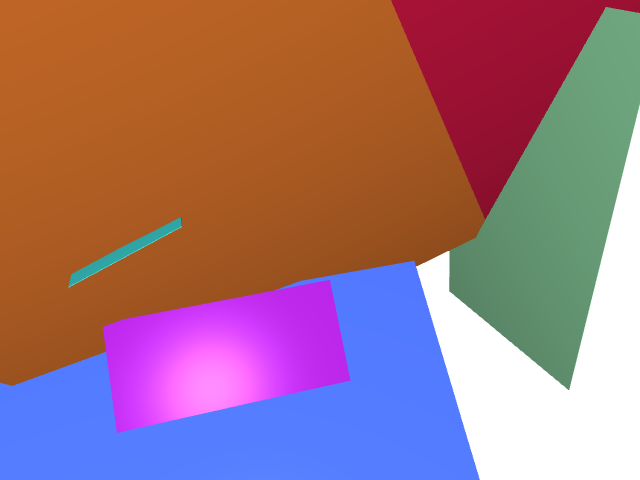}
			\includegraphics[width=0.16\linewidth]{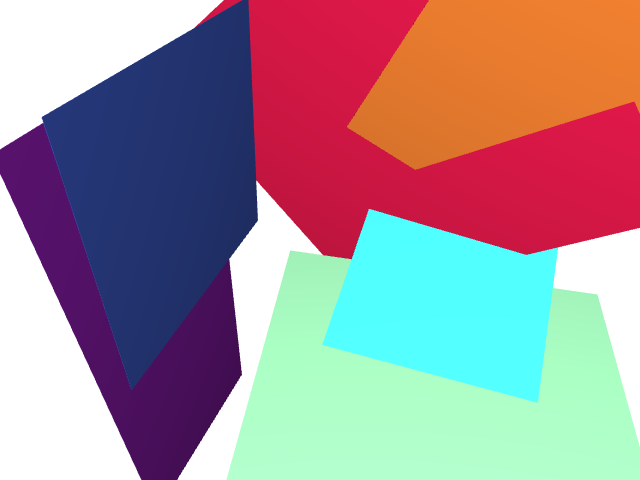}
			\includegraphics[width=0.16\linewidth]{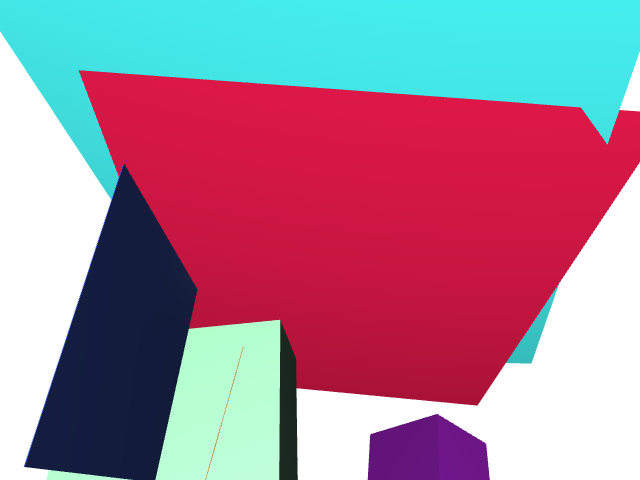}
			\vspace{-.5em}
			\caption{Recovered cuboids using occlusion-aware inlier counting, top view}
			\vspace{.35em}
		\end{subfigure}
		\begin{subfigure}[t]{0.99\linewidth}
			\centering
			\includegraphics[width=0.16\linewidth]{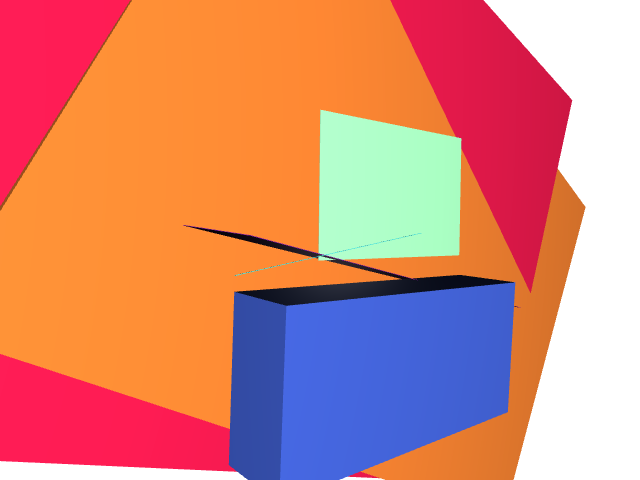}
			\includegraphics[width=0.16\linewidth]{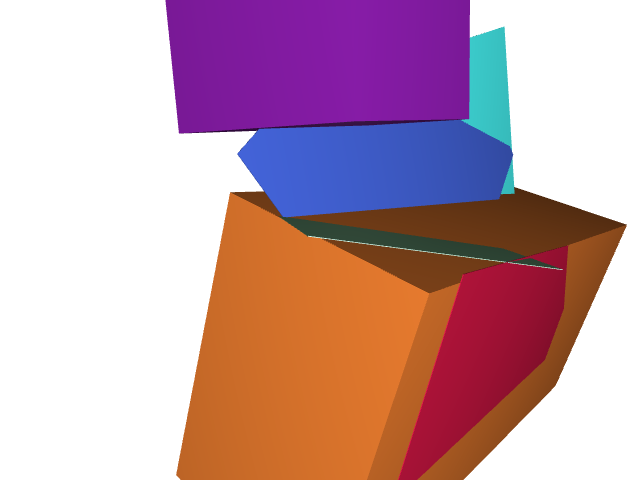}
			\includegraphics[width=0.16\linewidth]{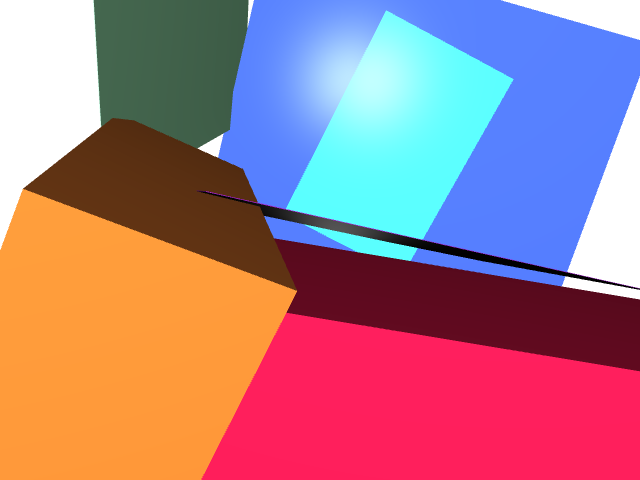}
			\includegraphics[width=0.16\linewidth]{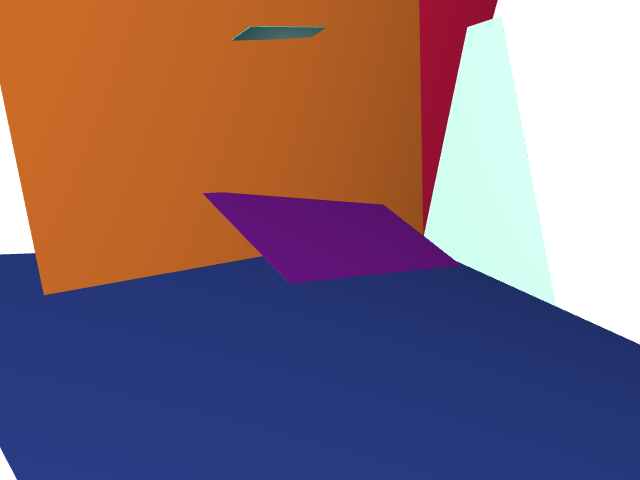}
			\includegraphics[width=0.16\linewidth]{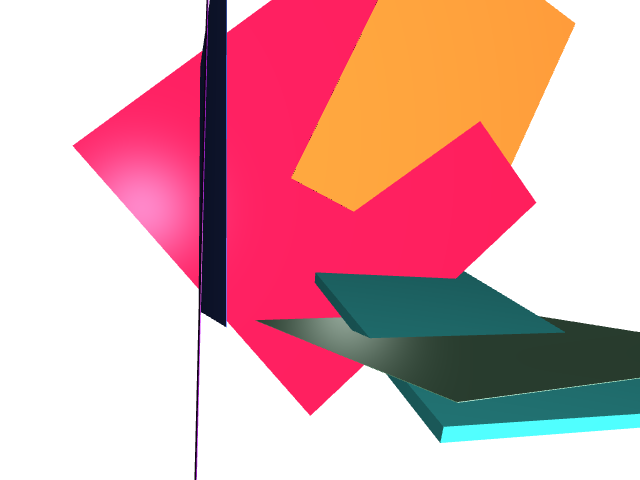}
			\includegraphics[width=0.16\linewidth]{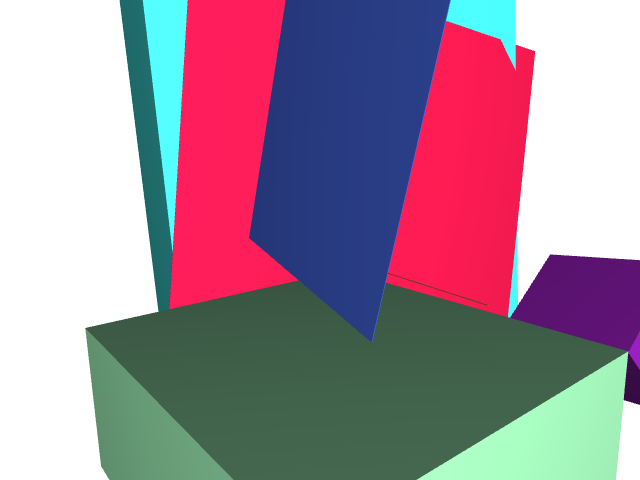}
			\vspace{-.5em}
			\caption{Recovered cuboids using occlusion-aware inlier counting, side view}
			\vspace{.75em}
		\end{subfigure}
	\begin{subfigure}[t]{0.99\linewidth}
			\centering
			\makebox[0.16\linewidth][l]{\small AUC$_{@\SI{5}{\centi\meter}}$: $17.5\%$}
			\makebox[0.16\linewidth][c]{\small $12.6\%$}
			\makebox[0.16\linewidth][c]{\small $0.0\%$}
			\makebox[0.16\linewidth][c]{\small $19.8\%$}
			\makebox[0.16\linewidth][c]{\small $0.0\%$}
			\makebox[0.16\linewidth][c]{\small $2.2\%$}
			\includegraphics[width=0.16\linewidth]{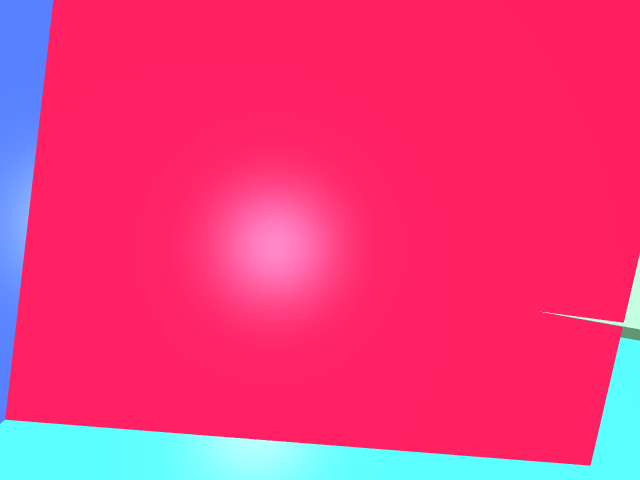}
			\includegraphics[width=0.16\linewidth]{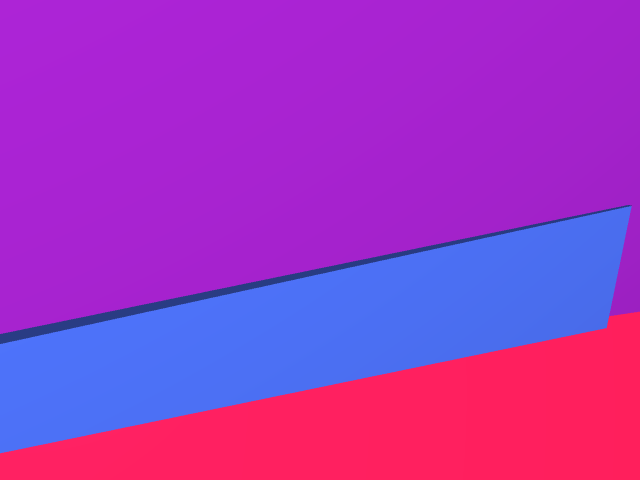}
			\includegraphics[width=0.16\linewidth]{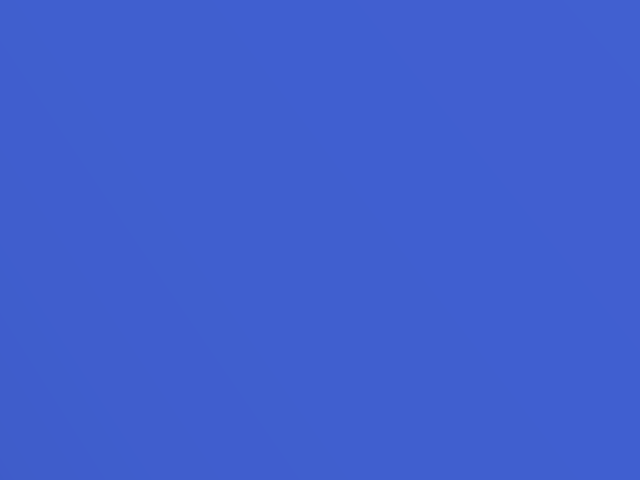}
			\includegraphics[width=0.16\linewidth]{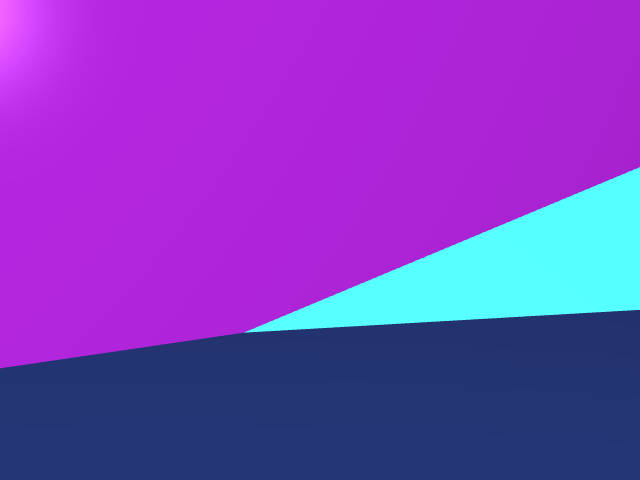}
			\includegraphics[width=0.16\linewidth]{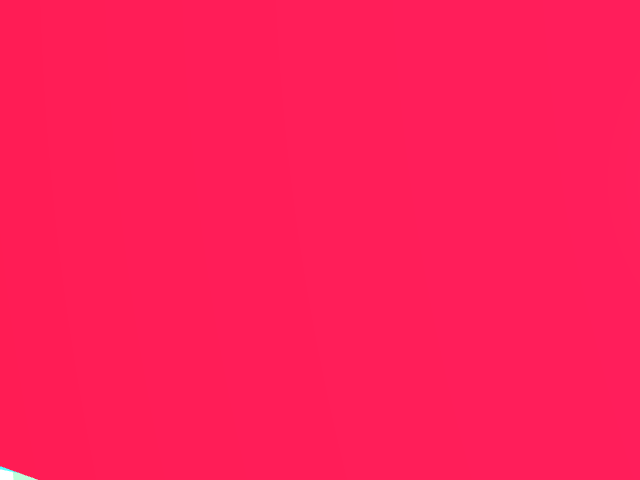}
			\includegraphics[width=0.16\linewidth]{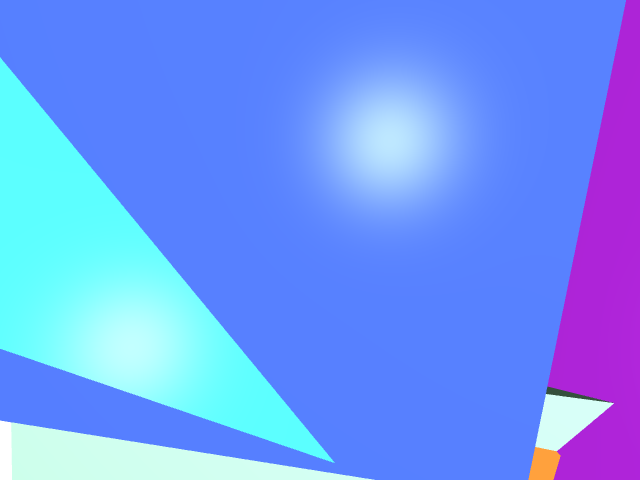}
			\vspace{-.5em}
			\caption{Recovered cuboids without occlusion-aware inlier counting, original view}
			\vspace{.35em}
		\end{subfigure}
	\begin{subfigure}[t]{0.99\linewidth}
			\centering
			\includegraphics[width=0.16\linewidth]{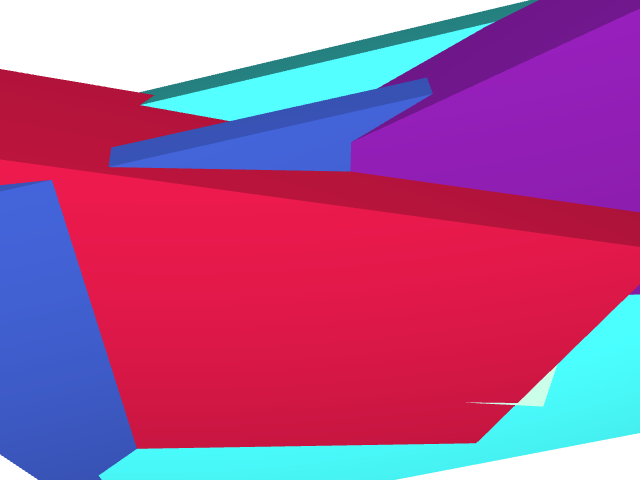}
			\includegraphics[width=0.16\linewidth]{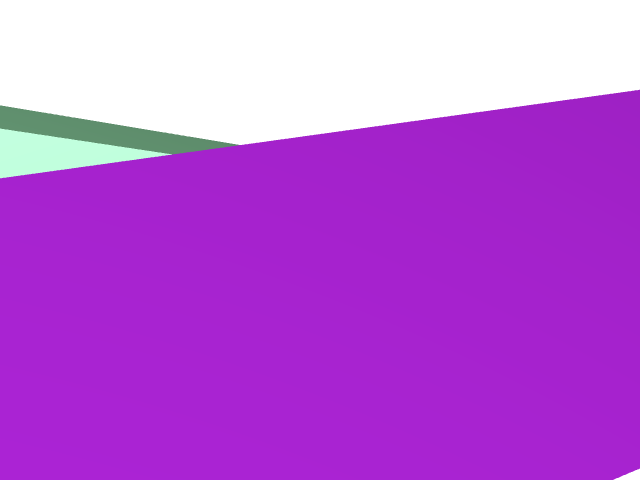}
			\includegraphics[width=0.16\linewidth]{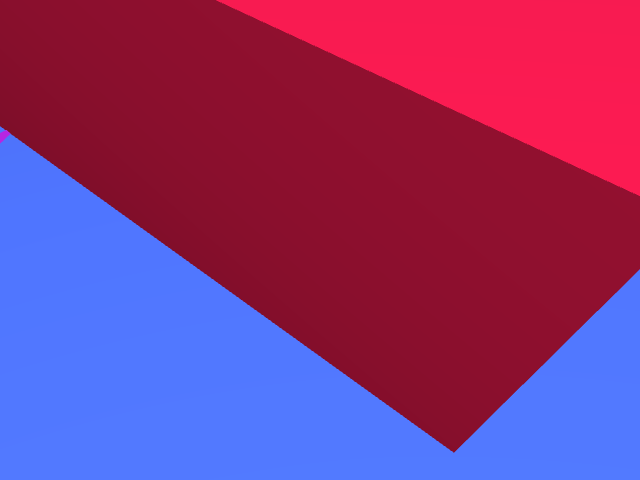}
			\includegraphics[width=0.16\linewidth]{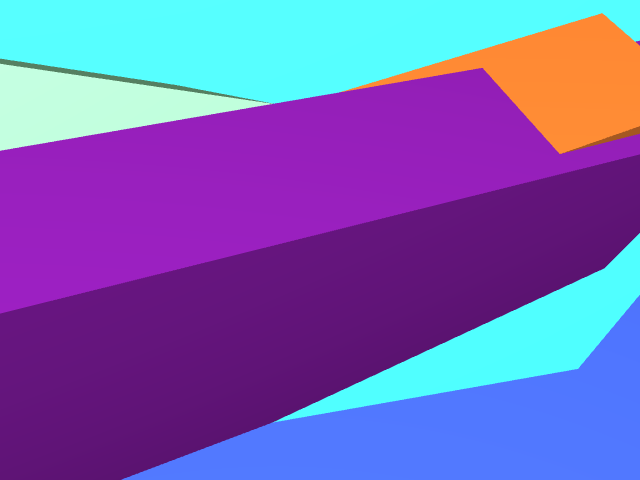}
			\includegraphics[width=0.16\linewidth]{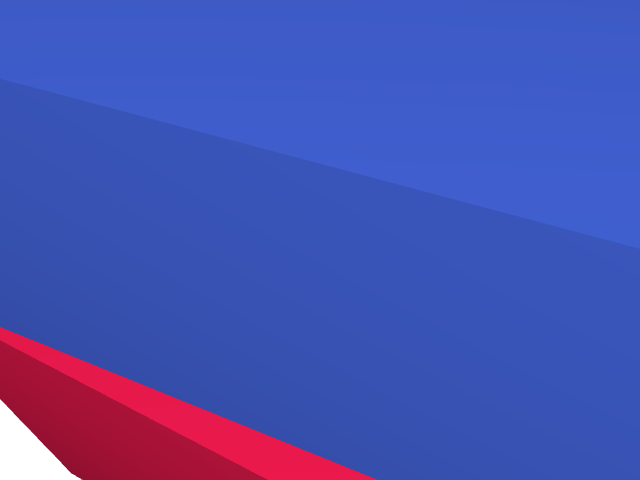}
			\includegraphics[width=0.16\linewidth]{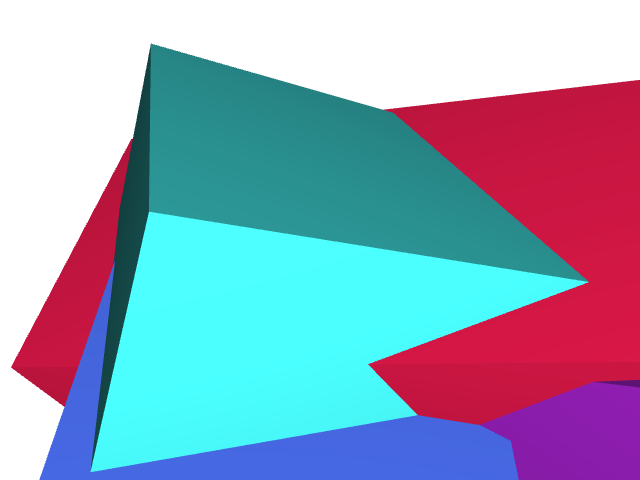}
			\vspace{-.5em}
			\caption{Recovered cuboids without occlusion-aware inlier counting, top view}
			\vspace{.35em}
		\end{subfigure}
	\begin{subfigure}[t]{0.99\linewidth}
			\centering
			\includegraphics[width=0.16\linewidth]{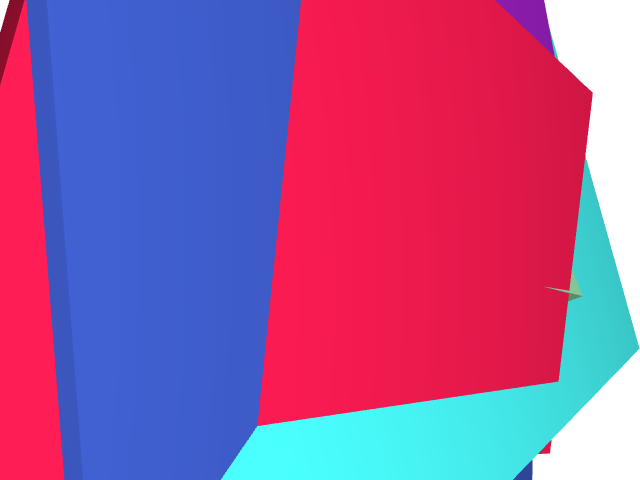}
			\includegraphics[width=0.16\linewidth]{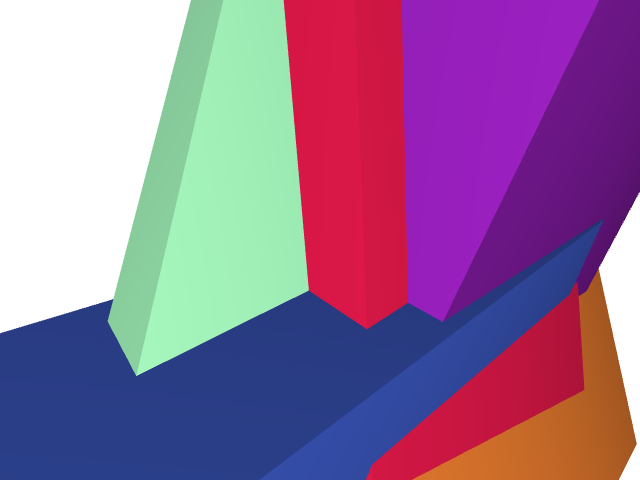}
			\includegraphics[width=0.16\linewidth]{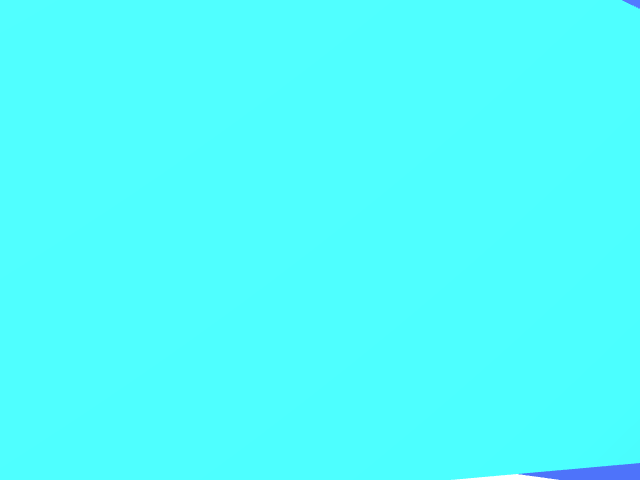}
			\includegraphics[width=0.16\linewidth]{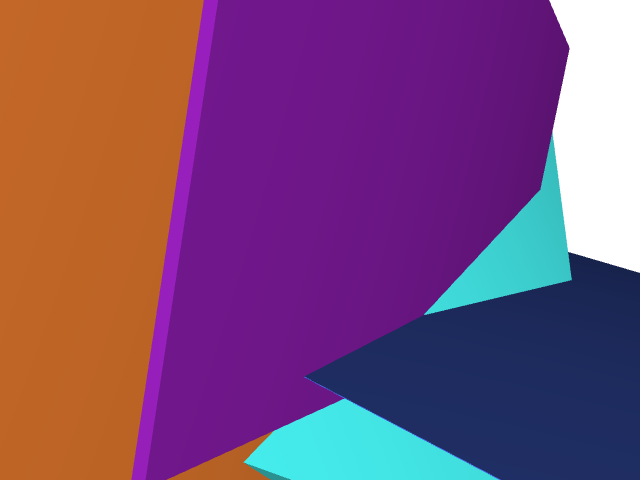}
			\includegraphics[width=0.16\linewidth]{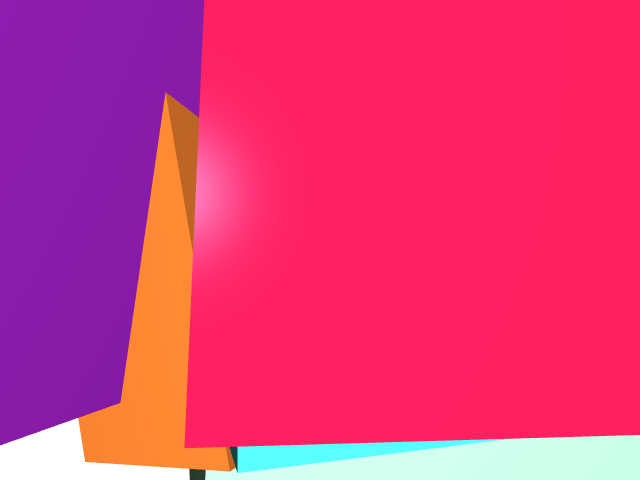}
			\includegraphics[width=0.16\linewidth]{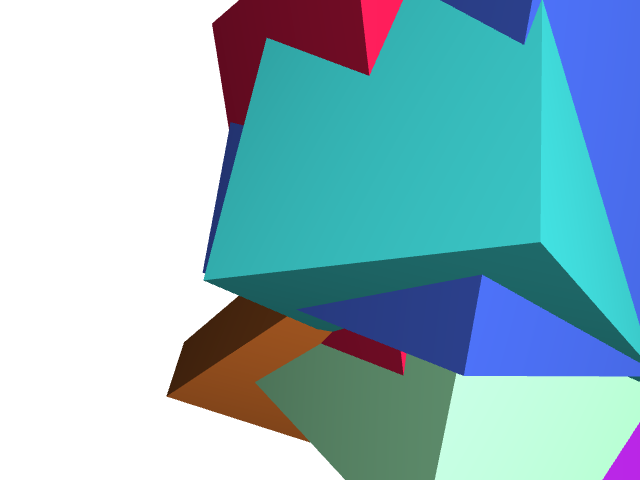}
			\vspace{-.5em}
			\caption{Recovered cuboids without occlusion-aware inlier counting, side view}
			\vspace{.35em}
		\end{subfigure}
		\caption{\textbf{Occlusion-Aware Inlier Counting.} First row: Original images from the NYU dataset. Rows (b)-(d): cuboids recovered from ground truth depth using our method with occlusion-aware inlier counting during inference. Rows (e)-(g): results \emph{without} occlusion-aware inlier counting. We show views from the perspective of the original image, as well as top and side views. We also report the  AUC$_{@\SI{5}{\centi\meter}}$ values for each example above rows (b) and (e). 
		See Sec.~\ref{subsec:qualitative_oai} for details.}
		\label{fig:oai_examples}
\end{figure*}

\vspace{-.3em}
\label{subsec:qualitative_oai}
Complementing the ablation study in Sec.~\ref{sec:ablation}, we compare example results with and without occlusion-aware inlier counting during inference in Fig.~\ref{fig:oai_examples}.
As in Sec.~\ref{sec:ablation}, results are based on ground truth depth input for the NYU dataset.
As these examples show, the occlusion-aware inlier counting enables our method to reasonably abstract key elements of the scenes.
Without occlusion-aware inlier counting, however, the method predicts cuboids which are too large and often intersecting, occluding most of the scene.
The AUC$_{@\SI{5}{\centi\meter}}$ values, which we also provide for each example in the figure, emphasise this observation.

\subsection{Failure Cases}
\begin{figure}
		\centering
		\begin{subfigure}[t]{0.99\linewidth}
			\centering
			\includegraphics[width=0.32\linewidth]{supp/fail/006_0134_0330_original.jpg}
			\includegraphics[width=0.32\linewidth]{supp/fail/011_0564_1275_original.jpg}
			\includegraphics[width=0.32\linewidth]{supp/fail/017_0043_0088_original.jpg}
			\vspace{-.5em}
			\caption{Original images}
			\vspace{.35em}
		\end{subfigure}
		\begin{subfigure}[t]{0.99\linewidth}
			\centering
			\makebox[0.32\linewidth][l]{\small AUC$_{@\SI{5}{\centi\meter}}$: $17.1\%$}
			\makebox[0.32\linewidth][c]{\small $13.1\%$}
			\makebox[0.32\linewidth][c]{\small $15.2\%$}
			\includegraphics[width=0.32\linewidth]{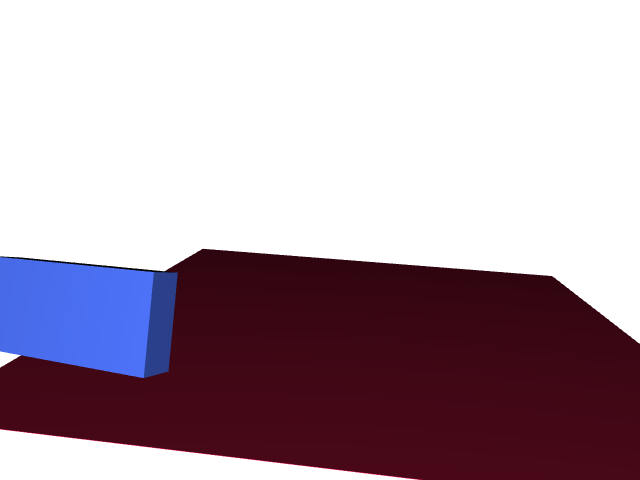}
			\includegraphics[width=0.32\linewidth]{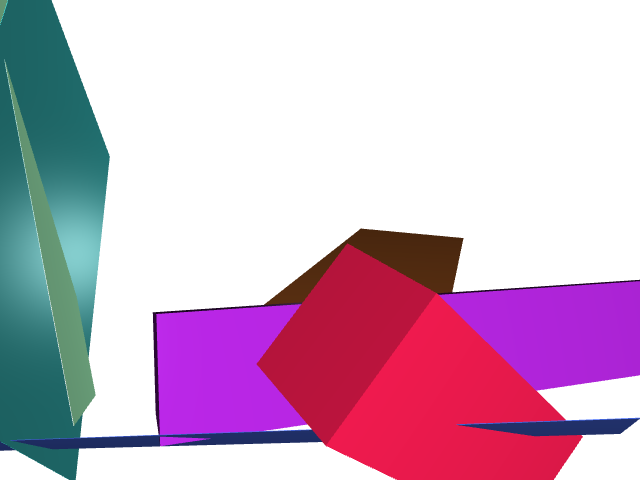}
			\includegraphics[width=0.32\linewidth]{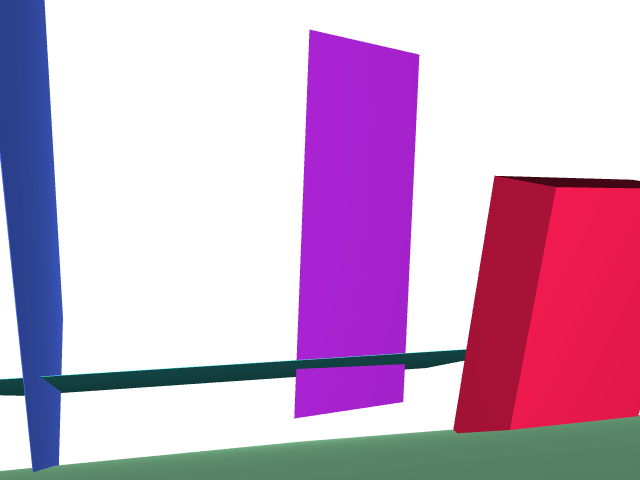}
			\vspace{-.5em}
			\caption{Recovered cuboids, original view}
			\vspace{.35em}
		\end{subfigure}
		\begin{subfigure}[t]{0.99\linewidth}
			\centering
			\includegraphics[width=0.32\linewidth]{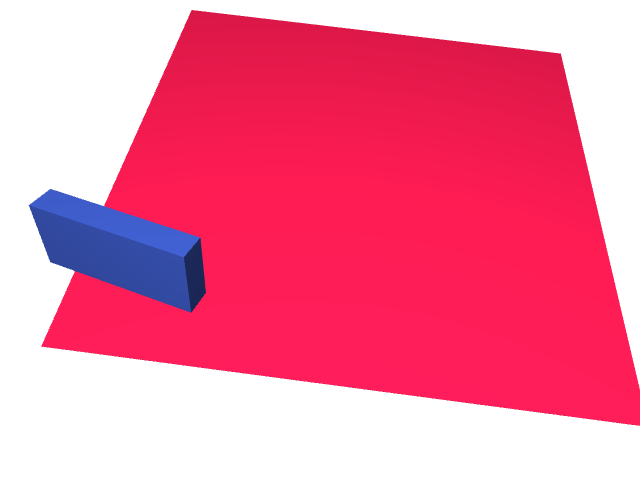}
			\includegraphics[width=0.32\linewidth]{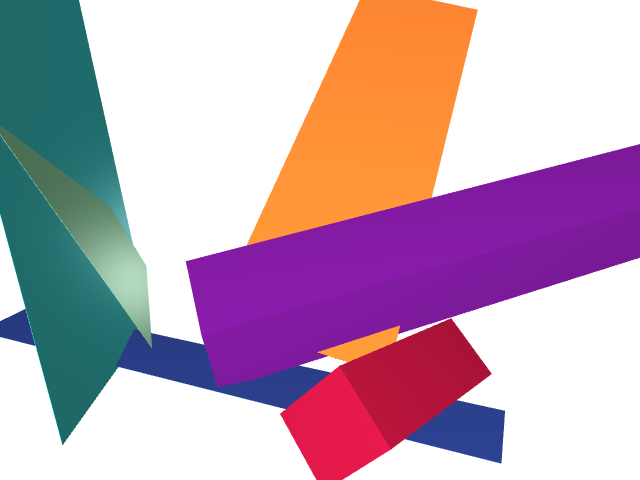}
			\includegraphics[width=0.32\linewidth]{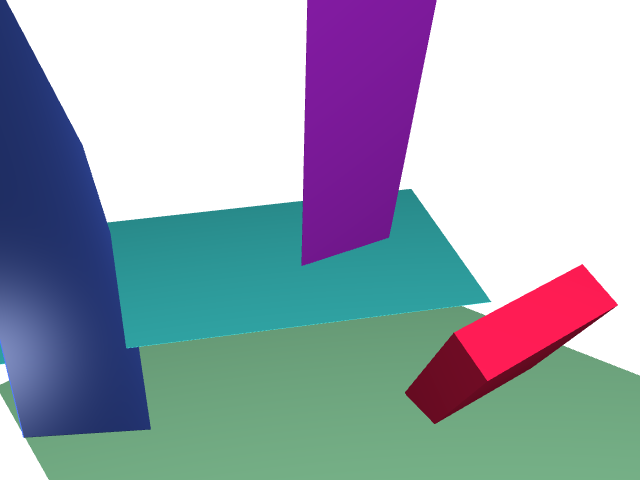}
			\vspace{-.5em}
			\caption{Recovered cuboids, top view}
			\vspace{.35em}
		\end{subfigure}
		\caption{\textbf{Failure Cases.} First row: Original images from the NYU dataset. Rows (b)-(c): cuboids recovered from ground truth depth using our method, showing views from the perspective of the original image, as well as top views. We also report the AUC$_{@\SI{5}{\centi\meter}}$ values for each example above row (b). 
		See Sec.~\ref{subsec:fail} for details.}
		\label{fig:fail_examples}
\vspace{-2mm}
\end{figure}

\vspace{-.3em}
\label{subsec:fail}
We show failure cases of our method, \ie examples with below average AUC, in Fig.~\ref{fig:fail_examples}.
Most common failure modes of our approach are: 
\begin{compactitem}
    \item Missing scene parts, for which no cuboid was fitted (first and third column).
    \item Cuboids which are too large, too small, improperly oriented, or simply too coarse (second column).
    \item Spurious cuboids, usually very thin and barely visible form the original view (second and third column).
\end{compactitem}
We conjecture that the first two failure modes can be mitigated via more effective sampling.
The third failure mode may require additional consideration in the inlier counting procedure, or possibly just an increased instance cutoff threshold (cf. Sec.~\ref{subsec:cutoff}).

\subsection{Cuboid Parsing Baseline}
\begin{figure*}	
		\centering
		\begin{subfigure}[t]{0.99\linewidth}
			\centering
			\includegraphics[width=0.16\linewidth]{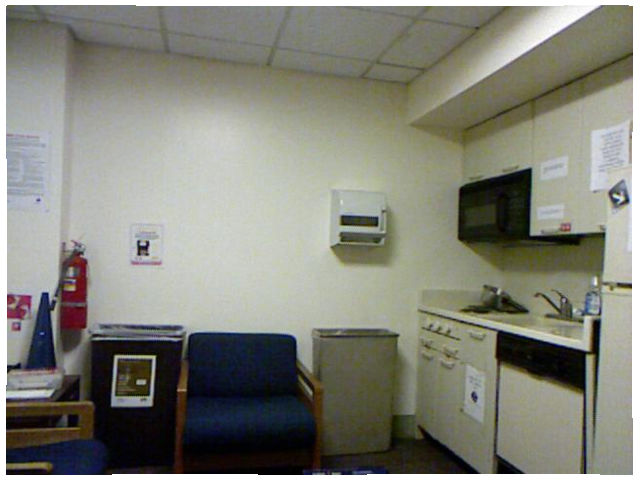}
			\includegraphics[width=0.16\linewidth]{supp/volprim/0010.jpg}
			\includegraphics[width=0.16\linewidth]{supp/volprim/0020.jpg}
			\includegraphics[width=0.16\linewidth]{supp/volprim/0836.jpg}
			\includegraphics[width=0.16\linewidth]{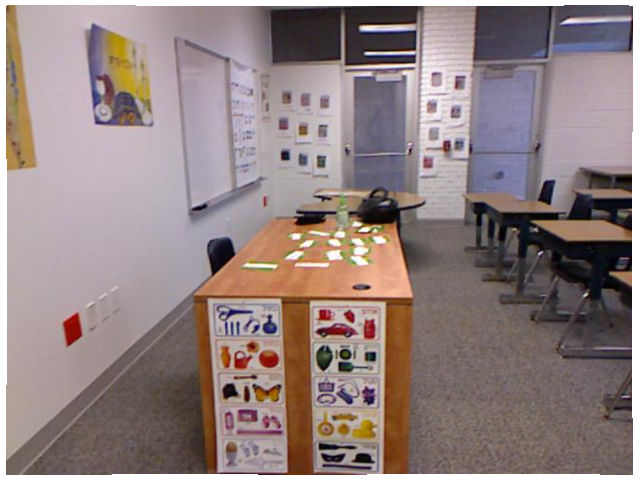}
			\includegraphics[width=0.16\linewidth]{supp/volprim/0570.jpg}
			\vspace{-.5em}
			\caption{Original images}
			\vspace{.35em}
		\end{subfigure}
		\begin{subfigure}[t]{0.99\linewidth}
			\centering
			\includegraphics[width=0.16\linewidth]{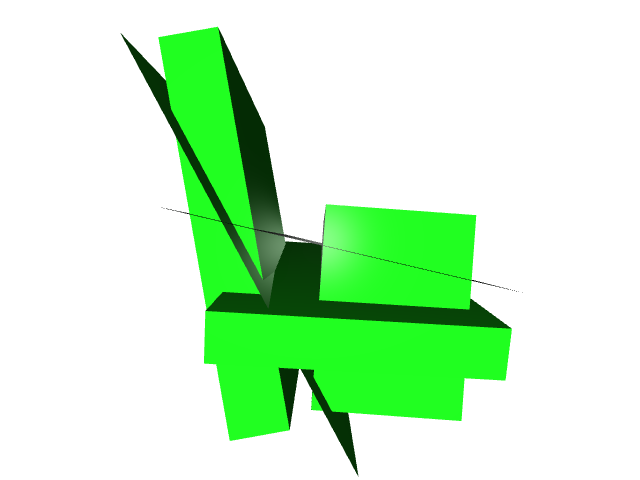}
			\includegraphics[width=0.16\linewidth]{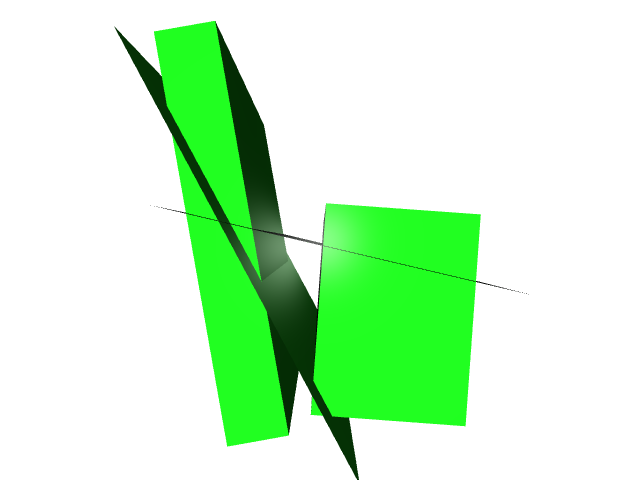}
			\includegraphics[width=0.16\linewidth]{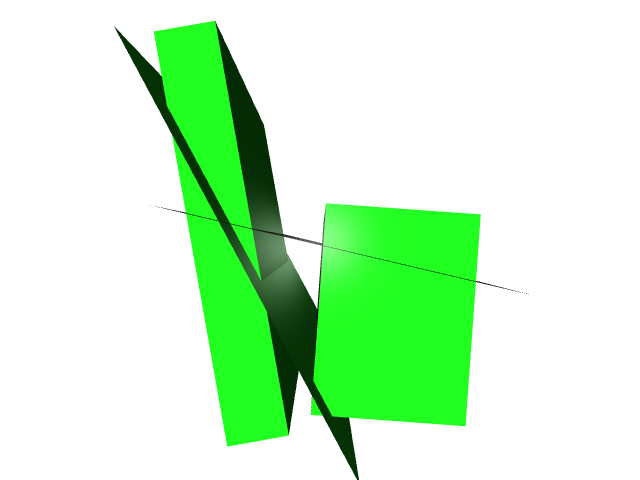}
			\includegraphics[width=0.16\linewidth]{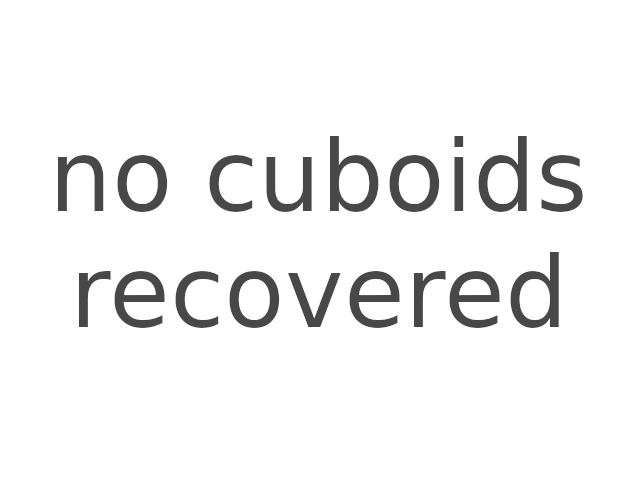}
			\includegraphics[width=0.16\linewidth]{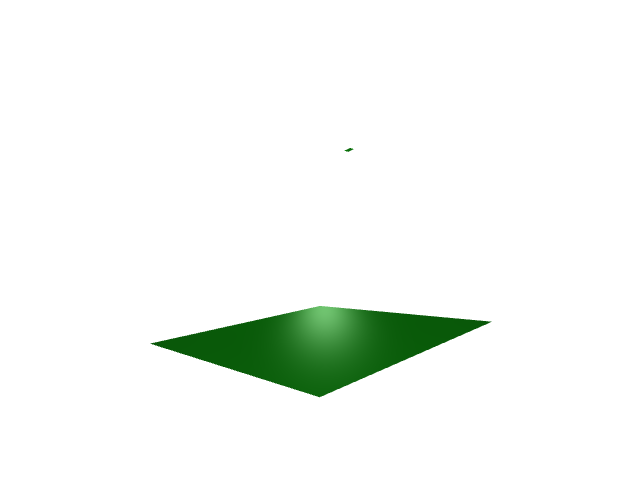}
			\includegraphics[width=0.16\linewidth]{supp/volprim/no_cuboids.png}
			\vspace{-.5em}
			\caption{Recovered cuboids using~\cite{tulsiani2017learning}, first training run}
			\vspace{.35em}
		\end{subfigure}
		\begin{subfigure}[t]{0.99\linewidth}
			\centering
			\includegraphics[width=0.16\linewidth]{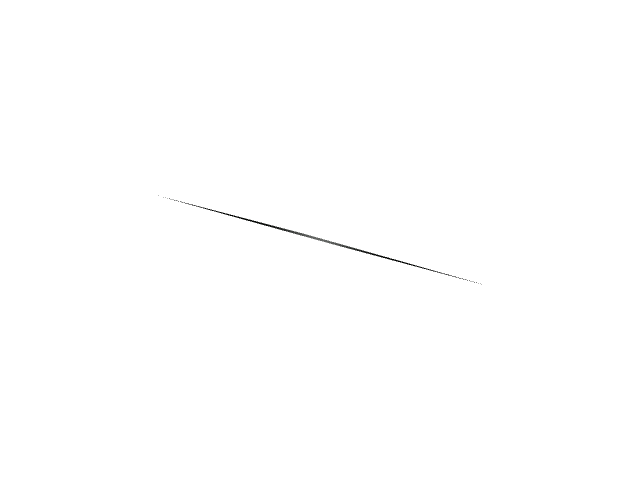}
			\includegraphics[width=0.16\linewidth]{supp/volprim/no_cuboids.png}
			\includegraphics[width=0.16\linewidth]{supp/volprim/no_cuboids.png}
			\includegraphics[width=0.16\linewidth]{supp/volprim/no_cuboids.png}
			\includegraphics[width=0.16\linewidth]{supp/volprim/no_cuboids.png}
			\includegraphics[width=0.16\linewidth]{supp/volprim/no_cuboids.png}
			\vspace{-.5em}
			\caption{Recovered cuboids using~\cite{tulsiani2017learning}, second training run}
			\vspace{.35em}
		\end{subfigure}
		\begin{subfigure}[t]{0.99\linewidth}
			\centering
			\includegraphics[width=0.16\linewidth]{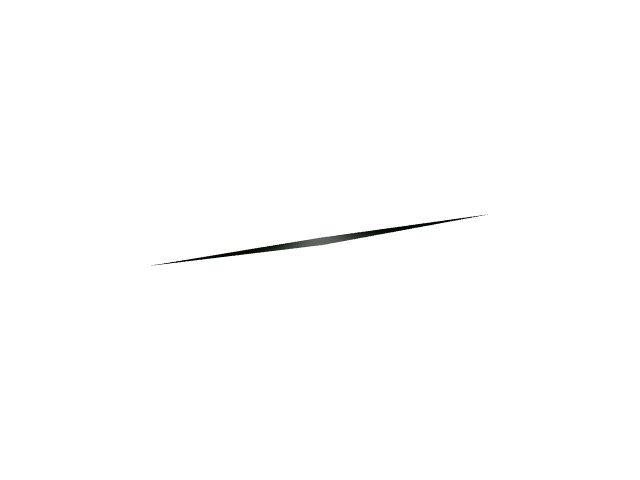}
			\includegraphics[width=0.16\linewidth]{supp/volprim/no_cuboids.png}
			\includegraphics[width=0.16\linewidth]{supp/volprim/no_cuboids.png}
			\includegraphics[width=0.16\linewidth]{supp/volprim/no_cuboids.png}
			\includegraphics[width=0.16\linewidth]{supp/volprim/no_cuboids.png}
			\includegraphics[width=0.16\linewidth]{supp/volprim/no_cuboids.png}
			\vspace{-.5em}
            \caption{Recovered cuboids using~\cite{tulsiani2017learning}, third training run}
			\vspace{.35em}
		\end{subfigure}
		
		\caption{\textbf{Cuboid Parsing Baseline:} We show qualitative results for the cuboid based approach of~\cite{tulsiani2017learning}. Row (a) shows the original images from the NYU dataset. Rows (b)-(d) show corresponding cuboid predictions by~\cite{tulsiani2017learning} for three different training runs using ground truth depth as input. As these examples show, the method is unable to recover sensible cuboid configurations for these real-world indoor scenes. See Sec.~\ref{subsec:volprim} for details. }
		\label{fig:volprim_examples}
\end{figure*}

\vspace{-.3em}
\label{subsec:volprim}
As mentioned in the main paper, we were unable to obtain sensible results with the cuboid based approach of~\cite{tulsiani2017learning} for the NYU dataset.
We trained their approach on NYU using ground truth depth input, following their instructions published together with their source code, using the same input data as we did for the superquadrics approach of~\cite{paschalidou2019superquadrics}, multiple times with different random seeds, but to no avail.
Their approach mostly predicts the same or very similar cuboid configurations for different images.
Often, no cuboids are recovered at all.
We show a couple of examples for three different training runs in Fig.~\ref{fig:volprim_examples}.

\end{appendices}

\newpage
{\small
\bibliographystyle{ieee_fullname}
\bibliography{cvpr}

\begin{thebibliography}{10}\itemsep=-1pt

\bibitem{amayo2018geometric}
Paul Amayo, Pedro Pini{\'e}s, Lina~M Paz, and Paul Newman.
\newblock Geometric {M}ulti-{M}odel {F}itting with a {C}onvex {R}elaxation
  {A}lgorithm.
\newblock In {\em CVPR}, 2018.

\bibitem{barath2018graph}
Daniel Barath and Ji{\v{r}}{\'\i} Matas.
\newblock Graph-{C}ut {RANSAC}.
\newblock In {\em CVPR}, 2018.

\bibitem{barath2018multi}
Daniel Barath and Jiri Matas.
\newblock Multi-{C}lass {M}odel {F}itting by {E}nergy {M}inimization and
  {M}ode-{S}eeking.
\newblock In {\em ECCV}, 2018.

\bibitem{barath2019progressive}
Daniel Barath and Jiri Matas.
\newblock Progressive-{X}: {E}fficient, {A}nytime, {M}ulti-{M}odel {F}itting
  {A}lgorithm.
\newblock {\em ICCV}, 2019.

\bibitem{barr1981superquadrics}
Alan~H Barr.
\newblock Superquadrics and angle-preserving transformations.
\newblock {\em CGA}, 1981.

\bibitem{biederman1987recognition}
Irving Biederman.
\newblock Recognition-by-{C}omponents: a {T}heory of {H}uman {I}mage
  {U}nderstanding.
\newblock {\em Psychological Review}, 1987.

\bibitem{Brachmann2017:DSAC}
Eric Brachmann, Alexander Krull, Sebastian Nowozin, Jamie Shotton, Frank
  Michel, Stefan Gumhold, and Carsten Rother.
\newblock {DSAC} - {D}ifferentiable {RANSAC} for {C}amera {L}ocalization.
\newblock In {\em CVPR}, 2017.

\bibitem{Brachmann2019:NGRANSAC}
Eric Brachmann and Carsten Rother.
\newblock Neural-{G}uided {RANSAC:} {L}earning {W}here to {S}ample {M}odel
  {H}ypotheses.
\newblock In {\em ICCV}, 2019.

\bibitem{Chang2015:ShapeNet}
Angel~X. Chang, Thomas Funkhouser, Leonidas Guibas, Pat Hanrahan, Qixing Huang,
  Zimo Li, Silvio Savarese, Manolis Savva, Shuran Song, Hao Su, Jianxiong Xiao,
  Li Yi, and Fisher Yu.
\newblock Shape{N}et: {A}n {I}nformation-{R}ich 3{D} {M}odel {R}epository.
\newblock arXiv preprint arXiv:1512.03012, 2015.

\bibitem{Chin2014:AccHypGen}
Tat-Jun Chin, Jin Yu, and David Suter.
\newblock Accelerated {H}ypothesis {G}eneration for {M}ulti-structure {D}ata
  via {P}reference {A}nalysis.
\newblock {\em TPAMI}, 2014.

\bibitem{criminisi2000single}
Antonio Criminisi, Ian Reid, and Andrew Zisserman.
\newblock Single view metrology.
\newblock {\em IJCV}, 2000.

\bibitem{deng2020cvxnet}
Boyang Deng, Kyle Genova, Soroosh Yazdani, Sofien Bouaziz, Geoffrey Hinton, and
  Andrea Tagliasacchi.
\newblock Cvx{N}et: {L}earnable {C}onvex {D}ecomposition.
\newblock In {\em CVPR}, 2020.

\bibitem{Dwibedi2016:Beyond2dBBs}
Debidatta Dwibedi, Tomasz Malisiewicz, Vijay Badrinarayanan, and Andrew
  Rabinovich.
\newblock Deep {C}uboid {D}etection: {B}eyond 2{D} {B}ounding {B}oxes.
\newblock {\em arXiv preprint arXiv:1611.10010}, 2016.

\bibitem{Eigen2014:DepthMap}
David Eigen, Christian Puhrsch, and Rob Fergus.
\newblock Depth {M}ap {P}rediction from a {S}ingle {I}mage using a
  {M}ulti-{S}cale {D}eep {N}etwork.
\newblock In {\em NeurIPS}, 2014.

\bibitem{Fischler1981:RANSAC}
Martin~A. Fischler and Robert~C. Bolles.
\newblock Random sample consensus: A paradigm for model fitting with
  applications to image analysis and automated cartography.
\newblock {\em ACM}, 1981.

\bibitem{Garg0216:UnsupervisedDepth}
R. Garg, V.~K. Bg, G. Carneiro, and I. Reid.
\newblock Unsupervised {CNN} for {S}ingle {V}iew {D}epth {E}stimation:
  {G}eometry to the {R}escue.
\newblock In {\em ECCV}, 2016.

\bibitem{genova2019learning}
Kyle Genova, Forrester Cole, Daniel Vlasic, Aaron Sarna, William~T Freeman, and
  Thomas Funkhouser.
\newblock Learning {S}hape {T}emplates with {S}tructured {I}mplicit
  {F}unctions.
\newblock In {\em ICCV}, 2019.

\bibitem{gkioxari2019mesh}
Georgia Gkioxari, Jitendra Malik, and Justin Johnson.
\newblock Mesh {R-CNN}.
\newblock In {\em ICCV}, 2019.

\bibitem{Godard2017:DepthEstimation}
Cl\'ement Godard, Oisin Mac~Aodha, and Gabriel~J. Brostow.
\newblock Unsupervised {M}onocular {D}epth {E}stimation with {L}eft-{R}ight
  {C}onsistency.
\newblock In {\em CVPR}, 2017.

\bibitem{Godard2019:SelfSupervised}
Cl{\'e}ment Godard, Oisin Mac~Aodha, Michael Firman, and Gabriel~J Brostow.
\newblock Digging {I}nto {S}elf-{S}upervised {M}onocular {D}epth {E}stimation.
\newblock In {\em ICCV}, 2019.

\bibitem{groueix2018papier}
Thibault Groueix, Matthew Fisher, Vladimir~G Kim, Bryan~C Russell, and Mathieu
  Aubry.
\newblock A {P}apier-{M}{\^a}ch{\'e} {A}pproach to {L}earning 3{D} {S}urface
  {G}eneration.
\newblock In {\em CVPR}, 2018.

\bibitem{GuptaEfrosHebert_ECCV10}
Abhinav Gupta, Alexei~A. Efros, and Martial Hebert.
\newblock Blocks {W}orld {R}evisited: {I}mage {U}nderstanding {U}sing
  {Q}ualitative {G}eometry and {M}echanics.
\newblock In {\em ECCV}, 2010.

\bibitem{he2015delving}
Kaiming He, Xiangyu Zhang, Shaoqing Ren, and Jian Sun.
\newblock Delving deep into rectifiers: {S}urpassing human-level performance on
  {ImageNet} classification.
\newblock In {\em ICCV}, 2015.

\bibitem{he2016deep}
Kaiming He, Xiangyu Zhang, Shaoqing Ren, and Jian Sun.
\newblock Deep residual learning for image recognition.
\newblock In {\em CVPR}, 2016.

\bibitem{huang2017densely}
Gao Huang, Zhuang Liu, Laurens Van Der~Maaten, and Kilian~Q Weinberger.
\newblock Densely connected convolutional networks.
\newblock In {\em CVPR}, 2017.

\bibitem{ioffe2015batch}
Sergey Ioffe and Christian Szegedy.
\newblock Batch normalization: {A}ccelerating deep network training by reducing
  internal covariate shift.
\newblock In {\em ICML}, 2015.

\bibitem{jaklic2000segmentation}
Ales Jaklic, Ales Leonardis, Franc Solina, and F Solina.
\newblock {\em Segmentation and recovery of superquadrics}.
\newblock Springer Science \& Business Media, 2000.

\bibitem{Jiang2013:Linear}
Hao Jiang and Jianxiong Xiao.
\newblock A {L}inear {A}pproach to {M}atching {C}uboids in {RGBD} {I}mages.
\newblock In {\em CVPR}, 2013.

\bibitem{kazhdan2006poisson}
Michael Kazhdan, Matthew Bolitho, and Hugues Hoppe.
\newblock Poisson {S}urface {R}econstruction.
\newblock In {\em Eurographics Symposium on Geometry Processing}, 2006.

\bibitem{Kehl2018:3dDetection}
W. Kehl, F. Manhardt, F. Tombari, S. Ilic, and N. Navab.
\newblock {SSD-6D}: {M}aking {RGB}-based 3{D} detection and 6{D} pose
  estimation great again.
\newblock In {\em ICCV}, 2017.

\bibitem{KingmaB14}
Diederik~P. Kingma and Jimmy Ba.
\newblock Adam: {A} {M}ethod for {S}tochastic {O}ptimization.
\newblock In {\em ICLR}, 2015.

\bibitem{kluger2017deep}
Florian Kluger, Hanno Ackermann, Michael~Ying Yang, and Bodo Rosenhahn.
\newblock Deep learning for vanishing point detection using an inverse gnomonic
  projection.
\newblock In {\em GCPR}, 2017.

\bibitem{kluger2020temporally}
Florian Kluger, Hanno Ackermann, Michael~Ying Yang, and Bodo Rosenhahn.
\newblock Temporally consistent horizon lines.
\newblock In {\em ICRA}, 2020.

\bibitem{KluBra2020}
Florian Kluger, Eric Brachmann, Hanno Ackermann, Carsten Rother, Michael~Ying
  Yang, and Bodo Rosenhahn.
\newblock {CONSAC}: {R}obust {M}ulti-{M}odel {F}itting by {C}onditional
  {S}ample {C}onsensus.
\newblock In {\em CVPR}, 2020.

\bibitem{Kuznietsov2017:Semi}
Yevhen Kuznietsov, J\"org St\"uckler, and Bastian Leibe.
\newblock Semi-{S}upervised {D}eep {L}earning for {M}onocular {D}epth {M}ap
  {P}rediction.
\newblock In {\em CVPR}, 2017.

\bibitem{lee2019big}
Jin~Han Lee, Myung-Kyu Han, Dong~Wook Ko, and Il~Hong Suh.
\newblock From {B}ig to {S}mall: {M}ulti-{S}cale {L}ocal {P}lanar {G}uidance
  for {M}onocular {D}epth {E}stimation.
\newblock {\em arXiv preprint arXiv:1907.10326}, 2019.

\bibitem{Lin2013:Holistic}
Dahua Lin, Sanja Fidler, and Raquel Urtasun.
\newblock Holistic {S}cene {U}nderstanding for 3{D} {O}bject {D}etection with
  {RGBD} {C}ameras.
\newblock In {\em ICCV}, 2013.

\bibitem{liu2019planercnn}
Chen Liu, Kihwan Kim, Jinwei Gu, Yasutaka Furukawa, and Jan Kautz.
\newblock Plane{RCNN}: 3{D} {P}lane {D}etection and {R}econstruction from a
  {S}ingle {I}mage.
\newblock In {\em CVPR}, 2019.

\bibitem{Liu2018:PlaneNet}
Chen Liu, Jimei Yang, Duygu Ceylan, Ersin Yumer, and Yasutaka Furukawa.
\newblock Plane{N}et: {P}iece-wise {P}lanar {R}econstruction from a {S}ingle
  {RGB} {I}mage.
\newblock In {\em CVPR}, 2018.

\bibitem{liu89}
Dong Liu and Jorge Nocedal.
\newblock On the limited memory {B}{F}{G}{S} method for large scale
  optimization.
\newblock {\em Mathematical Programming}, 1989.

\bibitem{Liu2015:Depth}
Fayao Liu, Chunhua Shen, Guosheng Lin, and Ian Reid.
\newblock Learning {D}epth from {S}ingle {M}onocular {I}mages using {D}eep
  {C}onvolutional {N}eural {F}ields.
\newblock {\em TPAMI}, 2015.

\bibitem{Ma2019:SelfSupervised}
Fangchang Ma, Guilherme~Venturelli Cavalheiro, and Sertac Karaman.
\newblock Self-supervised {S}parse-to-{D}ense: {S}elf-supervised {D}epth
  {C}ompletion from {L}idar and {M}onocular {C}amera.
\newblock In {\em ICRA}, 2019.

\bibitem{magri2019fitting}
Luca Magri and Andrea Fusiello.
\newblock Fitting {M}ultiple {H}eterogeneous {M}odels by {M}ulti-{C}lass
  {C}ascaded {T}-{L}inkage.
\newblock In {\em CVPR}, 2019.

\bibitem{Mahjourian2018:Unsupervised}
Reza Mahjourian, Martin Wicke, and Anelia Angelova.
\newblock Unsupervised {L}earning of {D}epth and {E}go-{M}otion from
  {M}onocular {V}ideo {U}sing 3{D} {G}eometric {C}onstraints.
\newblock In {\em CVPR}, 2018.

\bibitem{mescheder2019occupancy}
Lars Mescheder, Michael Oechsle, Michael Niemeyer, Sebastian Nowozin, and
  Andreas Geiger.
\newblock Occupancy {N}etworks: {L}earning 3{D} {R}econstruction in {F}unction
  {S}pace.
\newblock In {\em CVPR}, 2019.

\bibitem{Mousavian2017:3dBB}
A. Mousavian, D. Anguelov, J. Flynn, and J. Kosecka.
\newblock 3{D} {B}ounding {B}ox {E}stimation {U}sing {D}eep {L}earning and
  {G}eometry.
\newblock In {\em CVPR}, 2017.

\bibitem{Nie2020:Total3dUnderstanding}
Y. Nie, X. Han, S. Guo, Y. Zheng, J. Chang, and J. Zhang.
\newblock Total3{DU}nderstanding: {J}oint {L}ayout, {O}bject {P}ose and {M}esh
  {R}econstruction for {I}ndoor {S}cenes from a {S}ingle {I}mage.
\newblock In {\em CVPR}, 2020.

\bibitem{niu2018im2struct}
Chengjie Niu, Jun Li, and Kai Xu.
\newblock Im2{S}truct: {R}ecovering 3{D} {S}hape {S}tructure from a {S}ingle
  {RGB} image.
\newblock In {\em CVPR}, 2018.

\bibitem{Pan2019:Mesh}
Junyi Pan, Xiaoguang Han, Weikai Chen, Jiapeng Tang, and Kui Jia.
\newblock Deep mesh reconstruction from single {RGB} images via topology
  modification networks.
\newblock In {\em ICCV}, 2019.

\bibitem{Paschalidou2020:Decomposition}
Despoina Paschalidou, Luc~Van Gool, and Andreas Geiger.
\newblock Learning {U}nsupervised {H}ierarchical {P}art {D}ecomposition of 3{D}
  {O}bjects from a {S}ingle {RGB} {I}mage.
\newblock In {\em CVPR}, 2020.

\bibitem{paschalidou2019superquadrics}
Despoina Paschalidou, Ali~Osman Ulusoy, and Andreas Geiger.
\newblock Superquadrics {R}evisited: {L}earning 3{D} {S}hape {P}arsing {B}eyond
  {C}uboids.
\newblock In {\em CVPR}, 2019.

\bibitem{paszke2017automatic}
Adam Paszke, Sam Gross, Soumith Chintala, Gregory Chanan, Edward Yang, Zachary
  DeVito, Zeming Lin, Alban Desmaison, Luca Antiga, and Adam Lerer.
\newblock Automatic differentiation in {PyTorch}.
\newblock In {\em NIPS-W}, 2017.

\bibitem{pham2014interacting}
Trung~Thanh Pham, Tat-Jun Chin, Konrad Schindler, and David Suter.
\newblock Interacting {G}eometric {P}riors for {R}obust {M}ultimodel {F}itting.
\newblock {\em Transactions on Image Processing (TIP)}, 2014.

\bibitem{PurChi2014a}
Pulak Purkait, Tat-Jun Chin, Hanno Ackermann, and David Suter.
\newblock Clustering with {H}ypergraphs: {T}he {C}ase for {L}arge {H}yperedges.
\newblock In {\em ECCV}, 2014.

\bibitem{Qi2018:FrustumPNets}
C.~R. Qi, W. Liu, C. Wu, H. Su, and L.~J. Guibas.
\newblock Frustum {P}oint{N}ets for 3{D} {O}bject {D}etection from {RGB-D}
  {D}ata.
\newblock In {\em CVPR}, 2018.

\bibitem{Ren2016:3dObjectDetection}
Zhile Ren and Erik~B Sudderth.
\newblock Three-{D}imensional {O}bject {D}etection and {L}ayout {P}rediction
  using {C}louds of {O}riented {G}radients.
\newblock In {\em CVPR}, 2016.

\bibitem{roberts1963machine}
Lawrence~G Roberts.
\newblock {\em Machine {P}erception of {T}hree-{D}imensional {S}olids}.
\newblock PhD thesis, Massachusetts Institute of Technology, 1963.

\bibitem{Shi2019:PRCNN}
S. Shi, X. Wang, and H. Li.
\newblock Point{RCNN}: 3{D} {O}bject {P}roposal {G}eneration and {D}etection
  from {P}oint {C}loud.
\newblock In {\em CVPR}, 2019.

\bibitem{silberman2012indoor}
Nathan Silberman, Derek Hoiem, Pushmeet Kohli, and Rob Fergus.
\newblock Indoor {S}egmentation and {S}upport {I}nference from {RGBD} {I}mages.
\newblock In {\em ECCV}, 2012.

\bibitem{Thomas2019ICCV}
Hugues Thomas, Charles~R. Qi, Jean-Emmanuel Deschaud, Beatriz Marcotegui,
  Francois Goulette, and Leonidas~J. Guibas.
\newblock {KPC}onv: {F}lexible and {D}eformable {C}onvolution for {P}oint
  {C}louds.
\newblock In {\em ICCV}, 2019.

\bibitem{tulsiani2017learning}
Shubham Tulsiani, Hao Su, Leonidas~J Guibas, Alexei~A Efros, and Jitendra
  Malik.
\newblock Learning {S}hape {A}bstractions by {A}ssembling {V}olumetric
  {P}rimitives.
\newblock In {\em CVPR}, 2017.

\bibitem{ulyanov2016instance}
Dmitry Ulyanov, Andrea Vedaldi, and Victor Lempitsky.
\newblock Instance normalization: {T}he missing ingredient for fast
  stylization.
\newblock In {\em CoRR}, 2016.

\bibitem{vincent2001seqransac}
Etienne Vincent and Robert Lagani{\'e}re.
\newblock Detecting planar homographies in an image pair.
\newblock In {\em ISPA}, 2001.

\bibitem{WangZLFLJ18}
Nanyang Wang, Yinda Zhang, Zhuwen Li, Yanwei Fu, Wei Liu, and Yu{-}Gang Jiang.
\newblock Pixel2{M}esh: {G}enerating 3{D} {M}esh {M}odels from {S}ingle {RGB}
  {I}mages.
\newblock In {\em ECCV}, 2018.

\bibitem{Xiao2012:Localizing}
Jianxiong Xiao, Bryan Russell, and Antonio Torralba.
\newblock Localizing 3{D} cuboids in single-view images.
\newblock In {\em NeurIPS}, 2012.

\bibitem{Yin2018:GeoNet}
Zhichao Yin and Jianping Shi.
\newblock Geo{N}et: {U}nsupervised {L}earning of {D}ense {D}epth, {O}ptical
  {F}low and {C}amera {P}ose.
\newblock In {\em CVPR}, 2018.

\bibitem{Zhang2017:DeepContext}
Yinda Zhang, Mingru Bai, Pushmeet Kohli, Shahram Izadi, and Jianxiong Xiao.
\newblock Deep{C}ontext: {C}ontext-{E}ncoding {N}eural {P}athways for 3{D}
  {H}olistic {S}cene {U}nderstanding.
\newblock In {\em ICCV}, 2017.

\bibitem{zou20173d}
Chuhang Zou, Ersin Yumer, Jimei Yang, Duygu Ceylan, and Derek Hoiem.
\newblock {3D-PRNN}: {G}enerating {S}hape {P}rimitives with {R}ecurrent
  {N}eural {N}etworks.
\newblock In {\em ICCV}, 2017.

\end{thebibliography}
}

\end{document}